\documentclass[acmlarge]{acmart}

\usepackage{subfig}
\usepackage[capitalise,noabbrev]{cleveref}

\AtBeginDocument{%
  \providecommand\BibTeX{{%
    \normalfont B\kern-0.5em{\scshape i\kern-0.25em b}\kern-0.8em\TeX}}}

\acmJournal{HEALTH}

\usepackage{tikz}
\usepackage{pgfplots}
\pgfplotsset{compat=1.16} 
\usepgfplotslibrary{fillbetween}

\definecolor{gold}{rgb}{1,0.84,0.0}



\begin{document}
\title{A Framework for Generating Summaries from Temporal Personal Health Data}

\author{Jonathan J. Harris}
\orcid{0000-0002-8823-6602}
\email{harrij15@rpi.edu}
\affiliation{%
  \institution{Rensselaer Polytechnic Institute}
  \city{Troy}
  \state{NY}
  \country{USA}
}

\author{Ching-Hua Chen}
\orcid{0000-0002-1020-0861}
\email{chinghua@us.ibm.com}
\affiliation{%
  \institution{Center for Computational Health, IBM Research}
  \city{Yorktown Heights}
  \state{NY}
  \country{USA}
}

\author{Mohammed J. Zaki}
\orcid{0000-0003-4711-0234}
\authornotemark[1]
\email{zaki@cs.rpi.edu}
\affiliation{%
  \institution{Rensselaer Polytechnic Institute}
  \city{Troy}
  \state{NY}
  \country{USA}
}

\begin{abstract}
Whereas it has become easier for individuals to track their personal
health data (e.g., heart rate, step count, and nutrient intake data), there is still a
wide chasm between the collection of data and the generation of
meaningful summaries to help users better understand what their data
means to them. With an increased comprehension of their data, users will
be able to act upon the newfound information and work towards striving
closer to their health goals. We aim to bridge the gap between data
collection and summary generation by mining the data for interesting
behavioral findings that may provide hints about a user's tendencies.
Our focus is on improving the explainability of temporal personal health
data via a set of informative summary templates, or ``protoforms.''
These protoforms span both evaluation-based summaries that help users
evaluate their health goals and pattern-based summaries that explain
their implicit behaviors. In addition to individual-level summaries, the
protoforms we use are also designed for population-level summaries. We
apply our approach to generate summaries (both univariate and
multivariate) from real user health data and show that the summaries our system generates
are both interesting and useful.  
\end{abstract}



\maketitle

\section{Introduction} 

Smartphone apps and personal fitness devices have made it increasingly
easy for users to collect and monitor their personal health data. While
some of this data requires active entry by the user (e.g., dietary
behaviors), other types of data are passively and continuously collected
(e.g., physical activity, heart rate, and location). The increased ease of
data collection in the personal health domain has inspired the
quantified-self movement, where motivated individuals record almost
every aspect of their lives, including mental and physical health.
Likewise, users with chronic conditions regularly use their own health
information for health decision-making \cite{sundecision}, and
ineffective interpretation of one's data may adversely affect how they
take their medications, what they eat, how they exercise, and even how
they socialize \cite{peeldiabetes}. 

On the other hand, there are people who fit neither of the aforementioned groups and simply wish to live a healthier
lifestyle. For such people, it is widely reported that fitness devices
and health apps experience a high abandonment rate. While there are many
reported reasons for this, technology-related reasons include the lack
of desired features such as \textit{notifications} or \textit{decision support}.
Furthermore, if the user-perceived value of the data is low, this can
create a feedback loop where such a perception increases the chance of
erroneous or sparse data being recorded, which in turn lowers the
utility of the data and leads to further user disengagement
\cite{codella2018}. A key challenge for most users is often the lack of
meaningful interpretation of the health data \cite{choequantifiedself}. 

This concept can also be applied to personal health data. For those who
may wish to improve or maintain their health, it is important for
them to gain more insight into their own health logs to help them reach their personal
health goals, and to assess whether their efforts are bringing them closer to those
goals. However, individuals often find it challenging to understand
their own health data, especially when they record multiple types of
data over a long period of time. For example, in the quantified-self community,
structured recording of daily activities and outcomes is practiced
regularly. A key hurdle for this community is the extraction of high-level 
information from the sea of data and to interpret that information in a
meaningful way \cite{choequantifiedself}. This hurdle is also commonly
reported among patients living with chronic conditions
\cite{peeldiabetes}, who use data for daily decision-making on
medication dosages, food intake, and other behaviors. Ineffective
interpretation of one's data may affect the subsequent decision-making
process and anticipated health outcomes. The high frequency of data
usage by those populations makes it impractical to rely solely on
medical professionals to interpret their data. Automated methods to
support data interpretation is therefore an urgent need.

Today, a common approach to obtaining expert-generated information on
improving or maintaining health is through a search query online or
through contact with a health expert. While searching for health
information on the Internet works for the general case, it often lacks
the personalization required to accommodate individual needs. In
particular, every person has a different health experience, as
exemplified by the uniqueness of the data collected by their health
apps; human health experts may be able to relate an individual's data to
general health knowledge, but they are expensive to engage with and there are
not enough of them. Therefore, health consumers are often left on their
own to bridge the gap between the sea of general health knowledge and
the sea of personal health data. Addressing this gap via automation
requires a combination of methods for anticipating and understanding an
individual's needs, providing an answer or recommendation for meeting
that need, and, importantly, providing an explanation for that
recommendation. While black box approaches that generate recommendations
from data without explanation may be acceptable in some domains (e.g.,
manufacturing, advertising), this is rarely the case when it comes to
personal health and healthcare. We believe that an important aspect of
data-driven recommendation involves explaining how the data itself is
being interpreted, and how it can be used to support explanations of
downstream algorithms to produce a recommendation. 

\textit{The main motivation of our work is the need of
individuals (who wish to improve their health) to better understand
their past behaviors based on their personal health data that may be
inhibiting them from reaching their health goals. With additional
comprehension via a natural language summary and a refined focus on key
aspects of their health data, they will have the ability to take action by
making appropriate changes to their routine.}
We address this problem by creating a framework that 
provides individuals with \textit{personalized natural language summaries} based on
behavioral patterns found within their time-series data. 
Generating explainable summaries from personal health data is a
challenging task. Within the field of summarization, there are three
main approaches when it comes to linguistic summary generation:
probabilistic/statistical, neural, and rule-based methods
\cite{vanderlee}. Whereas state-of-the-art
probabilistic/statistical and neural methods generate the sentences
automatically, the textual output of these approaches
is of lower quality than that of the rule-based approach \cite{vanderlee}. 
Our work, therefore,
utilizes a rule-based approach inspired by the principles presented in \cite{prototype,fuzzyquant,ZADEH1975} and the database summarization approach in \cite{businessdata}.
Most existing methods within the
linguistic summarization community do not handle time-series data; the
few approaches that do either generate longer narratives of the data
\cite{gatt} or summaries of simpler trends \cite{trendsapproach}, such
as whether the trend is increasing, concave, etc. 
Our unsupervised approach takes advantage of a more comprehensive
set of summaries along with the use of time-series data mining methods
to generate more meaningful summaries.

\textit{Our work focuses on improving the explainability of personal
health data by generating temporal summaries in natural language from
time-series health data.} We propose a comprehensive framework to
generate summaries that can help a user evaluate their personal health
data, and compare their data against general health guidelines or goals.
In particular, we propose a systematic classification of summary types
that cover a wide range of applications in personal health, including
evaluation-based summaries that help users evaluate their health goals,
and pattern-based summaries that explain their ``hidden'' behaviors.
Our approach extracts temporal patterns from data
and generates clear and concise summaries. 
In particular, our summaries are based on a categorical (or symbolic) representation of
time-series data (via SAX~\cite{sax}), combined with frequent sequence
pattern mining (via SPADE~\cite{spade}) and
categorical clustering (via Squeezer~\cite{squeezer}), allowing us to
generate understandable descriptions of hidden and implicit trends (both
within and across multiple time series) that
are not obvious from the raw data. To generate these summaries, we employ
the linguistic summarization approach~\cite{prototype,yagerapproach}
that relies on our proposed time-series {\em protoforms} to describe comprehensible
natural language findings in personal health data.
A protoform is a sentence prototype or template with placeholders or
blanks that
are automatically chosen to reflect trends and patterns supported by
the data. For
example, consider the protoform  \textit{On $\langle$quantifier$\rangle$ $\langle$sub-time
window$\rangle$ in the past $\langle$time window$\rangle$, your $\langle$attribute$\rangle$ was
$\langle$summarizer$\rangle$}, where the blanks (represented as
$\langle$\textit{blank}$\rangle$) are of different types and must satisfy different constraints. 
An example summary generated from this 
protoform is: \textit{On most of the days in the past week, your step count
was high.} Here, the $\langle$quantifier$\rangle$ is `most of the.'
Each summary explains a particular pattern
for an attribute (or a set of attributes) to help the user make sense of
their own data.

It is important to note that our work significantly extends existing
approaches for both linguistic summarization and time-series data
mining. For example, whereas existing time-series data
mining methods extract patterns and trends, we extend them by generating
understandable summaries.
Likewise, existing linguistic summarization approaches primarily focus on
tabular and non-temporal data, whereas our approach is especially
focused on time-series data. Furthermore, our framework extends current 
time-series summarization approaches
\cite{businessdata,processes,trends,trendsapproach,kacprzyk2008linguistic}
both in terms of diversity of summaries that can be produced, 
as well as what patterns can be found within the data via mining. 
With the generation of natural language summaries from both univariate
and  multivariate temporal personal health data, our system can provide
important clues to a better understanding of a user's general behavior,
and can facilitate actionable changes to fix areas where they may be
falling short of their health goals. 
To summarize, our work makes the following significant contributions:
\begin{itemize}
    \item We propose a comprehensive framework of informative
        time-series protoforms
        to produce both evaluation-based and pattern-based summaries
        using time-series data mining methods.
        In particular, we provide a systemic classification of summary types to be
        applied to the temporal personal health domain.
    \item We generate meaningful natural language summaries from both
        univariate and multivariate time-series data to highlight hidden
        patterns found within and between multiple variables. Our
        approach also illustrates summary provenance via charts highlighting
        the appropriate data and/or pattern that support the summaries.
        A preliminary user evaluation confirms the usefulness and
    comprehensibility of our summaries.
    \item We showcase and evaluate the usefulness of the summaries on
        real user data including food nutrient logs (using MyFitnessPal
        dataset~\cite{mfp}) and fitness data (using 
        Insight4Wear~\cite{rawassizadeh_scalable_2016} dataset). 
        We highlight interesting summaries obtained from users'
        nutrient intake, heart rate, and step count data.
    \item We show that our framework is general and can be applied to
        different domains in addition to personal health. We illustrate
        the generalizability by showing summaries from weather and stock
    market data.
\end{itemize}

\section{Related Works}
\subsection{Data-To-Text Generation}
\label{sec:related}
In general, data-to-text generation methods include statistical \cite{koehn} and
neural \cite{klein} machine translation, and rule-based linguistic
summarization \cite{boran2016overview}.  Neural and statistical
methods rely on the use of automation to produce natural language
summaries, relying on measures like the BLEU score \cite{bleu} to
evaluate these summaries (the BLEU score measures the correspondence
between the algorithm output and the reference human sentences).
Rule-based methods typically use semantically meaningful templates or protoforms to generate their output,
and thus obviate the need for measures like the BLEU score. Instead, they
rely on human evaluation to judge this utility, but they can use
objective measures such as significance, frequency, and other metrics \cite{boran2016overview} to judge the quality of the
summary output.
\citet{vanderlee}
compared the performance and text quality between rule-based, neural,
and statistical methods. Their main and important conclusion is that rule-based methods generally
perform faster and produce higher text quality, although the manual
creation of the sentence prototypes is time-intensive. Furthermore, they observe that rule-based methods are generally restricted to more simple situations and may be less useful in more complex cases. On the other hand, statistical- and learning-based methods avoid manual creation of prototypes,
but are generally lacking in performance and text quality. 
It is important to note a significant drawback of the supervised neural
and statistical text generation methods: they typically need
a large set of training pairs containing the input data and the desired
natural language summary, which is often not available.
{\em Given that text
quality is extremely important within the personal health domain,
and given the lack of training data comprising input raw time-series
data and their corresponding natural language health summaries, we follow the rule-based
linguistic paradigm.}

\subsection{Time-Series Summarization}
Our work builds upon and extends linguistic database summarization methods \cite{businessdata,processes,trends,trendsapproach,kacprzyk2008linguistic}
that rely on the concept of protoforms and fuzzy logic
\cite{boran2016overview,prototype} to summarize data.
Linguistic summarization methods have also been applied to 
time-series data in various domains, 
such as elderly care \cite{eldercare}, physical activity
tracking \cite{selftrack}, driving simulation environments
\cite{driving}, deforestation analysis \cite{deforestation}, human gait
study
\cite{gait}, periodicity detection \cite{period}, time-series
forecasting \cite{forecast}, and generation of longer temporal
``narratives'' from neonatal intensive care data via the use of a
neonatal ontology \cite{gatt}. Other work includes the use of genetic
algorithms \cite{genetic} to generate linguistic summaries from
time series, and those that place emphasis on simple trends (e.g.,
increasing, concave) \cite{trendsapproach}. {\em In contrast to these
    works, we propose a more comprehensive set of summaries, and unlike
    all previous time-series summary-based works, we also apply data
    mining to discover interesting patterns across multiple variables to
produce more interesting summaries.~\footnote{While there is work on
summaries spanning multiple variables in the context of neonatal
intensive care data \cite{gatt}, that work is based on a neonatal
ontology and does not perform multivariate temporal pattern mining as in our
work.}} 

More recent work on time-series summaries includes \cite{murakami2017}
that
uses a neural encoder-decoder model to generate natural language summaries in
the financial domain, and \cite{aoki2018} that extends the model to
multiple external factors (e.g., relationships between the Nikkei and
Dow Jones stock market data). For training, \citet{murakami2017} pair a
time series with a market comment that aligns with it. They
used 16,276 headlines gathered from Nikkei Quick News (NQN) to
train their model. \citet{aoki2018}, in addition, use five-minute charts of seven stock market
indices from Thomson Reuters DataScope
Select\footnote{https://hosted.datascope.reuters.com/DataScope/} as
an external resource; however, summaries generated by both of these methods are limited to relatively simple conclusions (e.g., a
continual rising trend). As such, neural network based methods suffer
from several drawbacks, such as the aforementioned lack of high quality
text
\cite{vanderlee}, dependence on large training data and/or supervision
(which is not available for our personal health domain),
and lack of ability to explain patterns directly from raw temporal
data. 
{\em In contrast, our system is unsupervised and generates summaries
    that explain interesting patterns and trends that are not
immediately apparent (based on pattern mining and clustering)}.

The closest work to ours is \cite{temporalkc}, where they generate
linguistic descriptions of multivariate data via feature extraction,
primitive pattern extraction via neural networks, and rule generation
\cite{sig}. In their work, they use unsupervised neural networks to find primitive patterns of
events in time series with natural language names assigned to the
primitive patterns in a semi-automated manner.
An example summary of an event from their work is:
{\bf If} ``no airflow without snoring'' {\bf is more or less
simultaneous} ``no chest
and abdomen wall movements without snoring,'' 
where phrases (e.g., ``no'') resulting from fuzzy-membership functions are paired
with measured attributes (e.g., ``airflow'', ``chest wall movements'').
In our work, we mine frequent sequences to generate more interesting if-then pattern summaries, as well as
cluster-based pattern summaries. Our framework is also able to provide many more
informative summaries based on the comprehensive set of univariate and
multivariate protoforms. 

\subsection{Time-Series Data Mining}
There are many works on time-series data mining, reviewed by
\citet{BATYRSHIN2008}, including 
the construction of rules based on patterns found in the data
\cite{Das}, using derivatives to describe the concavity/convexity of
trends \cite{cheung1}, identification of pre-determined patterns using
shape descriptors \cite{baldwin}, transformation of time series into
state intervals to create association rules \cite{Hoppner}, generating
reports about stocks \cite{reiter}, and so on. 
{\em It is important to note that these approaches
find temporal patterns or rules based on shapes and trends, but they do
not generate explanations. In contrast, our work tries to explain the
important patterns via temporal natural language summaries.}

Our work utilizes the Symbolic Aggregate approXimation
    (SAX)~\cite{sax}
    approach to discretize time-series data into a symbolic sequence
from which we can mine patterns and trends. SAX is very effective for
motif discovery, dimensionality reduction, and other data mining tasks on time-series data.
Other related work includes Symbolic Fourier Approximation
(SFA)~\cite{sfa}, which is more data adaptive; ABBA \cite{abba} that
aims to better preserve shapes via adaptive polygonal chain
approximation and mean-based clustering; and cSAX \cite{csax}
that incorporates complexity invariance within SAX. In the future, we
plan to explore these alternative methods in terms of their effect on the mined
patterns for summary generation.

\section{Temporal Summaries for Personal Health Data}
\label{sec:summaries}


We begin by defining basic concepts we will use in the remainder of this
paper:
\begin{itemize}
    \item \textbf{Protoform} ($P$): A sentence prototype (or template)
        that can be used to generate a natural language summary
    \item \textbf{Summarizer} ($S$): A conclusive phrase for a summary
    \item \textbf{Quantifier} ($Q$): A word or phrase that specifies how
        often the summarizer $S$ is true
    \item \textbf{Attribute} ($A$): A variable of interest
    \item \textbf{Time window} ($TW$): A time window of interest
    \item \textbf{Sub-time window} ($sTW$): A time window at a smaller
        granularity than $TW$
    \item \textbf{Qualifier} ($R$): a word or phrase that adds more
        specificity to a summary
\end{itemize}

Given a set of quantifiers \textbf{Q}, a set of summarizers \textbf{S},
a specified time window granularity TW and sub-time window
granularity sTW, a set of protoforms \textbf{P}, and a set of
time series \textbf{T} for a corresponding set of attributes \textbf{A},
we generate natural language summaries of behavioral
patterns found in temporal personal health data. 
For example, consider the summary ``\textit{On most of the
days in the past week, your calorie intake was high},'' 
generated from the protoform 
``\textit{On $Q$ $sTW$ in the past $TW$, your $A$ was $S$},'' 
the quantifier $Q$ (``\textit{most of the}'')
represents how often the finding is found to be true in the data, the
attribute $A$
(`\textit{`calorie intake}'') represents the variable of interest, and the summarizer
$S$ (``\textit{high}'') represents the conclusion from the data. Here, the time
window $TW$ is ``\textit{weeks}'' and the sub-time window $sTW$ is ``\textit{days}.''
We use different protoforms to generate more complex
summaries describing interesting patterns within and across variables.



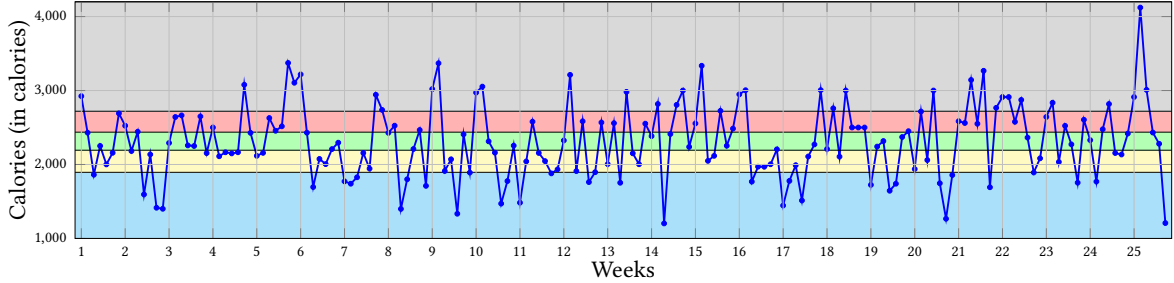
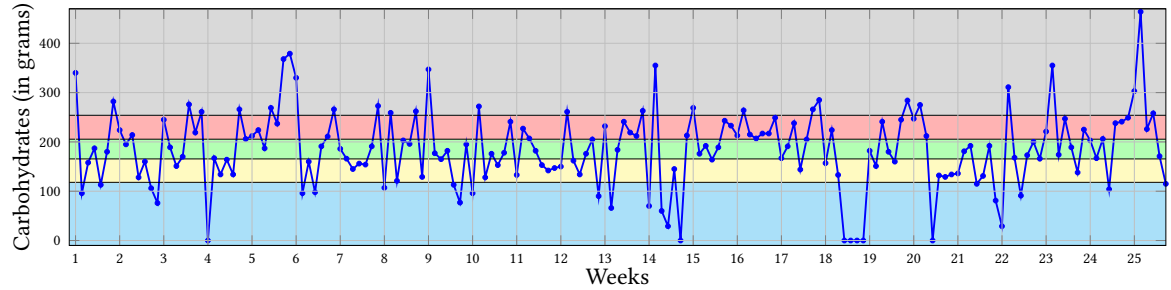
\begin{figure}[!ht]
    \centering
    \pgfplotsset{every axis/.append style = {
            tick label style={font=\tiny}},
            x coord trafo/.code={\pgfmathparse{(#1-1)/7+1}\pgfmathresult},
        x coord inv trafo/.code={\pgfmathparse{#1}\pgfmathresult},
    }
    \centerline{
        \hspace{-0.25in}
        \subfloat[Calorie Intake Data]{
        \label{fig:caloriedata}
            \scalebox{0.9}{
            \begin{tikzpicture}
            \begin{axis}[name=calorie_plot,%
                axis on top,
                width=7in, height=2in,
                grid=both,%
                xlabel=Weeks, xmin=0, xmax=175, ymin=1000,ymax=4200,%
                xtick distance=1, mark size = 1pt,%
                ylabel=Calories (in calories),%
                label shift = -5pt,
                ytick distance=1000, thick]
                \addplot+[color=blue, mark=*, mark options={fill=blue}] 
                    table[x=Day,y=Calories,col sep=comma]  {calorie_data.csv};
                \addplot[name path=A,black,no markers, line width=0pt]
                    coordinates {(0,0) (175,0)};
                \addplot[name path=B,black, no markers, line width=0pt]
                    coordinates {(0,1894) (175,1894)};
                \addplot[name path=C,black, no markers, line width=0pt]
                    coordinates {(0,2193) (175,2193)};
                \addplot[name path=D,black, no markers, line width=0pt]
                    coordinates {(0,2437) (175,2437)};
                \addplot[name path=E,black, no markers, line width=0pt]
                    coordinates {(0,2720) (175,2720)};
                \addplot[name path=F,black, no markers, line width=0pt]
                    coordinates {(0,4200) (175,4200)};
                \addplot[cyan!30] fill between[of=A and B];                
                \addplot[yellow!30] fill between[of=B and C];                
                \addplot[green!30] fill between[of=C and D];                
                \addplot[red!30] fill between[of=D and E];                
                \addplot[gray!30] fill between[of=E and F];                
            \end{axis}
            \end{tikzpicture}
        }}}
    \centerline{
        \hspace{-0.25in}
        \subfloat[Carbohydrate Intake Data]{
        \label{fig:carbdata}
            \scalebox{0.9}{
            \begin{tikzpicture}
            \begin{axis}[name=carb_plot,%
                axis on top,
                label shift = -5pt,
                width=7in, height=2in,
                grid=major, mark size=1pt, %
                xlabel=Weeks, xmin=0, xmax=174, ymin=-10, ymax=470,%
                xtick distance=1, ylabel=Carbohydrates (in grams),%
                ytick distance=100, thick]
            \addplot+[color=blue, mark=*, mark options={fill=blue}] 
                table[x=Day,y=Carbohydrates,col sep=comma]  {carb_data.csv};
                \addplot[name path=A,black,no markers, line width=0pt]
                    coordinates {(0,-10) (175,-10)};
                \addplot[name path=B,black, no markers, line width=0pt]
                    coordinates {(0,118) (175,118)};
                \addplot[name path=C,black, no markers, line width=0pt]
                    coordinates {(0,165.5) (175,165.5)};
                \addplot[name path=D,black, no markers, line width=0pt]
                    coordinates {(0,205.5) (175,205.5)};
                \addplot[name path=E,black, no markers, line width=0pt]
                    coordinates {(0,254) (175,254)};
                \addplot[name path=F,black, no markers, line width=0pt]
                    coordinates {(0,470) (175,470)};
                \addplot[cyan!30] fill between[of=A and B];                
                \addplot[yellow!30] fill between[of=B and C];                
                \addplot[green!30] fill between[of=C and D];                
                \addplot[red!30] fill between[of=D and E];                
                \addplot[gray!30] fill between[of=E and F];                
            \end{axis}
            \end{tikzpicture}
    }}
}
\vspace{-0.1in}
\caption{Calorie (a) and Carbohydrate (b) Intake Data for a user from
MyFitnessPal dataset~\cite{mfp}. The different colored regions
correspond to the different summarizers -- very low, low, moderate, high,
very high -- from bottom to top.}
        \label{fig:carbcaloriedata}
\vspace{-0.1in}
\end{figure}

\begin{table}[!ht]
\vspace{-0.1in}
    \centering
    \caption{Variable Assignments}
    \label{tab:assignments}
    \begin{tabular}{c|c}
    \toprule
    
        \textbf{Q} & \begin{tabular}{@{}c@{}} none of the, almost none of the, some of the, \\half of the, more than half of the, most of the, all of the\end{tabular}\\\hline
        \textbf{A} & calorie intake, carbohydrate intake \\\hline
        TW & weekly granularity \\\hline
        sTW & daily granularity \\
    \bottomrule
    \end{tabular}
\end{table}


\paragraph{Running Example:} To instantiate summaries that illustrate
each of our protoforms, we will consider real user data from 
the MyFitnessPal dataset~\cite{mfp}, in particular data on their 
intake of calories and carbohydrates for a period of about 6 months (actually
174 days). The
corresponding time series data is plotted in Figures
\ref{fig:caloriedata} and \ref{fig:carbdata}. The five horizontal 
ranges within the charts each correspond to the range of values for each summarizer in \textbf{S}.
For example, any data point within the top-most (gray) range can be
described as ``very high.'' These regions are found using
SAX~\cite{sax}, which will be explained later in \cref{sec:approach}.
Assume that the user is interested in finding patterns in a weekly time
window and has the goal
to limit their calorie and carbohydrate intake for a \textit{2000-calorie}
diet.
Table
\ref{tab:assignments} shows some of the possible quantifiers
(\textbf{Q}), and also the attributes of interest (i.e., Calories,
Carbohydrates) and the (sub-)time window
values (i.e., day, week) for our running example.
For the remainder of this paper, we use the data in
\cref{fig:carbcaloriedata} as a running example to explain the various
protoforms.

\begin{figure}[!h]
  \centering
  \includegraphics[width=\textwidth, height=2.4in]{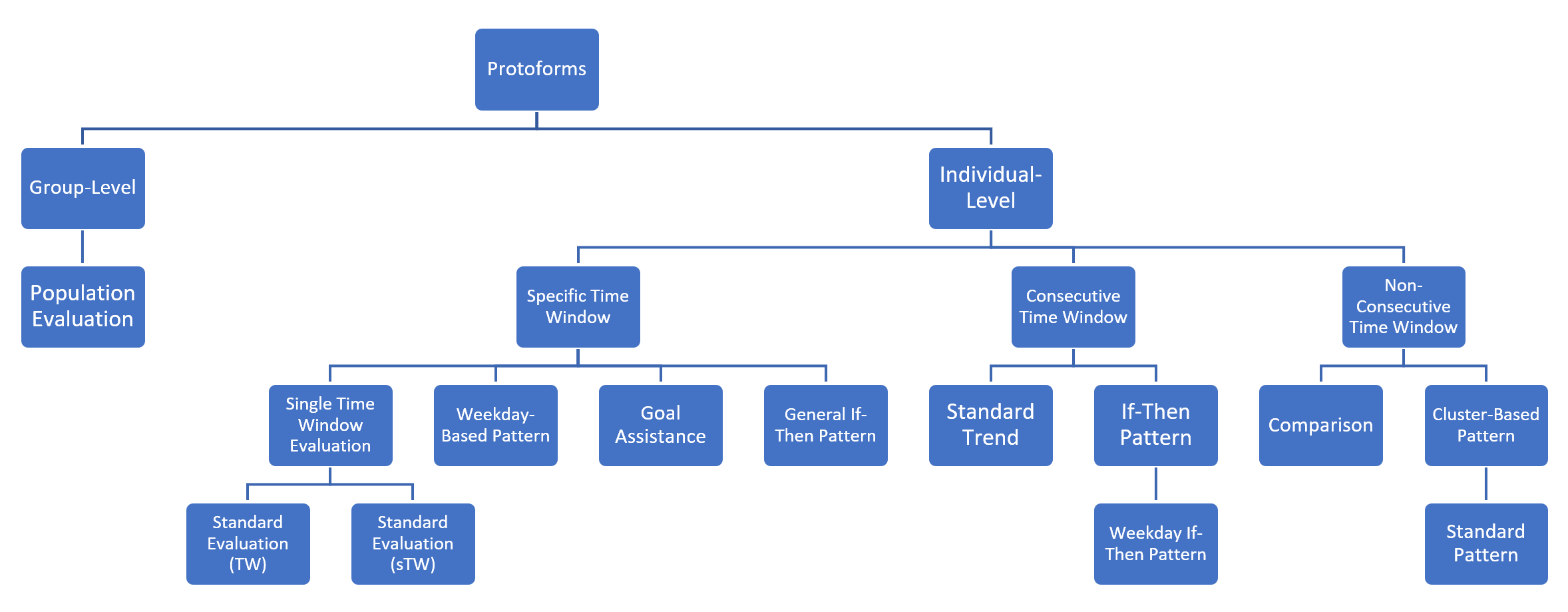}
\vspace{-0.1in}
  \caption{Hierarchy of Protoform or Summary Types. Protoforms can be
  group-level or individual-level. Within the individual level, we
  propose three types of summaries: i) based on a specific time window,
  ii) comparing two consecutive time windows, and iii) comparing any two
  time periods. Each of these is further divided into more specific
  summary types.}
  \label{fig:hierarchy}
\vspace{-0.1in}
\end{figure}



\subsection{Protoform Hierarchy}
We seek to automatically generate a diverse set of summaries of
time-series data related to a user's personal health. It
is important that we have a diverse set to allow us to take into
account the various ways we can look at a user's data. Each
kind of protoform our system generates highlights a different aspect from the
user's data that
may be helping or hurting their efforts to reach their personal health
goals.
As shown in Figure \ref{fig:hierarchy},  we propose
a number of different summary types
that are applicable to a wide range of personal health scenarios, and
are meant to be both useful and comprehensive. 
In particular, we propose 
three types of individual-level summaries: 1) specific time window summaries,
which look at trends within a specified time window, 2) consecutive time window based
summaries that compare two successive time periods, and 3)
non-consecutive time window based summaries that compare different time
periods. We also propose a group-level summary that is designed to see
the patterns in a population of users. 
In addition, these summaries can be augmented with goals or
guidelines to better help the user.
These protoforms
are equally applicable to quantified-selfers or general users who
want to understand their personal data. When a user looks at their own
data, they may try to look for patterns in the data that correlate with
their daily routine. These patterns reflect their behaviors and can
provide clues as to what aids or hampers progress
towards their health goals.



In the next following subsections, we
will explore various protoform types where each protoform \textit{p} $\in$
\textbf{P} has a corresponding set of summarizers \textbf{S}.
Each summary template or protoform
requires a set of quantifiers and a unique set of summarizers as
appropriate placeholders.
Table \ref{tab:summarizers} enumerates the
different types of summarizers (\textbf{S}), whereas
\cref{tab:assignments} shows the quantifiers (\textbf{Q}).
For each summary type, we will provide
univariate and multivariate examples
using the data from the running example in \cref{fig:carbcaloriedata},
using both calorie and carbohydrate intake as input variables.
In addition to showing the natural language summaries, for better 
    explainability, we display the provenance of the data supporting the
    summaries generated using the various protoform types. In
    particular, we automatically generate corresponding time-series charts showing where the discovered patterns
were found in the data, which can be a great aid in understanding the
summaries. Furthermore, we describe the summaries from the perspective of a user
who is exploring their data looking for summaries from the simple to the
more complex, which also motivates the need for each summary type.

\begin{table}[!h]
\centering
\vspace{-0.1in}
  \caption{Summarizers by Protoform Type}
  \label{tab:summarizers}
  \begin{tabular}{p{2.2in}p{2.2in}}
    \toprule
    Protoform Type & Possible Summarizers\\\hline
    \midrule
    \texttt{Standard Evaluation} & very low, low, moderate, high, very high \\\hline
    \texttt{Standard Evaluation (w/ goal)} & reached, did not reach \\ \hline
    \texttt{Goal Assistance} & increase, decrease \\ \hline
    \texttt{Day-Based Pattern} & very low, low, moderate, high, very high \\\hline
    \texttt{Standard Trend} & increased, decreased, stayed the same \\\hline
    \texttt{If-Then Pattern} & very low, low, moderate, high, very high \\ \hline
    \texttt{Comparison} & higher, lower, about the same \\\hline
    \texttt{Comparison (w/ goal)} & better, not do as well, about the same  \\\hline
    \texttt{Cluster-Based Pattern} & rose, dropped, stayed the same \\
    \bottomrule\\
  \end{tabular}
  
\end{table}

\begin{figure}[!hbt]
  \centering
  \includegraphics[width=0.5\textwidth, height=1.5in]{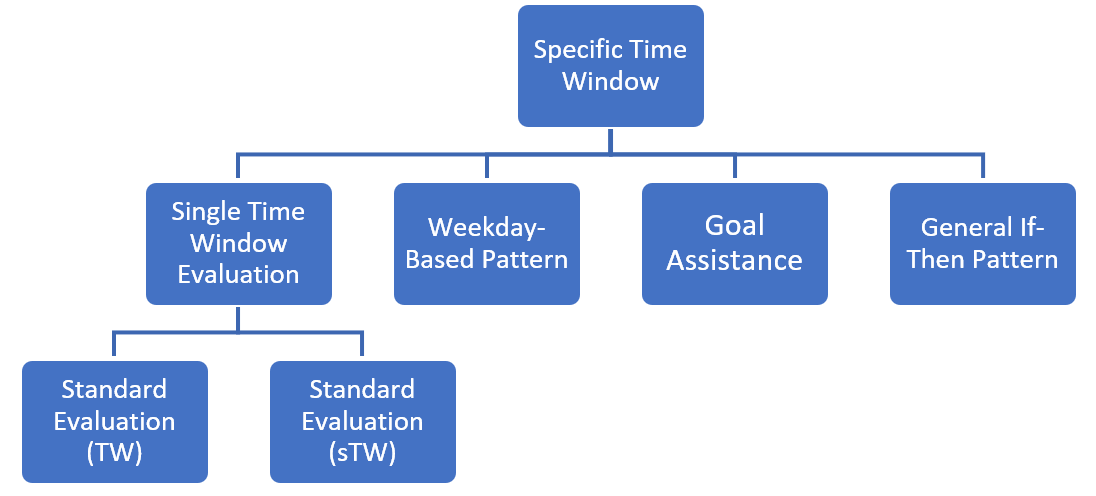}
\vspace{-0.1in}
  \caption{Specific Time Window Hierarchy}
  \label{fig:specific_tw}
\vspace{-0.1in}
\end{figure}

\subsection{Specific Time Window Summaries}
When looking for behavioral patterns in a time series, a user may want
to search for patterns within a specific time window and to evaluate
themselves within that time period. Since TW is set to a weekly
granularity and sTW is set to a daily granularity in our
running example, this user would be looking at a particular week (or
days within that week) in their intake data so they can evaluate their
progress. 
The summaries we generate for these patterns are called specific time
window summaries; this sub-hierarchy of summaries is shown in Figure
\ref{fig:specific_tw}. Specific summary types
  include standard evaluation summaries, those
  that elicit the particular weekdays, those that aid goal assistance,
  and those based on if-then patterns, all within a given time window
  (e.g., within the past week).

\subsubsection{Standard evaluation summaries:}
Standard evaluation summaries are descriptions of evaluations made over
the specified time window by pairing the standard evaluation summarizers
from Table \ref{tab:summarizers} with the ``best'' quantifier from Table
\ref{tab:assignments}. These summaries contain conclusions drawn from
both TW and sTW and use summarizer set \textbf{S} =
\{\textit{very low}, \textit{low}, \textit{moderate}, \textit{high}, \textit{very high}\}.

Suppose the user is interested in knowing how well they have been doing
for the past week. In order to find this information in \textbf{T}, the
user may compare the past week with other weeks in the data. Our
framework can generate standard evaluation summaries at the TW
granularity with this protoform:
\begin{description}
\small
\item \textbf{Standard Evaluation Protoform} (TW granularity): In the
    past full $\langle$time window$\rangle$, your $\langle$attribute 1$\rangle$ has been
    $\langle$summarizer 1$\rangle$,..., and your $\langle$attribute $n\rangle$ has been
    $\langle$summarizer $n\rangle$.\\~\\
\textbf{Univariate Example} (TW granularity): In the past full
    \textbf{week}, your \textbf{calorie intake} has been
    \textbf{moderate}.\\
\textbf{Multivariate Example} (TW granularity): In the past full
\textbf{week}, your \textbf{calorie intake} has been \textbf{moderate} and
your \textbf{carbohydrate intake} has been \textbf{moderate}.
\end{description}



\begin{figure}[!t]
  \centering
    \pgfplotsset{every axis/.append style = {
            tick label style={font=\tiny}},
            x coord trafo/.code={\pgfmathparse{(#1-1)/7+1}\pgfmathresult},
        x coord inv trafo/.code={\pgfmathparse{#1}\pgfmathresult},
    }
    \centerline{
        \hspace{-0.25in}
        \subfloat[Calorie Intake Data]{
          \label{fig:calorie_setw}
            \scalebox{0.9}{
            \begin{tikzpicture}
            \begin{axis}[%
                axis on top,
                width=7in, height=1.75in,
                label shift = -5pt,
                grid=both,%
                xlabel=Weeks, xmin=0, xmax=175, ymin=1000,ymax=4200,%
                xtick distance=1, mark size = 1pt,%
                ylabel=Calories,%
                ytick distance=1000, thick]
                \addplot+[color=blue, mark=*, mark options={fill=blue}] 
                    table[x=Day,y=Calories,col sep=comma]  {calorie_data.csv};
                \addplot[name path=C,black, no markers, line width=0pt]
                    coordinates {(0,2205) (175,2205)};
                \addplot[name path=D,black, no markers, line width=0pt]
                    coordinates {(0,2432) (175,2432)};
                \addplot[green!30] fill between[of=C and D];                
                \addplot[name path=vA,black, no markers, line width=0pt]
                    coordinates {(162,0) (162,4200)};
                \addplot[name path=vB,black, no markers, line width=0pt]
                    coordinates {(169,0) (169,4200)};
                \addplot[gold!70, fill opacity=0.7] fill between[of=vA and vB];                
                \addplot[color=green, no markers, line width=3pt] coordinates{
                    (162, 2300.0) (169, 2300.0)
                };
           \end{axis}
            \end{tikzpicture}
}}}
\centerline{
    \hspace{-0.25in}
        \subfloat[Carbohydrate Intake Data]{
          \label{fig:carb_setw}
            \scalebox{0.9}{
            \begin{tikzpicture}
            \begin{axis}[name=carb_plot,%
                axis on top,
                width=7in, height=1.75in,
                label shift = -5pt,
                grid=major, mark size=1pt, %
                xlabel=Weeks, xmin=0, xmax=174, ymin=-10, ymax=470,%
                xtick distance=1, ylabel=Carbohydrates,%
                ytick distance=100, thick]
            \addplot+[color=blue, mark=*, mark options={fill=blue}] 
                table[x=Day,y=Carbohydrates,col sep=comma]  {carb_data.csv};
                \addplot[name path=C,black, no markers, line width=0pt]
                    coordinates {(0,165.5) (175,165.5)};
                \addplot[name path=D,black, no markers, line width=0pt]
                    coordinates {(0,205.5) (175,205.5)};
                \addplot[green!30] fill between[of=C and D];                
                \addplot[name path=vA,black, no markers, line width=0pt]
                    coordinates {(162,-10) (162,4200)};
                \addplot[name path=vB,black, no markers, line width=0pt]
                    coordinates {(169,-10) (169,4200)};
                \addplot[gold!70, fill opacity=0.7] fill between[of=vA and vB];                
                \addplot[color=green, no markers, line width=3pt] coordinates{
                    (162, 201.1) (169, 201.1)
                };
            \end{axis}
            \end{tikzpicture}
    }}}
\vspace{-0.1in}
  \caption{Summary Provenance - Standard Evaluation (TW granularity).
  Horizontal green range denotes ``moderate,'' and the green segment the weekly
  average. Vertical gold range denotes the week of interest.}
\vspace{-0.1in}
\end{figure}
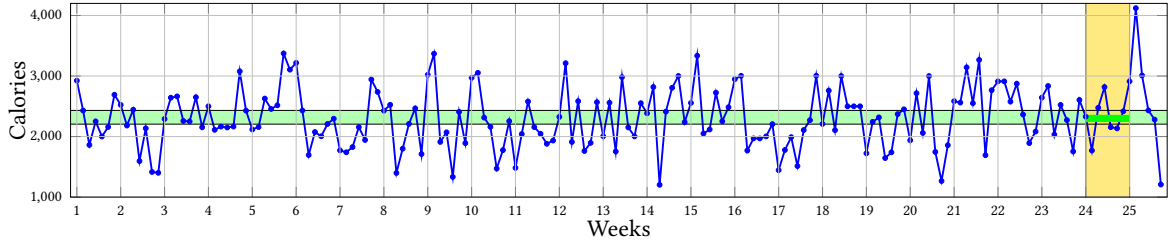
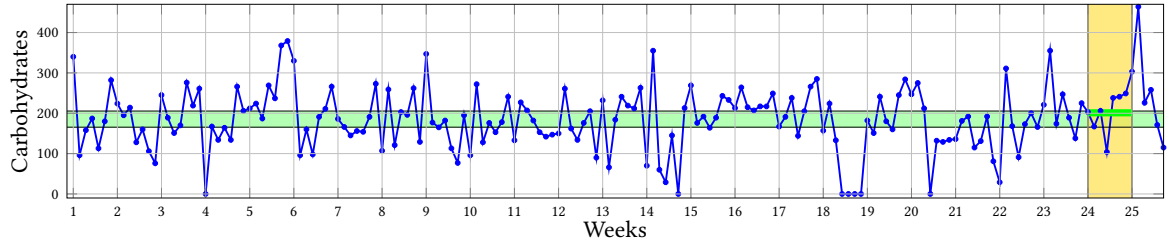

Here, $n$ is the number of attributes.
When the user receives these summaries on the TW (weekly)
granularity, they are able to evaluate their past full week as a whole
relative to other weeks in their data. 
The provenance of the example summaries above are shown in Figures \ref{fig:calorie_setw} and 
\ref{fig:carb_setw}, where the green range 
represents the summarizer ``moderate'' and the gold
vertical range represents the time window of interest, which corresponds
to the last {\em full} week, namely week 24.

Suppose the user wants more detail on what happened during the past week? We can switch to the
sTW (daily) sub-time window granularity by looking at summaries modeled
by the following
protoform:
\begin{description}
\small
\item \textbf{Standard Evaluation Protoform} (sTW granularity): On
    $\langle$quantifier$\rangle$ $\langle$sub-time window$\rangle$ in the past $\langle$time
    window$\rangle$, your $\langle$attribute 1$\rangle$ was $\langle$summarizer 1$\rangle$,..., and
    your $\langle$attribute $n\rangle$ was  $\langle$summarizer $n\rangle$.\\~\\
\textbf{Univariate Example} (sTW granularity): On \textbf{some of the
\textit{days}} in the past \textbf{week}, your \textbf{calorie intake}
has been \textbf{low}.\\
\textbf{Multivariate Example} (sTW granularity): On \textbf{some of the}
\textbf{\textit{days}} in the past \textbf{week}, your \textbf{calorie
intake} has been \textbf{low} and your \textbf{carbohydrate intake} has
been \textbf{high}.
\end{description}






\begin{figure}[!t]
  \centering
    \pgfplotsset{every axis/.append style = {
            tick label style={font=\tiny}},
            x coord trafo/.code={\pgfmathparse{(#1-1)/7+1}\pgfmathresult},
        x coord inv trafo/.code={\pgfmathparse{#1}\pgfmathresult},
    }
    \centerline{
        \hspace{-0.25in}
        \subfloat[Calorie Intake Data]{
          \label{fig:calorie_sestw}
            \scalebox{0.9}{
            \begin{tikzpicture}
            \begin{axis}[%
                axis on top,
                width=7in, height=1.75in,
                grid=both,%
                label shift = -5pt,
                xlabel=Weeks, xmin=0, xmax=175, ymin=1000,ymax=4200,%
                xtick distance=1, mark size = 1pt,%
                ylabel=Calories,%
                ytick distance=1000, thick]
                \addplot+[color=blue, mark=*, mark options={fill=blue}] 
                    table[x=Day,y=Calories,col sep=comma]  {calorie_data.csv};
                \addplot[name path=B,black, no markers, line width=0pt]
                    coordinates {(0,1894) (175,1894)};
                \addplot[name path=C,black, no markers, line width=0pt]
                    coordinates {(0,2193) (175,2193)};
                \addplot[yellow!30] fill between[of=B and C];                
                \addplot[name path=vA,black, no markers, line width=0pt]
                    coordinates {(162,0) (162,4200)};
                \addplot[name path=vB,black, no markers, line width=0pt]
                    coordinates {(169,0) (169,4200)};
                \addplot[gold!70, fill opacity=0.7] fill between[of=vA and vB];                
                \addplot[only marks, color=red] coordinates {
                        (166, 2155.0)
                        (167, 2135.0)
                    };
            \end{axis}
            \end{tikzpicture}
        }}
    }
    \centerline{
        \hspace{-0.25in}
        \subfloat[Carbohydrate Intake Data]{
          \label{fig:carb_sestw}
            \scalebox{0.9}{
            \begin{tikzpicture}
            \begin{axis}[name=carb_plot,%
                axis on top,
                width=7in, height=1.75in,
                grid=major, mark size=1pt, %
                label shift = -5pt,
                xlabel=Weeks, xmin=0, xmax=174, ymin=-10, ymax=470,%
                xtick distance=1, ylabel=Carbohydrates,%
                ytick distance=100, thick]
            \addplot+[color=blue, mark=*, mark options={fill=blue}] 
                table[x=Day,y=Carbohydrates,col sep=comma]  {carb_data.csv};
                \addplot[name path=D,black, no markers, line width=0pt]
                    coordinates {(0,205.5) (175,205.5)};
                \addplot[name path=E,black, no markers, line width=0pt]
                    coordinates {(0,254) (175,254)};
                \addplot[red!30] fill between[of=D and E];                
                %
                \addplot[name path=vA,black, no markers, line width=0pt]
                    coordinates {(162,0) (162,4200)};
                \addplot[name path=vB,black, no markers, line width=0pt]
                    coordinates {(169,0) (169,4200)};
                \addplot[gold!70, fill opacity=0.7] fill between[of=vA and vB];                
                \addplot[only marks, color=red] coordinates {
                        (164, 206.0)
                        (166, 238.0)
                        (167, 241.0)
                        (168, 249.0)
                    };
            \end{axis}
            \end{tikzpicture}
    }}}
\vspace{-0.1in}
  \caption{Summary Provenance - Standard Evaluation (sTW granularity).
  The (horizontal) yellow region denotes ``low'' and the red region denotes
  ``high.'' The vertical range in gold highlights the week of interest.}
\vspace{-0.1in}
\end{figure}
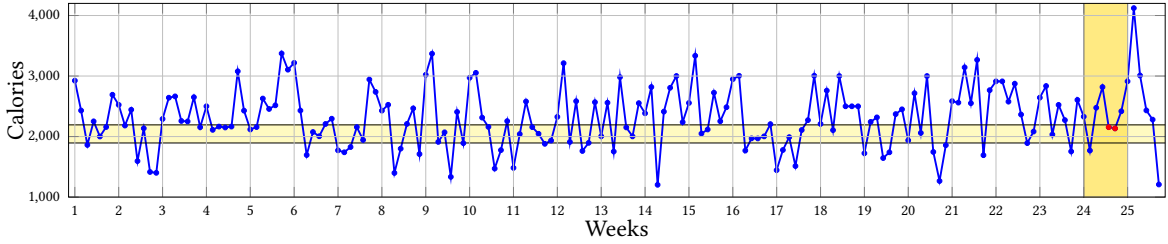
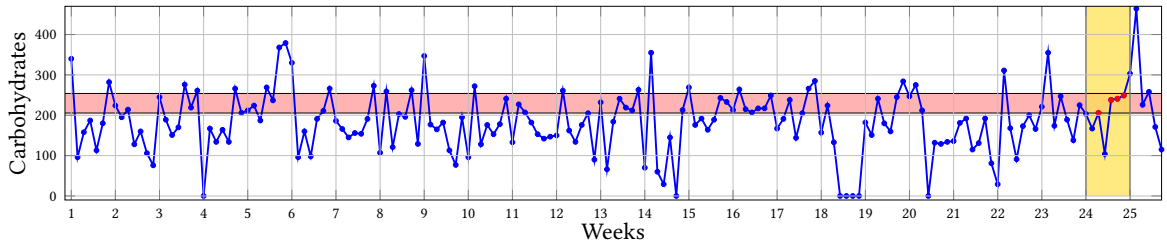

With these summaries, the user gains the knowledge that their calorie
intake was actually low on some of the days in the past week. Figures \ref{fig:calorie_sestw} and 
    \ref{fig:carb_sestw} shows specific days (red points) that support the summary. 
    In these charts, the yellow range 
represents the ``low'' summarizer, while the red range represents the
``high'' summarizer. The vertical range in gold represents the time period of
focus with relevant data points in red. They can
use these summaries to examine  
their behavior on those days, and try to get
closer to their goals. The multivariate example implies that there may
be a behavioral pattern between the user's calorie and carbohydrate
intake.

Often, the user may be interested in specific conditions under which a
pattern manifests itself. The following protoform can be used when
enhanced with a qualifier, which adds more context:

\begin{description}
\small
\item \textbf{Standard Evaluation Protoform} (w/ qualifier): On
    $\langle$quantifier$\rangle$ $\langle$sub-time window$\rangle$ in the past $\langle$time window$\rangle$
    $\langle$qualifier$\rangle$, your $\langle$attribute $n+1\rangle$ was $\langle$summarizer
    $n+1\rangle$,...\\~\\
\textbf{Multivariate Example} (sTW granularity w/ qualifier): On
\textbf{all of the} \textbf{\textit{days}} in the past \textbf{week},
\textbf{\textit{when your calorie intake was very low}}, your
\textbf{carbohydrate intake} was \textbf{moderate}.
\end{description}



\begin{figure}[!t]
  \centering
    \pgfplotsset{every axis/.append style = {
            tick label style={font=\tiny}},
            x coord trafo/.code={\pgfmathparse{(#1-1)/7+1}\pgfmathresult},
        x coord inv trafo/.code={\pgfmathparse{#1}\pgfmathresult},
    }
    \centerline{
        \hspace{-0.25in}
        \subfloat[Calorie Intake Data]{
  \label{fig:calorie_sestwq}
            \scalebox{0.9}{
            \begin{tikzpicture}
            \begin{axis}[%
                axis on top,
                width=7in, height=1.75in,
                grid=both,%
                label shift = -5pt,
                xlabel=Weeks, xmin=0, xmax=175, ymin=1000,ymax=4200,%
                xtick distance=1, mark size = 1pt,%
                ylabel=Calories,%
                ytick distance=1000, thick]
                \addplot+[color=blue, mark=*, mark options={fill=blue}] 
                    table[x=Day,y=Calories,col sep=comma]  {calorie_data.csv};
                \addplot[name path=A,black,no markers, line width=0pt]
                    coordinates {(0,0) (175,0)};
                \addplot[name path=B,black, no markers, line width=0pt]
                    coordinates {(0,1894) (175,1894)};
                \addplot[cyan!30] fill between[of=A and B];                
                \addplot[name path=vA,black, no markers, line width=0pt]
                    coordinates {(162,0) (162,4200)};
                \addplot[name path=vB,black, no markers, line width=0pt]
                    coordinates {(169,0) (169,4200)};
                \addplot[gold!70, fill opacity=0.7] fill between[of=vA and vB];                
                \addplot[only marks, color=red] coordinates {
                        (163, 1769.0)
                    };
            \end{axis}
            \end{tikzpicture}
        }}
    }
    \centerline{
        \hspace{-0.25in}
        \subfloat[Carbohydrate Intake Data]{
  \label{fig:carb_sestwq}
            \scalebox{0.9}{
            \begin{tikzpicture}
            \begin{axis}[name=carb_plot,%
                axis on top,
                width=7in, height=1.75in,
                grid=major, mark size=1pt, %
                label shift = -5pt,
                xlabel=Weeks, xmin=0, xmax=174, ymin=-10, ymax=470,%
                xtick distance=1, ylabel=Carbohydrates,%
                ytick distance=100, thick]
            \addplot+[color=blue, mark=*, mark options={fill=blue}] 
                table[x=Day,y=Carbohydrates,col sep=comma]  {carb_data.csv};
                \addplot[name path=C,black, no markers, line width=0pt]
                    coordinates {(0,165.5) (175,165.5)};
                \addplot[name path=D,black, no markers, line width=0pt]
                    coordinates {(0,205.5) (175,205.5)};
                \addplot[green!30] fill between[of=C and D];                
                %
                \addplot[name path=vA,black, no markers, line width=0pt]
                    coordinates {(162,-10) (162,4200)};
                \addplot[name path=vB,black, no markers, line width=0pt]
                    coordinates {(169,-10) (169,4200)};
                \addplot[gold!70, fill opacity=0.7] fill between[of=vA and vB];                
                \addplot[only marks, color=red] coordinates {
                        (163, 167.0)
                    };
            \end{axis}
            \end{tikzpicture}
    }}}
\vspace{-0.1in}
  \caption{Summary Provenance - Standard Evaluation (sTW granularity w/
  qualifier). Cyan region corresponds to ``very low,'' and green to
  ``moderate.'' The vertical region in gold denotes week of interest.}
\vspace{-0.1in}
\end{figure}
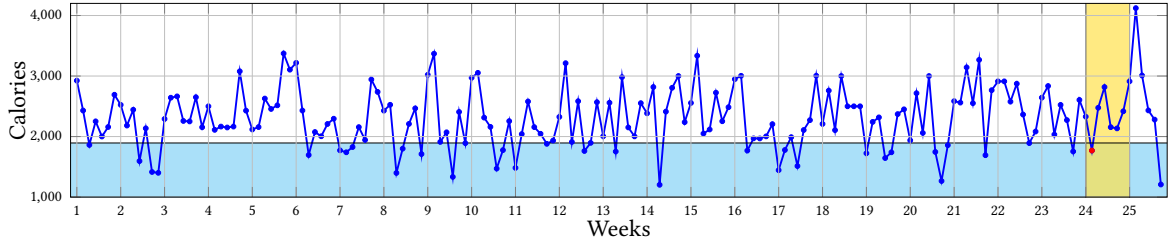
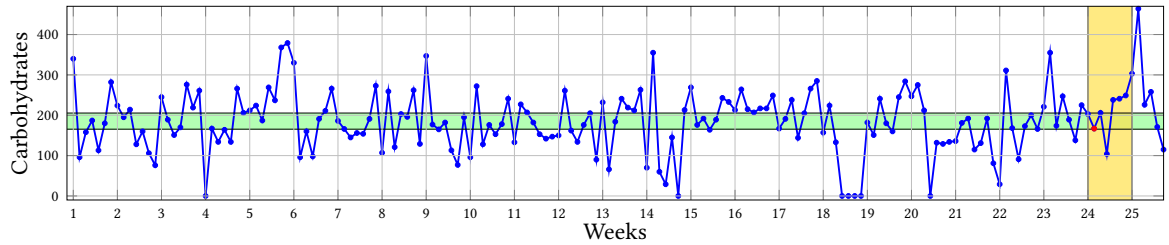

For this summary, the user can clearly see a behavioral pattern that
occurred in the past week. Whenever they had a very low calorie intake
in the past week (the qualifier), their carbohydrate intake was moderate. They can use
this summary to lower their carbohydrate intake if they choose to eat
similar foods as on the day(s) they had a very low calorie intake. These
summaries enable the user to comprehend how well they have performed in
specific aspects (e.g., their calorie intake) within a specified time
window. They can also look at Figures \ref{fig:calorie_sestwq} and 
\ref{fig:carb_sestwq} to verify the multivariate summary. In these charts, the blue range 
represents the ``very low'' range, while the green range represents the ``moderate'' range.
The vertical range represents the data the summary
is describing and the data points that agree with the summary are in red.


\subsubsection{Day-based pattern summaries:}
These summaries focus on patterns in the user's behavior in terms of
certain attributes during ``named'' days of the week (e.g., Mondays).
After receiving the standard evaluation summaries above, our user may
wonder how they typically perform on certain days of the week. It is
possible that they perform better for their health goals on certain
days. The following protoform can be used with \textbf{S} = \{\textit{very
low}, \textit{low}, \textit{moderate}, \textit{high}, \textit{very high}\}:

\begin{description}
\small
\item \textbf{Day-Based Pattern Protoform}: Your $\langle$attribute 1$\rangle$ tends
    to be $\langle$summarizer 1$\rangle$,..., and your $\langle$attribute $n\rangle$ tends to
    be $\langle$summarizer $n\rangle$ on $\langle$specified day$\rangle$.\\~\\
\textbf{Univariate Example}: Your \textbf{calorie intake} tends to be
\textbf{very high} on \textbf{Sundays}.\\
\textbf{Multivariate Example}: Your \textbf{calorie intake} tends to be
\textbf{very high} and your \textbf{carbohydrate intake} tends to be
\textbf{very high} on \textbf{Sundays}.
\end{description}



\begin{figure}[!t]
  \centering
    \pgfplotsset{every axis/.append style = {
            tick label style={font=\tiny}},
            x coord trafo/.code={\pgfmathparse{(#1-1)/7+1}\pgfmathresult},
        x coord inv trafo/.code={\pgfmathparse{#1}\pgfmathresult},
    }
    \centerline{
        \hspace{-0.25in}
        \subfloat[Calorie Intake Data]{
  \label{fig:calorie_db}
            \scalebox{0.9}{
            \begin{tikzpicture}
            \begin{axis}[%
                axis on top,
                width=7in, height=1.75in,
                grid=both,%
                label shift = -5pt,
                xlabel=Weeks, xmin=0, xmax=175, ymin=1000,ymax=4200,%
                xtick distance=1, mark size = 1pt,%
                ylabel=Calories,%
                ytick distance=1000, thick]
                \addplot+[color=blue, mark=*, mark options={fill=blue}] 
                    table[x=Day,y=Calories,col sep=comma]  {calorie_data.csv};
                \addplot[name path=E,black, no markers, line width=0pt]
                    coordinates {(0,2720) (175,2720)};
                \addplot[name path=F,black, no markers, line width=0pt]
                    coordinates {(0,4200) (175,4200)};
                \addplot[gray!30] fill between[of=E and F];                
                \addplot[only marks, color=red] coordinates {
                        (1, 2924.0)
                        (35, 3105.0)
                        (65, 3053.0)
                        (79, 3212.0)
                        (93, 2818.0)
                        (107, 3004.0)
                        (149, 2912.0)
                        (156, 2835.0)
                        (170, 4122.0)
                    };
                \addplot[green!30, no markers, line width=2pt]
                    coordinates {\input{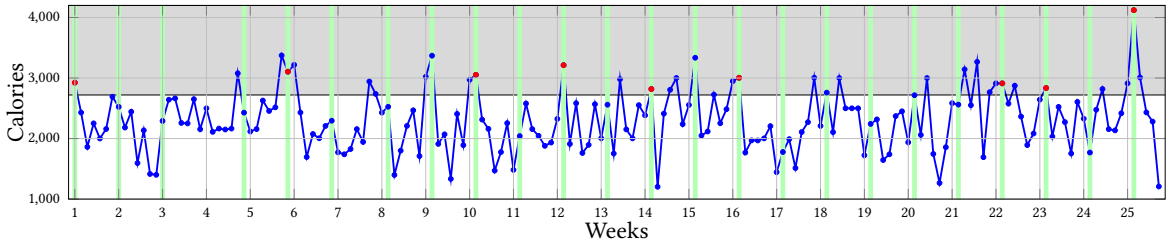} };
            \end{axis}
            \end{tikzpicture}
        }}
    }
    \centerline{
        \hspace{-0.25in}
        \subfloat[Carbohydrate Intake Data]{
  \label{fig:carb_db}
            \scalebox{0.9}{
            \begin{tikzpicture}
            \begin{axis}[name=carb_plot,%
                axis on top,
                width=7in, height=1.75in,
                grid=major, mark size=1pt, %
                label shift = -5pt,
                xlabel=Weeks, xmin=0, xmax=174, ymin=-10, ymax=470,%
                xtick distance=1, ylabel=Carbohydrates,%
                ytick distance=100, thick]
            \addplot+[color=blue, mark=*, mark options={fill=blue}] 
                table[x=Day,y=Carbohydrates,col sep=comma]  {carb_data.csv};
                \addplot[name path=E,black, no markers, line width=0pt]
                    coordinates {(0,254) (175,254)};
                \addplot[name path=F,black, no markers, line width=0pt]
                    coordinates {(0,470) (175,470)};
                \addplot[gray!30] fill between[of=E and F];                
                %
                \addplot[only marks, color=red] coordinates {
                        (1, 340.0)
                        (35, 379.0)
                        (65, 272.0)
                        (79, 261.0)
                        (93, 355.0)
                        (107, 264.0)
                        (149, 311.0)
                        (156, 355.0)
                        (170, 464.0)
                    };
                \addplot[green!30, no markers, line width=2pt]
                    coordinates {\input{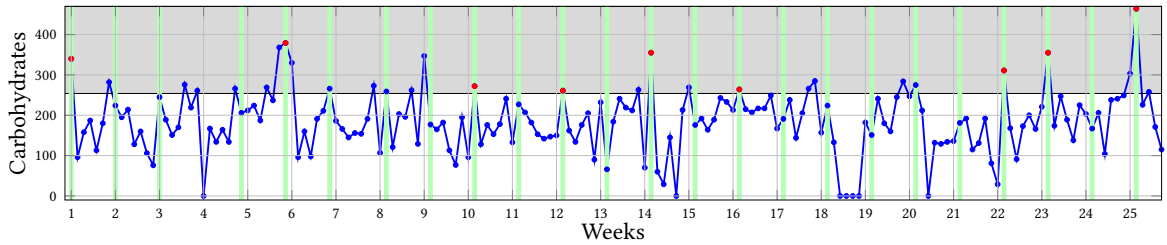}  };
            \end{axis}
            \end{tikzpicture}
    }}}
\vspace{-0.1in}
  \caption{Summary Provenance - Day-Based Pattern. The data
  points supporting the conclusion are in red. Specific days of interest are shown as green line
  segments. The gray region denotes ``very high.''}
\vspace{-0.1in}
\end{figure}

According to the summaries above, the user does not perform well on
Sundays; both calorie and carbohydrate intake are typically very high on
Sundays. Using these conclusions, the
user can monitor how they usually eat on that day and take preventive
action. Figures \ref{fig:calorie_db} and 
\ref{fig:carb_db} illustrate the multivariate summary. In these charts,
the gray range 
represents the ``very high'' range. The vertical green bars represent the day of the week specified.

\subsubsection{Goal assistance summaries:}
Goal evaluation can be added to any summary type, to evaluate a certain
attribute against a goal or a guideline. 
How would a user evaluate their progress towards their goals? If our
user wishes to evaluate how well they limit their carbohydrate and
calorie intake in a specific week, this protoform can be used:
\begin{description}
\small
\item \textbf{Goal Evaluation Protoform}: On $\langle$quantifier$\rangle$
    $\langle$sub-time window$\rangle$ in the past $\langle$time window$\rangle$, you
    $\langle$summarizer 1$\rangle$ your goal to keep your $\langle$attribute 1$\rangle$ $\langle$goal
    1$\rangle$,..., and you $\langle$summarizer 1$\rangle$ your goal to keep your
    $\langle$attribute $n\rangle$ $\langle$goal $n\rangle$.\\~\\
\textbf{Univariate Example}: On \textbf{most of the \textit{days}} in
the past \textbf{week}, you \textbf{did not reach} your goal to keep
your \textbf{calorie intake \textit{low}}.\\
\textbf{Multivariate Example}: On \textbf{some of the}
\textbf{\textit{days}} in the past \textbf{week}, you \textbf{did not
reach} your goal to keep your \textbf{calorie intake \textit{low}} and
you \textbf{reached} your goal to keep your \textbf{carbohydrate intake
\textit{low}}.
\end{description}

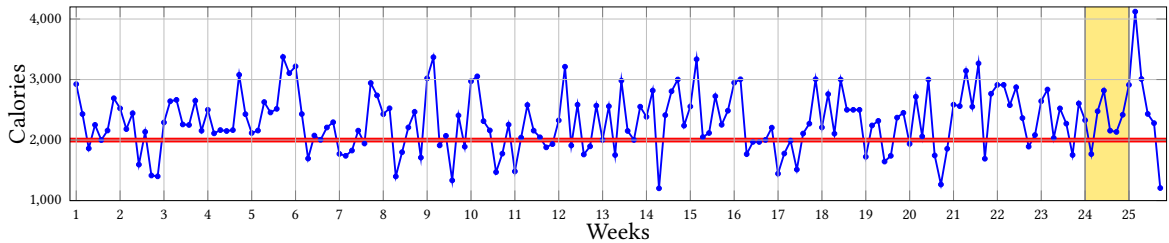
\begin{figure}[!hbt]
  \centering
    \pgfplotsset{every axis/.append style = {
            tick label style={font=\tiny}},
            x coord trafo/.code={\pgfmathparse{(#1-1)/7+1}\pgfmathresult},
        x coord inv trafo/.code={\pgfmathparse{#1}\pgfmathresult},
    }
    \centerline{
        \hspace{-0.25in}
            \scalebox{0.9}{
            \begin{tikzpicture}
            \begin{axis}[%
                axis on top,
                width=7in, height=1.75in,
                grid=both,%
                label shift = -5pt,
                xlabel=Weeks, xmin=0, xmax=175, ymin=1000,ymax=4200,%
                xtick distance=1, mark size = 1pt,%
                ylabel=Calories,%
                ytick distance=1000, thick]
                \addplot+[color=blue, mark=*, mark options={fill=blue}] 
                    table[x=Day,y=Calories,col sep=comma]  {calorie_data.csv};
                \addplot[red, no markers, line width=2pt]
                    coordinates {(0,2000) (175,2000)};
                \addplot[name path=vA,black, no markers, line width=0pt]
                    coordinates {(162,0) (162,4200)};
                \addplot[name path=vB,black, no markers, line width=0pt]
                    coordinates {(169,0) (169,4200)};
                \addplot[gold!70, fill opacity=0.7] fill between[of=vA and vB];                
            \end{axis}
            \end{tikzpicture}
        }}
\vspace{-0.1in}
  \caption{Summary Provenance - Goal Evaluation/Assistance.  
  The red line denotes the goal of 2000 calories, and the vertical
  region in gold denotes week of interest.}
  \label{fig:ge}
\vspace{-0.1in}
\end{figure}

This protoform is similar to the one used for the standard evaluation
summary at the sub-time window granularity, but with the summarizer set
\textbf{S} = \{\textit{reached}, \textit{did not reach}\}. These
summaries can be used to realize that they fail to reach their calorie intake goal.
On the bright side, they have some days where they reach their
carbohydrate intake goals. These goals can be extracted from official
health guidelines such as the ADA Lifestyle guidelines or suggested by
health physicians \cite{ada}. Figure \ref{fig:ge} 
illustrates the univariate summary. In this chart, the horizontal red line 
represents the calorie intake goal and the vertical range represents the data the summary
is describing.

In addition to goal evaluation summaries, we also allow for goal
assistance summaries that not only evaluate a user's progress towards a
goal, but are also constructed to assist the user if they seem to be
struggling. These summaries evaluate the user's data for multiple
attributes against guidelines that are less defined, such as certain
diets. 
The system must also determine which attributes to mention in the final
summary without making the summary too lengthy. Goal assistance
summaries can be thought of as a combination of goal evaluation
summaries. This time, the set of summarizers is \textbf{S} = \{\textit{increase},
\textit{decrease}\}. If the user wishes to receive a more direct summary of what
they should be working on, we can use this protoform:
\begin{description}
\small
\item \textbf{Goal Assistance Protoform}: In order to better follow the
    $\langle$goal$\rangle$, you should $\langle$summarizer 1$\rangle$ your $\langle$attribute 1$\rangle$,
    $\langle$summarizer 2$\rangle$ your $\langle$attribute 2$\rangle$, ..., and $\langle$summarizer
    $n\rangle$ your $\langle$attribute $n\rangle$.\\~\\
\textbf{Univariate Example}: In order to better follow the
\textbf{2000-calorie diet}, you should \textbf{decrease} your
\textbf{calorie intake}.\\
\textbf{Multivariate Example}: In order to better follow the
\textbf{2000-calorie diet}, you should \textbf{decrease} your
\textbf{calorie intake} and \textbf{increase} your \textbf{carbohydrate
intake.}
\end{description}

Looking at the above summary, the user may actually want to increase
their carbohydrate intake while lowering their calorie intake based on
how they performed last week. This is an expected output as the
\textit{2000-calorie} diet recommends a higher amount of carbohydrates while the
user wishes to limit their carbohydrate intake. It may be best for the
user to switch instead to a \textit{low-carbohydrate eating plan}.
The provenance illustration for the univariate summary is the same as in 
Figure \ref{fig:ge}.

\subsubsection{General if-then pattern summaries:}
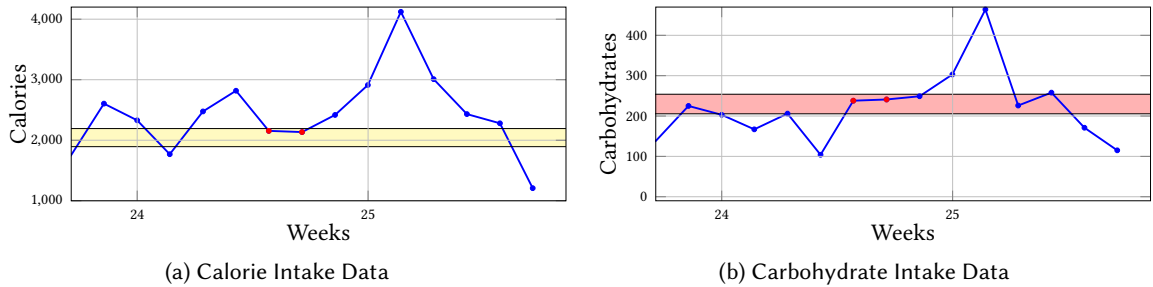
\begin{figure}[!hb]
  \centering
    \pgfplotsset{every axis/.append style = {
            tick label style={font=\tiny}},
            x coord trafo/.code={\pgfmathparse{(#1-1)/7+1}\pgfmathresult},
        x coord inv trafo/.code={\pgfmathparse{#1}\pgfmathresult},
    }
    \centerline{
        \hspace{-0.25in}
        \subfloat[Calorie Intake Data]{
  \label{fig:calorie_git}
            \scalebox{0.9}{
            \begin{tikzpicture}
            \begin{axis}[%
                axis on top,
                width=3.5in, height=1.75in,
                grid=both,%
                label shift = -5pt,
                xlabel=Weeks, xmin=160, xmax=175, ymin=1000,ymax=4200,%
                xtick distance=1, mark size = 1pt,%
                ylabel=Calories,%
                ytick distance=1000, thick]
                \addplot+[color=blue, mark=*, mark options={fill=blue}] 
                    table[x=Day,y=Calories,col sep=comma]  {calorie_data.csv};
                \addplot[name path=B,black, no markers, line width=0pt]
                    coordinates {(0,1894) (175,1894)};
                \addplot[name path=C,black, no markers, line width=0pt]
                    coordinates {(0,2193) (175,2193)};
                \addplot[yellow!30] fill between[of=B and C];                
                \addplot[only marks, color=red] coordinates {
                     (166, 2155.0)
                     (167, 2135.0)
                    };
            \end{axis}
            \end{tikzpicture}
        }}
        \subfloat[Carbohydrate Intake Data]{
      \label{fig:carb_git}
            \scalebox{0.9}{
            \begin{tikzpicture}
            \begin{axis}[name=carb_plot,%
                axis on top,
                width=3.5in, height=1.75in,
                grid=major, mark size=1pt, %
                label shift = -5pt,
                xlabel=Weeks, xmin=160, xmax=175, ymin=-10, ymax=470,%
                xtick distance=1, ylabel=Carbohydrates,%
                ytick distance=100, thick]
                \addplot+[color=blue, mark=*, mark options={fill=blue}] 
                    table[x={Day},y={Carbohydrates},col sep=comma]  {carb_data.csv};
                \addplot[name path=D,black, no markers, line width=0pt]
                    coordinates {(0,205.5) (175,205.5)};
                \addplot[name path=E,black, no markers, line width=0pt]
                    coordinates {(0,254) (175,254)};
                \addplot[red!30] fill between[of=D and E];                
                %
                \addplot[only marks, color=red] coordinates {
                         (166, 238)
                         (167, 241)
                    };
            \end{axis}
            \end{tikzpicture}
    }}}
\vspace{-0.1in}
  \caption{Summary Provenance - General If-Then Pattern. Yellow region
  denotes ``low,'' red denotes ``high.''}
\vspace{-0.1in}
\end{figure}

These summaries find possible correlations between multiple variables
pertaining to a user's behavior over the entire time window. What if the
user wishes to find a possible correlation between a certain behavior
and an inhibiting action they take?
The following protoform generates a summary that describes this
correlation:
\begin{description}
\small
\item \textbf{General If-Then Pattern Protoform}: In general, if your
    $\langle$attribute 1$\rangle$ is $\langle$summarizer 1$\rangle$,..., and your $\langle$attribute
    $n\rangle$ is $\langle$summarizer $n\rangle$, then your $\langle$attribute $n+1\rangle$ is
    $\langle$summarizer $n+1\rangle$,..., and your $\langle$attribute $n+m\rangle$ is
    $\langle$summarizer $n+m\rangle$\\~\\
\textbf{Example}: In general, if your \textbf{calorie intake} is
\textbf{low},
then your \textbf{carbohydrate intake} is \textbf{high}.
\end{description}
These summaries have summarizer set \textbf{S} = \{\textit{very low},
\textit{low}, \textit{moderate}, \textit{high}, \textit{very high}\}.
Figures \ref{fig:calorie_git} and 
\ref{fig:carb_git} verify the multivariate summary, focusing on the last
two weeks of data. The yellow range 
represents the ``low'' range and the red range 
represents the ``high'' range. The data points that agree with the summary are in red.



\subsection{Consecutive Time Window Summaries}
\begin{figure}[!t]
  \centering
  \includegraphics[width=0.25\textwidth, height=1.5in]{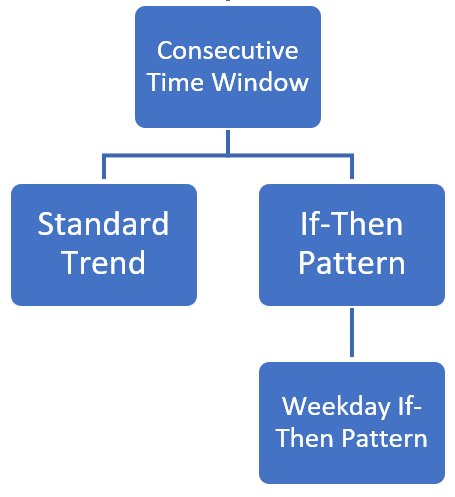}
\vspace{-0.1in}
  \caption{Consecutive Time Window Summary Types}
  \label{fig:consecutive_tw}
\vspace{-0.1in}
\end{figure}

After searching within specific time windows to find behavioral
patterns, our framework allows the user to move on to comparisons
between time windows. Naturally, the user would start with
consecutive time windows, or time windows that are next to each other.
With TW set to a weekly granularity and sTW set to a
daily granularity, our user will find patterns between consecutive weeks
and consecutive days. The summaries below are referred to as consecutive
time window summaries, and their inter-relationships are shown in Figure
\ref{fig:consecutive_tw}.

\subsubsection{Standard trend summaries:}

Suppose the user wishes to know how they perform from day to day.
Looking at the data, how often does their calorie intake increase or
decrease between days? We can use a standard trend summary to see this.

Standard trend summaries describe trends from one sub-time window to the
next. These summaries can be used to describe a user's tendency between
two consecutive sub-time windows and use summarizer set \textbf{S} =
\{\textit{increased}, \textit{decreased}, \textit{stayed the same}\}.
\begin{description}
\small
\item \textbf{Standard Trend Protoform}: $\langle$Quantifier$\rangle$ time, your
    $\langle$attribute 1$\rangle$ $\langle$summarizer 1$\rangle$,..., and your $\langle$attribute
    $n\rangle$ $\langle$summarizer $n\rangle$ from one $\langle$sub-time window$\rangle$ to the
    next.\\~\\
\textbf{Univariate Example}: \textbf{Half of the} time, your
\textbf{calorie intake \textit{increases}} from one \textbf{day} to the
next.\\
\textbf{Multivariate Example}: \textbf{Some of the }time, your
\textbf{calorie intake} \textbf{\textit{increases}} and your
\textbf{carbohydrate intake} \textbf{\textit{increases}} from one
\textbf{day} to the next.
\end{description}



\begin{figure}[!htb]
  \centering
    \pgfplotsset{every axis/.append style = {
            tick label style={font=\tiny}},
            x coord trafo/.code={\pgfmathparse{(#1-1)/7+1}\pgfmathresult},
        x coord inv trafo/.code={\pgfmathparse{#1}\pgfmathresult},
    }
    \centerline{
        \hspace{-0.25in}
        \subfloat[Calorie Intake Data]{
  \label{fig:calorie_st}
            \scalebox{0.9}{
            \begin{tikzpicture}
            \begin{axis}[%
                axis on top,
                width=7in, height=1.75in,
                grid=both,%
                label shift = -5pt,
                xlabel=Weeks, xmin=0, xmax=175, ymin=1000,ymax=4200,%
                xtick distance=1, mark size = 1pt,%
                ylabel=Calories,%
                ytick distance=1000, thick]
                \addplot+[color=blue, mark=*, mark options={fill=blue}] 
                    table[x=Day,y=Calories,col sep=comma]  {calorie_data.csv};
                \addplot[color=red, mark=o, line width=2pt] coordinates {
                        \input{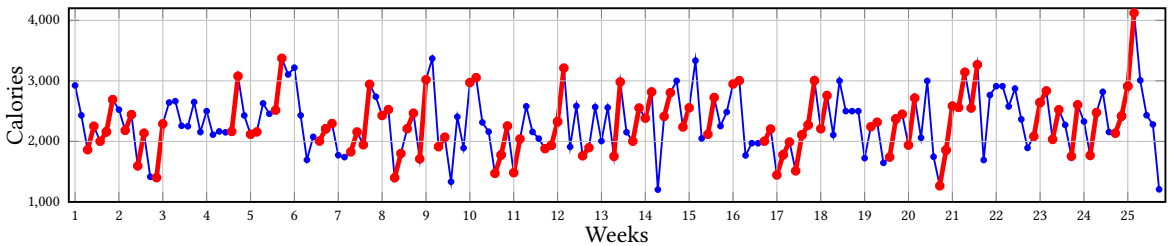} 
                    };
            \end{axis}
            \end{tikzpicture}
        }}}
        \centerline{
        \hspace{-0.25in}
        \subfloat[Carbohydrate Intake Data]{
          \label{fig:carb_st}
            \scalebox{0.9}{
            \begin{tikzpicture}
            \begin{axis}[name=carb_plot,%
                axis on top,
                width=7in, height=1.75in,
                grid=major, mark size=1pt, %
                label shift = -5pt,
                xlabel=Weeks, xmin=0, xmax=175, ymin=-10, ymax=470,%
                xtick distance=1, ylabel=Carbohydrates,%
                ytick distance=100, thick]
                \addplot+[color=blue, mark=*, mark options={fill=blue}] 
                    table[x={Day},y={Carbohydrates},col sep=comma]  {carb_data.csv};
                %
                \addplot[color=red, mark=o, line width=2pt] coordinates {
                        \input{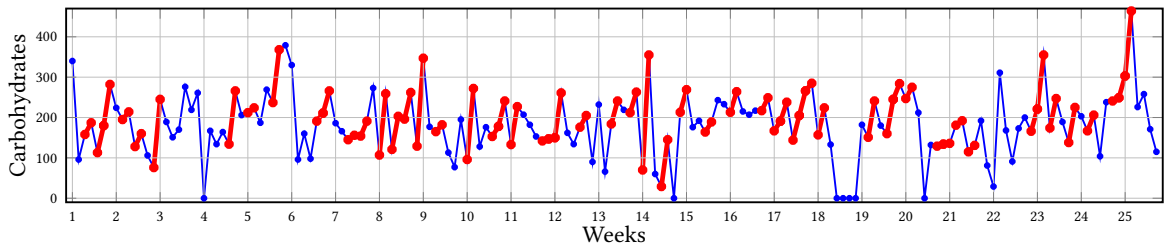} 
                    };
            \end{axis}
            \end{tikzpicture}
    }}}
\vspace{-0.1in}
  \caption{Summary Provenance - Standard Trend. Red line segments
  illustrate the pattern.}
\vspace{-0.1in}
\end{figure}

These two summaries allow our user to know that there is around a 50\%
chance that their calorie intake will increase the next day and that
there is a relatively smaller chance that their calorie intake and their carbohydrate
intake will both increase the next day. While similar to standard
evaluation summaries (which evaluate the attribute on each day), here we
evaluate the attribute between one day and the next; these summaries are
ratio-based and span the entire dataset, instead of a specified time
window. Figures \ref{fig:calorie_st} and 
\ref{fig:carb_st} verify the multivariate summary. In these charts, the red line segments
indicate where the pattern is found.

\subsubsection{If-then pattern summaries:}
What if the user wishes to know more about how their past and current
behaviors predict the trends in the near future? For answering this, we
propose if-then pattern summaries that provide more interesting patterns based on
frequent sequence mining \cite{spade}. These patterns span multiple
consecutive sub-time windows and are of variable length,  constrained by
the size of the time window. They use summarizer set \textbf{S} = \{\textit{very
low}, \textit{low}, \textit{moderate}, \textit{high},\textit{ very high}\}. The protoform is:
\begin{description}
\small
\item \textbf{If-Then Pattern Protoform}: There is $\langle$confidence
    value$\rangle$ confidence that, when your $\langle$attribute 1$\rangle$ is
    $\langle$summarizer 1:1$\rangle$, then $\langle$summarizer 2:1$\rangle$,..., then
    $\langle$summarizer $m$:1$\rangle$,..., and your $\langle$attribute $n\rangle$ is
    $\langle$summarizer 1:$n\rangle$, then $\langle$summarizer 2:$n\rangle$,..., then
    $\langle$summarizer $m$:$n\rangle$, your $\langle$attribute 1$\rangle$ tends to be
    $\langle$summarizer $(m+1)$:1$\rangle$,..., and your $\langle$attribute $n\rangle$ tends
    to be $\langle$summarizer $(m+1)$:$n\rangle$ the next $\langle$time window$\rangle$.\\~\\
\textbf{Univariate Example}: There is \textbf{100\%} confidence that,
when your \textbf{calorie intake} follows the pattern of being
\textbf{moderate}, your \textbf{calorie intake} tends to be \textbf{very
low} the next \textbf{day}.\\
\textbf{Multivariate Example}: There is \textbf{100\%} confidence that,
when your \textbf{calorie intake} follows the pattern of being
\textbf{very high}, your \textbf{calorie intake} tends to be \textbf{very high} and your \textbf{carbohydrate intake}
tends to be \textbf{very high} the next \textbf{day}.
\end{description}



\begin{figure}[!ht]
  \centering
    \pgfplotsset{every axis/.append style = {
            tick label style={font=\tiny}},
            x coord trafo/.code={\pgfmathparse{(#1-1)/7+1}\pgfmathresult},
        x coord inv trafo/.code={\pgfmathparse{#1}\pgfmathresult},
    }
    \centerline{
        \hspace{-0.25in}
        \subfloat[Calorie Intake Data]{
  \label{fig:calorie_it}
            \scalebox{0.9}{
            \begin{tikzpicture}
            \begin{axis}[%
                axis on top,
                width=7in, height=1.75in,
                grid=both,%
                label shift = -5pt,
                xlabel=Weeks, xmin=0, xmax=175, ymin=1000,ymax=4200,%
                xtick distance=1, mark size = 1pt,%
                ylabel=Calories,%
                ytick distance=1000, thick]
                \addplot+[color=blue, mark=*, mark options={fill=blue}] 
                    table[x=Day,y=Calories,col sep=comma]  {calorie_data.csv};
                \addplot[name path=E,black, no markers, line width=0pt]
                    coordinates {(0,2720) (175,2720)};
                \addplot[name path=F,black, no markers, line width=0pt]
                    coordinates {(0,4200) (175,4200)};
                \addplot[gray!30] fill between[of=E and F];                
                \addplot[color=red, mark=o, line width=2pt] coordinates {
                        \input{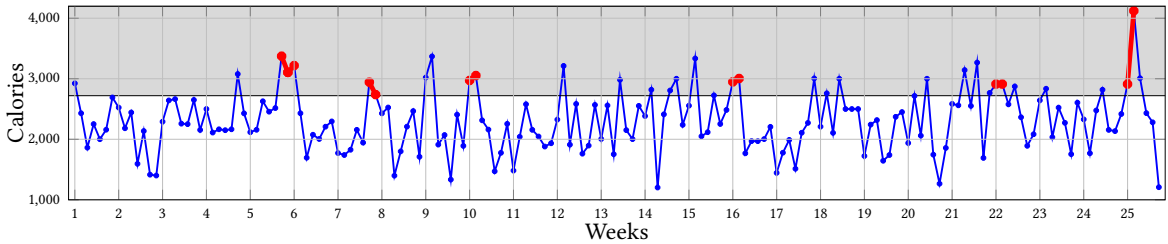}
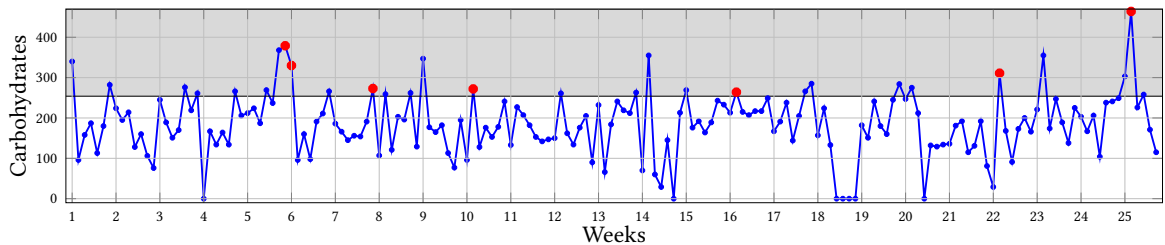 
                    };
            \end{axis}
            \end{tikzpicture}
        }}}
        \centerline{
        \hspace{-0.25in}
        \subfloat[Carbohydrate Intake Data]{
  \label{fig:carb_it}
            \scalebox{0.9}{
            \begin{tikzpicture}
            \begin{axis}[name=carb_plot,%
                axis on top,
                width=7in, height=1.75in,
                grid=major, mark size=1pt, %
                label shift = -5pt,
                xlabel=Weeks, xmin=0, xmax=175, ymin=-10, ymax=470,%
                xtick distance=1, ylabel=Carbohydrates,%
                ytick distance=100, thick]
                \addplot+[color=blue, mark=*, mark options={fill=blue}] 
                    table[x={Day},y={Carbohydrates},col sep=comma]  {carb_data.csv};
                \addplot[name path=E,black, no markers, line width=0pt]
                    coordinates {(0,254) (175,254)};
                \addplot[name path=F,black, no markers, line width=0pt]
                    coordinates {(0,470) (175,470)};
                \addplot[gray!30] fill between[of=E and F];                
                %
                \addplot[color=red, line width=2pt, only marks] coordinates {
                        (35, 379.0) 
                        (36, 330.0)
                        (49, 273.0)
                        (65, 272.0)
                        (107, 264.0)
                        (149, 311.0)
                        (170, 464.0)
                    };
            \end{axis}
            \end{tikzpicture}
    }}}
\vspace{-0.1in}
  \caption{Summary Provenance - If-Then Pattern. Gray region denotes
  ``very high.'' Red points denote points supporting the pattern.}
\vspace{-0.1in}
\end{figure}

In the above protoform, \textit{m} represents the number of summarizers
per attribute and \textit{n} represents the total number of attributes.
From these summaries, the user can conclude that their calorie intake is
typically very low after having a moderate intake the previous day. 
On the other hand, if they have a very high calorie intake, then both
calorie and carbohydrate intake remains very high the next day as well.
Figures \ref{fig:calorie_it} and 
\ref{fig:carb_it} verify the multivariate summary. In these charts, the
gray range 
represents the ``very high'' range.
The data points that agree with the summary are in red.

How about if the user wants to see these behavioral patterns pertaining
to days of the week? If-then pattern summaries can also be made
dependent on the day of the week, via the protoform:
\begin{description}
\small
\item \textbf{Day If-Then Pattern Protoform}: There is $\langle$confidence
    value$\rangle$ confidence that, when your $\langle$attribute 1$\rangle$ is
    $\langle$summarizer 1:1$\rangle$ on a $\langle$day 1:1$\rangle$, then $\langle$summarizer 2:1$\rangle$
    on a $\langle$day 2:1$\rangle$,..., then $\langle$summarizer $m$:1$\rangle$ on a $\langle$day
    $m$:1$\rangle$,..., and your $\langle$attribute $n\rangle$ is $\langle$summarizer 1:$n\rangle$
    on a $\langle$day 1:$n\rangle$, then $\langle$summarizer 2:$n\rangle$ on a $\langle$day
    2:$n\rangle$,..., then $\langle$summarizer $m$:$n\rangle$ on a $\langle$day $m$:$n\rangle$,
    your $\langle$attribute 1$\rangle$ tends to be $\langle$summarizer $(m+1)$:1$\rangle$ the
    next $\langle$day $(m+1)$:1$\rangle$,..., and your $\langle$attribute $n\rangle$ tends to
    be $\langle$summarizer $(m+1)$:$n\rangle$ the next $\langle$day $(m+1)$:$n\rangle$.\\~\\
\textbf{Univariate Example}: There is \textbf{100\%} confidence that,
when your \textbf{calorie intake} follows the pattern of being
\textbf{very high }on a \textbf{Saturday}, your \textbf{calorie intake}
tends to be \textbf{very high} the next \textbf{Sunday}.\\
\textbf{Multivariate Example}: There is \textbf{100\%} confidence that,
when your \textbf{calorie intake} follows the pattern of being
\textbf{very high} on a \textbf{Saturday}, your \textbf{calorie intake} tends to be
\textbf{very high} the next \textbf{Sunday} and your \textbf{carbohydrate intake} tends to be \textbf{very high} the next \textbf{Sunday}.
\end{description}



\begin{figure}[!t]
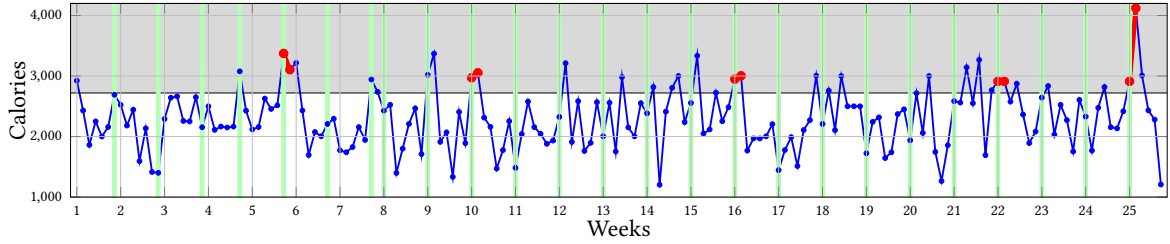
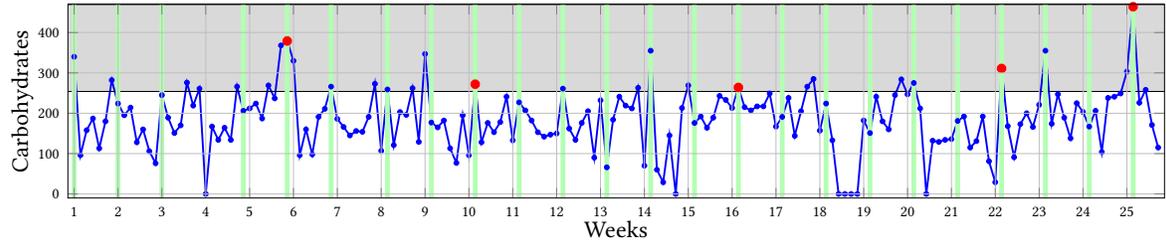

  \centering
    \pgfplotsset{every axis/.append style = {
            tick label style={font=\tiny}},
            x coord trafo/.code={\pgfmathparse{(#1-1)/7+1}\pgfmathresult},
        x coord inv trafo/.code={\pgfmathparse{#1}\pgfmathresult},
    }
    \centerline{
        \hspace{-0.25in}
        \subfloat[Calorie Intake Data]{
  \label{fig:calorie_wit}
            \scalebox{0.9}{
            \begin{tikzpicture}
            \begin{axis}[%
                axis on top,
                width=7in, height=1.75in,
                grid=both,%
                label shift = -5pt,
                xlabel=Weeks, xmin=0, xmax=175, ymin=1000,ymax=4200,%
                xtick distance=1, mark size = 1pt,%
                ylabel=Calories,%
                ytick distance=1000, thick]
                \addplot+[color=blue, mark=*, mark options={fill=blue}] 
                    table[x=Day,y=Calories,col sep=comma]  {calorie_data.csv};
                \addplot[name path=E,black, no markers, line width=0pt]
                    coordinates {(0,2720) (175,2720)};
                \addplot[name path=F,black, no markers, line width=0pt]
                    coordinates {(0,4200) (175,4200)};
                \addplot[gray!30] fill between[of=E and F];                
                \addplot[color=red, mark=o, line width=2pt] coordinates {
                    \input{calorie_wit.dat} };
                \addplot[green!30, no markers, line width=2pt]
coordinates {\input{calorie_wit2.dat} 
};
            \end{axis}
            \end{tikzpicture}
        }}}
        \centerline{
        \hspace{-0.25in}
        \subfloat[Carbohydrate Intake Data]{
  \label{fig:carb_wit}
            \scalebox{0.9}{
            \begin{tikzpicture}
            \begin{axis}[name=carb_plot,%
                axis on top,
                width=7in, height=1.75in,
                grid=major, mark size=1pt, %
                label shift = -5pt,
                xlabel=Weeks, xmin=0, xmax=175, ymin=-10, ymax=470,%
                xtick distance=1, ylabel=Carbohydrates,%
                ytick distance=100, thick]
                \addplot+[color=blue, mark=*, mark options={fill=blue}] 
                    table[x={Day},y={Carbohydrates},col sep=comma]  {carb_data.csv};
                \addplot[name path=E,black, no markers, line width=0pt]
                    coordinates {(0,254) (175,254)};
                \addplot[name path=F,black, no markers, line width=0pt]
                    coordinates {(0,470) (175,470)};
                \addplot[gray!30] fill between[of=E and F];                
                %
                \addplot[color=red, line width=2pt, only marks] coordinates {
                        (35, 379.0) 
                        (65, 272.0)
                        (107, 264.0)
                        (149, 311.0)
                        (170, 464.0)
                    };
                \addplot[green!30, no markers, line width=2pt]
coordinates {\input{carb_wit.dat} 
};
            \end{axis}
            \end{tikzpicture}
    }}}
\vspace{-0.1in}
  \caption{Summary Provenance - Day If-Then Pattern. Red points support
  the summary, and green line segments denote the days of interest.}
\vspace{-0.1in}
\end{figure}

In the above protoform, \textit{m} represents the number of
summarizer-day pairs per attribute and \textit{n} represents the total
number of attributes. 
The user can observe that 
if they have Saturdays with
very high calorie intake, they typically consume a lot of calories and carbohydrates
on Sunday. This can allow the user to make changes in their weekend
diet. Figures \ref{fig:calorie_wit} and 
    \ref{fig:carb_wit} verify the multivariate summary. In these charts,
    the gray range 
represents the ``very high'' range.
The data points that agree with the summary are in red and the vertical green bars show the specified day of the week.

\begin{figure}[!t]
  \centering
  \includegraphics[width=0.25\textwidth, height=1.5in]{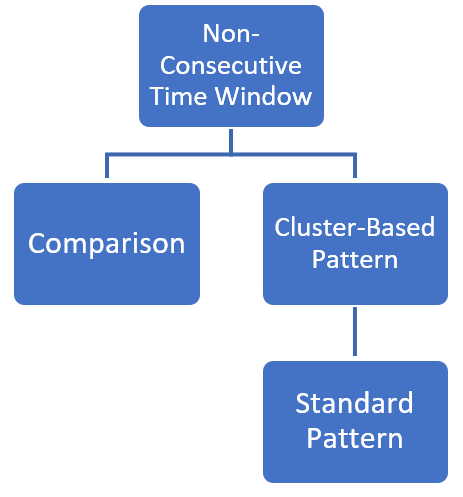}
\vspace{-0.1in}
  \caption{Non-Consecutive Time Window Summary Types}
  \label{fig:nonconsecutive_tw}
\vspace{-0.1in}
\end{figure}

\subsection{Non-Consecutive Time Window Summaries}
Having examined the patterns that can be found between consecutive time
windows, the user may try to find patterns across time windows that are
not consecutive. Perhaps the past week they had was similar to another
week that occurred in an earlier month. Summaries explaining
these types of patterns are called non-consecutive time window
summaries. 
These summaries look at time windows that do not necessarily have to be
consecutive; they compare discovered trends found in one time window
with those of another time window in the data. 

\subsubsection{Comparison summaries:}
What if the user wants to make comparisons between their most recent
week of logging and a week in a much earlier part of their data?
Comparison summaries provide comparisons between any two different time
windows to help users evaluate their behavioral differences. These
summaries use summarizer set \textbf{S} = \{\textit{higher}, \textit{lower}, \textit{about the
same}\}. The protoform is:
\begin{description}
\small
\item \textbf{Comparison Protoform}: Your $\langle$attribute 1$\rangle$ was
    $\langle$summarizer 1$\rangle$,..., and your $\langle$attribute $n\rangle$ was
    $\langle$summarizer $n\rangle$ on $\langle$time window 1$\rangle$ $\langle$number 1$\rangle$ than they
    were on $\langle$time window 2$\rangle$ $\langle$number 2$\rangle$.\\~\\
\textbf{Univariate Example}: Your \textbf{calorie intake} was
\textbf{about the same} in \textbf{week \textit{24}} than it was in \textbf{week
\textit{12}}.\\ 
\textbf{Multivariate Example}: Your \textbf{calorie intake} was
\textbf{about the same} and your \textbf{carbohydrate intake} was
\textbf{about the same} in \textbf{week \textit{24}} than they were in
\textbf{week \textit{12}}.
\end{description}



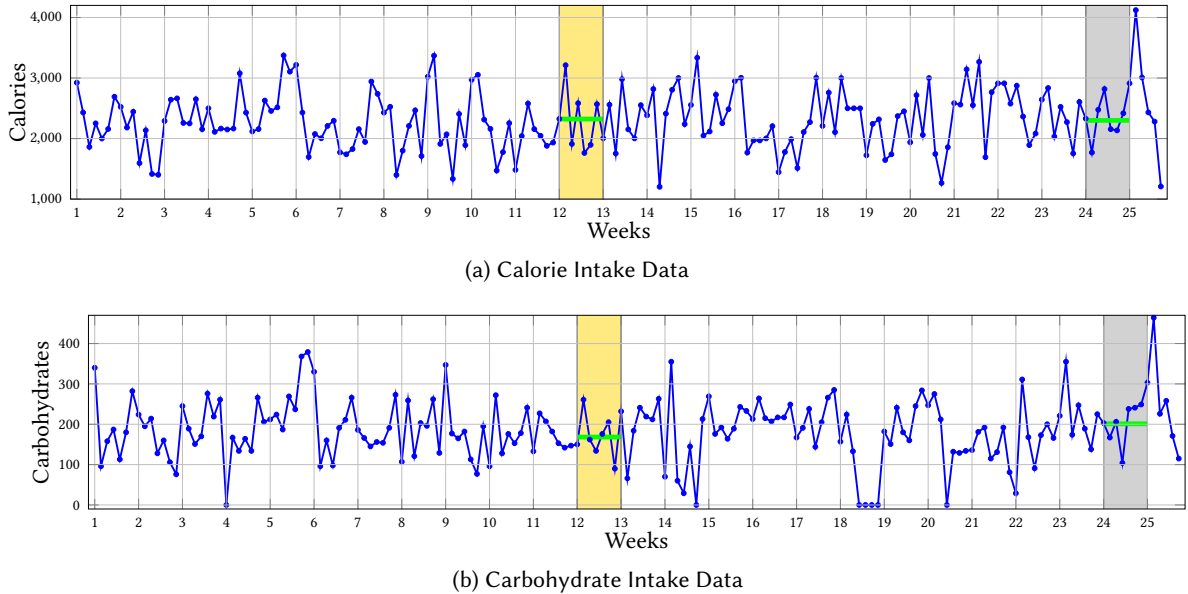
\begin{figure}[!ht]
  \centering
    \pgfplotsset{every axis/.append style = {
            tick label style={font=\tiny}},
            x coord trafo/.code={\pgfmathparse{(#1-1)/7+1}\pgfmathresult},
        x coord inv trafo/.code={\pgfmathparse{#1}\pgfmathresult},
    }
    \centerline{
        \hspace{-0.25in}
        \subfloat[Calorie Intake Data]{
  \label{fig:calorie_ec}
            \scalebox{0.9}{
            \begin{tikzpicture}
            \begin{axis}[%
                axis on top,
                width=7in, height=1.75in,
                grid=both,%
                label shift = -5pt,
                xlabel=Weeks, xmin=0, xmax=175, ymin=1000,ymax=4200,%
                xtick distance=1, mark size = 1pt,%
                ylabel=Calories,%
                ytick distance=1000, thick]
                \addplot+[color=blue, mark=*, mark options={fill=blue}] 
                    table[x=Day,y=Calories,col sep=comma]  {calorie_data.csv};
                \addplot[green, no markers, line width=2pt]
                    coordinates {(78,2322.6) (85,2322.6)};
                \addplot[green, no markers, line width=2pt]
                    coordinates {(162,2300) (169,2300)};
                \addplot[name path=vA,black, no markers, line width=0pt]
                    coordinates {(78,0) (78,4200)};
                \addplot[name path=vB,black, no markers, line width=0pt]
                    coordinates {(85,0) (85,4200)};
                \addplot[gold!70,fill opacity=0.7] fill between[of=vA and vB];                
                \addplot[name path=vC,black, no markers, line width=0pt]
                    coordinates {(162,0) (162,4200)};
                \addplot[name path=vD,black, no markers, line width=0pt]
                    coordinates {(169,0) (169,4200)};
                \addplot[gray!70, fill opacity=0.5] fill between[of=vC and vD];                
            \end{axis}
            \end{tikzpicture}
}}}
\centerline{
        \subfloat[Carbohydrate Intake Data]{
  \label{fig:carb_ec}
            \scalebox{0.9}{
            \begin{tikzpicture}
            \begin{axis}[name=carb_plot,%
                axis on top,
                width=7in, height=1.75in,
                grid=major, mark size=1pt, %
                label shift = -5pt,
                xlabel=Weeks, xmin=0, xmax=175, ymin=-10, ymax=470,%
                xtick distance=1, ylabel=Carbohydrates,%
                ytick distance=100, thick]
                \addplot+[color=blue, mark=*, mark options={fill=blue}] 
                    table[x={Day},y={Carbohydrates},col sep=comma]  {carb_data.csv};
                %
                \addplot[green, no markers, line width=2pt]
                    coordinates {(78,168.3) (85,168.3)};
                \addplot[green, no markers, line width=2pt]
                    coordinates {(162,201.1) (169,201.1)};
                \addplot[name path=vA,black, no markers, line width=0pt]
                    coordinates {(78,-10) (78,470)};
                \addplot[name path=vB,black, no markers, line width=0pt]
                    coordinates {(85,-10) (85,470)};
                \addplot[gold!70, fill opacity=0.7] fill between[of=vA and vB];                
                \addplot[name path=vC,black, no markers, line width=0pt]
                    coordinates {(162,-10) (162,470)};
                \addplot[name path=vD,black, no markers, line width=0pt]
                    coordinates {(169,-10) (169,470)};
                \addplot[gray!70, fill opacity=0.5] fill between[of=vC and vD];                
            \end{axis}
            \end{tikzpicture}
    }}}
\vspace{-0.1in}
  \caption{Summary Provenance - Comparison. The gray vertical region is the past
  week, and the gold region is the week being compared against. Green line
  segments denote the week-based summarizer, which is ``moderate.''}
\vspace{-0.1in}
\end{figure}

Looking at the past week and another week earlier in the data, the user
can see that their intakes of calories and carbohydrates of this past
week were about the same as they were three months before. 
Figures \ref{fig:calorie_ec} and 
\ref{fig:carb_ec} verify the multivariate summary. In these charts, the
gray vertical range represents the past week, while the golden vertical range represents the week it is compared against.

Comparison summaries can also be enhanced with a goal using summarizer
set \textbf{S} = \{\textit{better}, \textit{not do as well}, \textit{about the same}\}. We display
the protoform below:
\begin{description}
\small
\item \textbf{Goal Comparison Protoform}: You did $\langle$summarizer 1$\rangle$
    overall with keeping your $\langle$attribute 1$\rangle$ $\langle$goal 1$\rangle$ ,..., and
    you did $\langle$summarizer $n\rangle$ overall with keeping your $\langle$attribute
    $n\rangle$ $\langle$goal $n\rangle$ in $\langle$time window 1$\rangle$ $\langle$number 1$\rangle$ than you did in $\langle$time window 2$\rangle$ $\langle$number 2$\rangle$.\\
    
\textbf{Univariate Example}: You did \textbf{about the same} overall with keeping your \textbf{calorie intake \textit{low}} in \textbf{week \textit{24}} than you did in \textbf{week \textit{12}}.

\textbf{Multivariate Example}: You did \textbf{about the same} overall with keeping your \textbf{calorie intake \textit{low}} and you did \textbf{about the same} overall with keeping your \textbf{carbohydrate intake \textit{low}} in \textbf{week \textit{24}} than you did in \textbf{week \textit{12}}.
\end{description}



Figures \ref{fig:calorie_ec} illustrates the multivariate summary.
In these charts, the gray vertical range represents the past week,
while the yellow vertical range represents the week it is compared
against. The green segments denote the average nutrient level for that
week.

\subsubsection{Cluster-based pattern summaries:}
The user may also want to predict how they will act the following week
based on their behavior in the past week. One method to achieve this
would be to find other weeks most similar to this past one and to see
what happened in the weeks that followed them. We use cluster-based pattern
summaries to display these patterns. 

These summaries factor in all of the other time windows that are similar
to the time window in question, resulting in a cluster. For example, if
we are looking at the current week, our system will factor in every
other week that has a similar representation (using the Squeezer
\cite{squeezer} clustering algorithm). 
These summaries use
summarizer set \textbf{S} = \{\textit{rose}, \textit{dropped}, \textit{stayed the same}\}.
In addition to a protoform, we
also add a description of the preceding week. 
\begin{description}
\small
\item \textbf{Preceding Time Window Description Protoform}: In $\langle$time
    window$\rangle$ $\langle$week number$\rangle$, your $\langle$attribute 1$\rangle$ was
    $\langle$summarizer 1:1$\rangle$, then $\langle$summarizer 2:1$\rangle$,..., then
    $\langle$summarizer $m_1$:1$\rangle$,..., and your $\langle$attribute $n\rangle$ was
    $\langle$summarizer $n$:1$\rangle$, then $\langle$summarizer $n$:2$\rangle$,..., then
    $\langle$summarizer $m_n$:$n\rangle$.
\item \textbf{Cluster-Based Pattern Protoform}: During $\langle$quantifier$\rangle$
    $\langle$time window (plural)$\rangle$ similar to $\langle$time window$\rangle$ $\langle$week
    number$\rangle$, your $\langle$attribute 1$\rangle$ $\langle$summarizer 1$\rangle$,..., and your
    $\langle$attribute $n\rangle$ $\langle$summarizer $n\rangle$ the next $\langle$time
    window$\rangle$.\\~\\
\textbf{Univariate Example}: In \textbf{week \textit{24}}, your
\textbf{calorie intake} was \textbf{moderate}, then \textbf{very low},
then \textbf{high}, then \textbf{very high}, then \textbf{low}, then
\textbf{moderate}. During \textbf{more than half of the \textit{weeks}} similar to
\textbf{week \textit{24}}, your \textbf{calorie intake \textit{dropped}}
the next \textbf{week}.\\
\textbf{Multivariate Example}: In \textbf{week \textit{24}}, your
\textbf{calorie intake} was \textbf{moderate}, then \textbf{very low},
then \textbf{high}, then \textbf{very high}, then \textbf{low}, then
\textbf{moderate} and your \textbf{carbohydrate intake} was
\textbf{moderate}, then \textbf{high}, then \textbf{very low}, then
\textbf{high}. During \textbf{half of the \textit{weeks}} similar to
\textbf{week \textit{24}}, your \textbf{calorie intake \textit{dropped}}
and your \textbf{carbohydrate intake \textit{stayed the same}} the next
\textbf{week}.
\end{description}



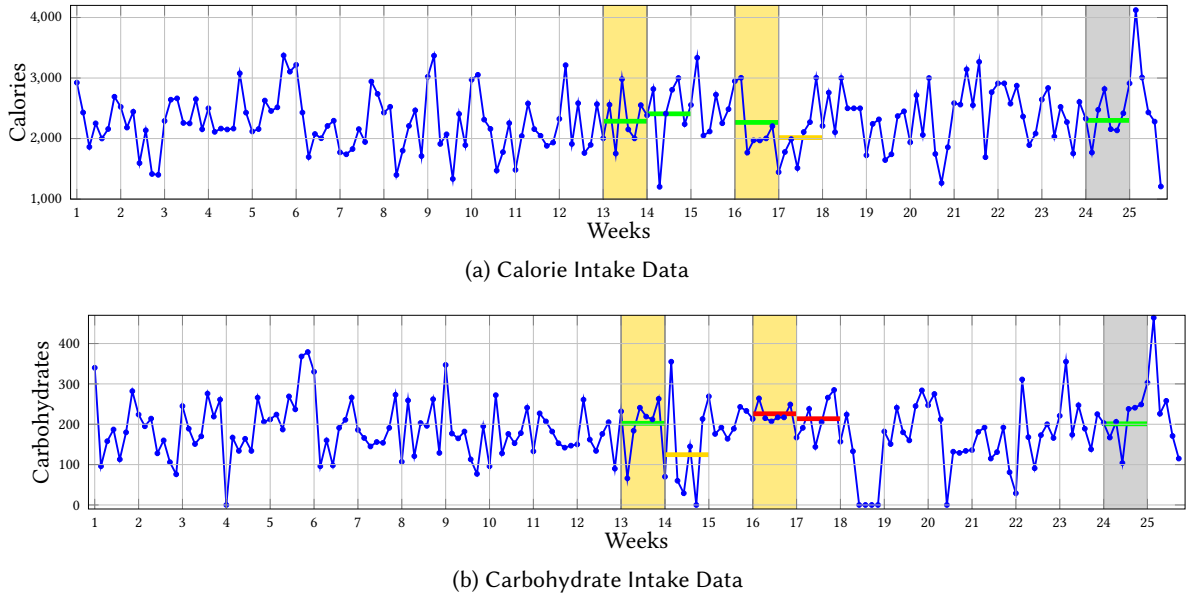
\begin{figure}[!t]
  \centering
    \pgfplotsset{every axis/.append style = {
            tick label style={font=\tiny}},
            x coord trafo/.code={\pgfmathparse{(#1-1)/7+1}\pgfmathresult},
        x coord inv trafo/.code={\pgfmathparse{#1}\pgfmathresult},
    }
    \centerline{
        \hspace{-0.25in}
        \subfloat[Calorie Intake Data]{
  \label{fig:calorie_cb}
            \scalebox{0.9}{
            \begin{tikzpicture}
            \begin{axis}[%
                axis on top,
                width=7in, height=1.75in,
                grid=both,%
                label shift = -5pt,
                xlabel=Weeks, xmin=0, xmax=175, ymin=1000,ymax=4200,%
                xtick distance=1, mark size = 1pt,%
                ylabel=Calories,%
                ytick distance=1000, thick]
                \addplot+[color=blue, mark=*, mark options={fill=blue}] 
                    table[x=Day,y=Calories,col sep=comma]  {calorie_data.csv};
                \addplot[green, no markers, line width=2pt]
                    coordinates {(85,2286.4) (92,2286.4)};
                \addplot[green,no markers, line width=2pt]
                    coordinates {(92,2408.7) (99,2408.7)};
                \addplot[green, no markers, line width=2pt]
                    coordinates {(106,2266.7) (113,2266.7)};
                \addplot[gold, no markers, line width=2pt]
                    coordinates {(113,2015.7) (120,2015.7)};
                \addplot[green, no markers, line width=2pt]
                    coordinates {(162,2300) (169,2300)};
                \addplot[name path=vA1,black, no markers, line width=0pt]
                    coordinates {(85,0) (85,4200)};
                \addplot[name path=vB1,black, no markers, line width=0pt]
                    coordinates {(92,0) (92,4200)};
                \addplot[gold!70,fill opacity=0.7] fill between[of=vA1
                    and vB1];                
                \addplot[name path=vA,black, no markers, line width=0pt]
                    coordinates {(106,0) (106,4200)};
                \addplot[name path=vB,black, no markers, line width=0pt]
                    coordinates {(113,0) (113,4200)};
                \addplot[gold!70,fill opacity=0.7] fill between[of=vA
                    and vB];                
                \addplot[name path=vC,black, no markers, line width=0pt]
                    coordinates {(162,0) (162,4200)};
                \addplot[name path=vD,black, no markers, line width=0pt]
                    coordinates {(169,0) (169,4200)};
                \addplot[gray!70, fill opacity=0.5] fill between[of=vC and vD];                
            \end{axis}
            \end{tikzpicture}
}}}
\centerline{
        \subfloat[Carbohydrate Intake Data]{
  \label{fig:carb_cb}
            \scalebox{0.9}{
            \begin{tikzpicture}
            \begin{axis}[name=carb_plot,%
                axis on top,
                width=7in, height=1.75in,
                grid=major, mark size=1pt, %
                label shift = -5pt,
                xlabel=Weeks, xmin=0, xmax=175, ymin=-10, ymax=470,%
                xtick distance=1, ylabel=Carbohydrates,%
                ytick distance=100, thick]
                \addplot+[color=blue, mark=*, mark options={fill=blue}] 
                    table[x={Day},y={Carbohydrates},col sep=comma]  {carb_data.csv};
                %
                \addplot[green, no markers, line width=2pt]
                    coordinates {(85,202.4) (92,202.4)};
                \addplot[gold, no markers, line width=2pt]
                    coordinates {(92,124.6) (99,124.6)};
                \addplot[red, no markers, line width=2pt]
                    coordinates {(106,226.0) (113,226.0)};
                \addplot[red, no markers, line width=2pt]
                    coordinates {(113,213.7) (120,213.7)};
                \addplot[green, no markers, line width=2pt]
                    coordinates {(162,201.1) (169,201.1)};
                \addplot[name path=vA1,black, no markers, line width=0pt]
                    coordinates {(85,-10) (85,4200)};
                \addplot[name path=vB1,black, no markers, line width=0pt]
                    coordinates {(92,-10) (92,4200)};
                \addplot[gold!70,fill opacity=0.7] fill between[of=vA1
                    and vB1];                
                \addplot[name path=vA,black, no markers, line width=0pt]
                    coordinates {(106,-10) (106,4200)};
                \addplot[name path=vB,black, no markers, line width=0pt]
                    coordinates {(113,-10) (113,4200)};
                \addplot[gold!70,fill opacity=0.7] fill between[of=vA
                    and vB];                
                \addplot[name path=vC,black, no markers, line width=0pt]
                    coordinates {(162,-10) (162,4200)};
                \addplot[name path=vD,black, no markers, line width=0pt]
                    coordinates {(169,-10) (169,4200)};
                \addplot[gray!70, fill opacity=0.5] fill between[of=vC and vD];                
            \end{axis}
            \end{tikzpicture}
    }}}
\vspace{-0.1in}
  \caption{Summary Provenance - Cluster-Based Pattern. The gray region is the
  past week and the golden regions are the weeks similar to it. The different
  colored line segments denote the summarizers at week level -- yellow
  is ``low,'' green ``moderate,'' and red ``high.''}
\vspace{-0.1in}
\end{figure}

Here $m_i$ is the number of summarizers for attribute $i$, and $n$ is
the number of attributes.
Note that the quantifier is calculated from the cluster alone instead of
the entire dataset. We can see that, in every summary of this type, the
description of the time window comes first. The description is then
followed by the actual protoform. From these summaries, the user in our
running example is able
to know how exactly their past week went for each nutrient. The user can
also conclude that their calorie intake will likely drop the next week,
while their carbohydrate intake has around a 50\% chance of staying the
same.
Figures \ref{fig:calorie_cb} and 
\ref{fig:carb_cb} illustrate the multivariate summary. In these charts,
the gray vertical range represents the past week, while the golden
vertical ranges represents the weeks considered similar to the past
week. 

The user may also wish to focus on the most recent week. Despite the
conclusions stated by the cluster-based pattern summaries, it is
possible that the user has not behaved this way recently. Cluster-based
pattern summaries can also be used for what we call a standard pattern
protoform:
\begin{description}
\small
\item \textbf{Standard Pattern Protoform}: The last time you had a
    $\langle$time window$\rangle$ similar to $\langle$time window$\rangle$ $\langle$number$\rangle$,
    your $\langle$attribute 1$\rangle$ $\langle$summarizer 1$\rangle$,..., and your $\langle$attribute
    $n\rangle$ $\langle$summarizer $n\rangle$ the next $\langle$time window$\rangle$.\\
\textbf{Univariate Example}: The last time you had a \textbf{week} similar to \textbf{week \textit{24}}, your \textbf{calorie intake \textit{dropped}} the next \textbf{week}.

\textbf{Multivariate Example}: The last time you had a \textbf{week} similar to \textbf{week \textit{24}}, your \textbf{calorie intake \textit{dropped}} and your \textbf{carbohydrate intake \textit{stayed the same}} the next \textbf{week}.
\end{description}



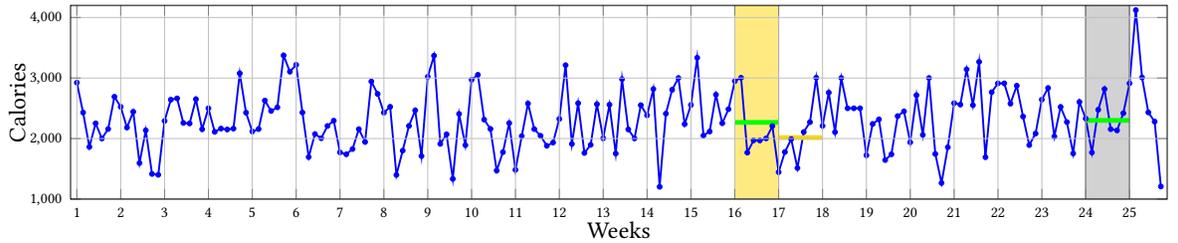
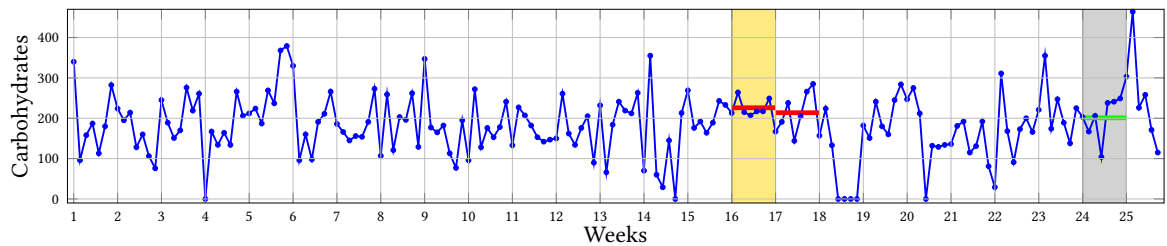
\begin{figure}[!t]
  \centering
    \pgfplotsset{every axis/.append style = {
            tick label style={font=\tiny}},
            x coord trafo/.code={\pgfmathparse{(#1-1)/7+1}\pgfmathresult},
        x coord inv trafo/.code={\pgfmathparse{#1}\pgfmathresult},
    }
    \centerline{
        \hspace{-0.25in}
        \subfloat[Calorie Intake Data]{
            \label{fig:calorie_sp}
            \scalebox{0.9}{
            \begin{tikzpicture}
            \begin{axis}[%
                axis on top,
                width=7in, height=1.75in,
                grid=both,%
                label shift = -5pt,
                xlabel=Weeks, xmin=0, xmax=175, ymin=1000,ymax=4200,%
                xtick distance=1, mark size = 1pt,%
                ylabel=Calories,%
                ytick distance=1000, thick]
                \addplot+[color=blue, mark=*, mark options={fill=blue}] 
                    table[x=Day,y=Calories,col sep=comma]  {calorie_data.csv};
                \addplot[green, no markers, line width=2pt]
                    coordinates {(106,2266.7) (113,2266.7)};
                \addplot[gold, no markers, line width=2pt]
                    coordinates {(113,2015.7) (120,2015.7)};
                \addplot[green, no markers, line width=2pt]
                    coordinates {(162,2300) (169,2300)};
                \addplot[name path=vA,black, no markers, line width=0pt]
                    coordinates {(106,0) (106,4200)};
                \addplot[name path=vB,black, no markers, line width=0pt]
                    coordinates {(113,0) (113,4200)};
                \addplot[gold!70,fill opacity=0.7] fill between[of=vA and vB];                
                \addplot[name path=vC,black, no markers, line width=0pt]
                    coordinates {(162,0) (162,4200)};
                \addplot[name path=vD,black, no markers, line width=0pt]
                    coordinates {(169,0) (169,4200)};
                \addplot[gray!70,fill opacity=0.5] fill between[of=vC
                    and vD];                
            \end{axis}
            \end{tikzpicture}
}}}
\centerline{
        \subfloat[Carbohydrate Intake Data]{
            \label{fig:carb_sp}
            \scalebox{0.9}{
            \begin{tikzpicture}
            \begin{axis}[name=carb_plot,%
                axis on top,
                width=7in, height=1.75in,
                grid=major, mark size=1pt, %
                label shift = -5pt,
                xlabel=Weeks, xmin=0, xmax=175, ymin=-10, ymax=470,%
                xtick distance=1, ylabel=Carbohydrates,%
                ytick distance=100, thick]
                \addplot+[color=blue, mark=*, mark options={fill=blue}] 
                    table[x={Day},y={Carbohydrates},col sep=comma]  {carb_data.csv};
                %
                \addplot[red, no markers, line width=2pt]
                    coordinates {(106,226.0) (113,226.0)};
                \addplot[red, no markers, line width=2pt]
                    coordinates {(113,213.7) (120,213.7)};
                \addplot[green, no markers, line width=2pt]
                    coordinates {(162,201.1) (169,201.1)};
                \addplot[name path=vA,black, no markers, line width=0pt]
                    coordinates {(106,-10) (106,4200)};
                \addplot[name path=vB,black, no markers, line width=0pt]
                    coordinates {(113,-10) (113,4200)};
                \addplot[gold!70,fill opacity=0.7] fill between[of=vA and vB];                
                \addplot[name path=vC,black, no markers, line width=0pt]
                    coordinates {(162,-10) (162,4200)};
                \addplot[name path=vD,black, no markers, line width=0pt]
                    coordinates {(169,-10) (169,4200)};
                \addplot[gray!70,fill opacity=0.5] fill between[of=vC
                    and vD];                
            \end{axis}
            \end{tikzpicture}
    }}}
\vspace{-0.1in}
  \caption{Summary Provenance - Standard Pattern. The gray region is the
  past week and the golden region is the most recent week similar to it. The different
  colored line segments denote the summarizers at week level -- yellow
  is ``low,'' green ``moderate,'' and red ``high.''}
\vspace{-0.1in}
\end{figure}

The user can now see from the standard pattern summaries that (for the
most part) the behavior found by the cluster-based pattern summaries
still repeats in the most recent week. This reinforces the user's
motivation to work towards changing or maintaining their behavior during the next week.
Figures \ref{fig:calorie_sp} and 
\ref{fig:carb_sp} verify the multivariate summary.

\begin{figure}[!t]
 \centering
 \includegraphics[width=0.1\textwidth, height=1in]{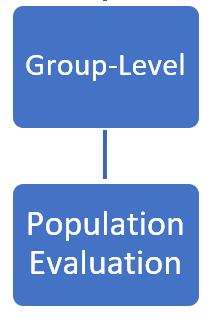}
\vspace{-0.1in}
 \caption{Summary Types (Group-Level)}
 \label{fig:grouplevel}
\vspace{-0.1in}
\end{figure}

\subsection{Group-Level Summaries}
Moving away from the point of view of the user, suppose we have researchers or clinicians
who wish to evaluate an entire population of users or patients. What if they want to know how
the entire user population is faring as a whole or how a particular user compares to other users
(where such data is available)?
We can use group-level summaries to find
this answer.

\paragraph{Population evaluation summaries}
Our system currently generates
population evaluation summaries for nutrient intake data (see Figure
\ref{fig:grouplevel}), which has further sub-types described below.
Our approach can summarize the study population as a whole using the
individual summaries previously generated by our system. If the user
wishes to know how they compare against other users in terms of their
calorie and carbohydrate intake in the past week, this
protoform can be used:
\begin{description}
\small
\item \textbf{Population Evaluation Protoform}: $\langle$Quantifier 1$\rangle$ participants
    in this study had a $\langle$summarizer 1$\rangle$ $\langle$attribute 1$\rangle$, a
    $\langle$summarizer 2$\rangle$ $\langle$attribute 2$\rangle$,..., and a $\langle$summarizer $n\rangle$
    $\langle$attribute $n\rangle$ $\langle$sub-protoform$\rangle$.\\
\textbf{Univariate Example (Standard Evaluation (TW))}: \textbf{Some of the} participants in this study had a \textbf{moderate \textit{calorie intake} in the past full week}.

\textbf{Multivariate Example (Standard Evaluation (TW))}: \textbf{Some of the} participants in this study had a \textbf{high \textit{calorie intake}} and a \textbf{very high \textit{carbohydrate intake} in the past full week}.
\end{description}

Here we define $\langle$sub-protoform$\rangle$ as a portion of an actual summary used to
describe a number of users in the dataset. This ``sub-protoform''
identifies the summary type the population has been evaluated on and the conclusion found.
The user in our running example can use these summaries to know that at least some of the users
in the study did better at managing their calorie intake in the past
full week. Now, the user may be more motivated to make changes in their
diet for the upcoming week.

There are also special cases of the group-level protoforms for some of the aforementioned protoform types, 
namely the cluster-based pattern, standard pattern, and (day) if-then pattern protoforms.
For the cluster-based pattern protoform, we have:
\begin{description}
\small
\item \textbf{Population Evaluation Protoform (Cluster-Based Pattern)}: After looking at clusters containing $\langle$time window (plural)$\rangle$ similar to this past one, it can be seen that $\langle$quantifier$\rangle$ participants with these clusters may see $\langle$summarizer 1$\rangle$ in their $\langle$attribute 1$\rangle$, ... , and $\langle$summarizer n$\rangle$ in their $\langle$attribute n$\rangle$ next $\langle$time window$\rangle$.\\
\textbf{Univariate Example (Cluster-Based Pattern)}: After looking at clusters containing \textbf{weeks} similar to this past one, it can be seen that \textbf{some of the} participants with these clusters may see \textbf{a rise} in their \textbf{calorie intake} next \textbf{week}.

\textbf{Multivariate Example (Cluster-Based Pattern)}: After looking at clusters containing \textbf{weeks} similar to this past one, it can be seen that \textbf{almost none of the} participants with these clusters may see \textbf{a rise} in their \textbf{calorie intake} and \textbf{little to no change} in their \textbf{carbohydrate intake} next \textbf{week}.
\end{description}
For the standard pattern protoform, we have:
\begin{description}
\small
\item \textbf{Population Evaluation Protoform (Standard Pattern)}: Based on the most recent $\langle$time window$\rangle$ similar to this past one, it can be seen that $\langle$quantifier$\rangle$ participants may see $\langle$summarizer 1$\rangle$ in their $\langle$attribute 1$\rangle$, ... , and $\langle$summarizer n$\rangle$ in their $\langle$attribute n$\rangle$ next $\langle$time window$\rangle$.\\
\textbf{Univariate Example (Standard Pattern)}: Based on the most recent \textbf{weeks} similar to this past one, it can be seen that \textbf{some of the} participants may see \textbf{a drop} in their \textbf{calorie intake} next \textbf{week}.

\textbf{Multivariate Example (Standard Pattern)}: Based on the most recent \textbf{weeks} similar to this past one, it can be seen that \textbf{almost none of the} participants may see \textbf{a rise} in their \textbf{calorie intake} and \textbf{little to no change} in their \textbf{carbohydrate intake} next \textbf{week}.
\end{description}
Finally, for the if-then pattern protoform, we have:
\begin{description}
\small
\item \textbf{Population Evaluation Protoform (If-Then Pattern)}: For $\langle$quantifier$\rangle$ participants in this study, it is true that when your $\langle$attribute 1$\rangle$ is
    $\langle$summarizer 1:1$\rangle$, then $\langle$summarizer 2:1$\rangle$,..., then $\langle$summarizer $m$:1$\rangle$,..., and your $\langle$attribute $n\rangle$ is $\langle$summarizer 1:$n\rangle$, then $\langle$summarizer 2:$n\rangle$,..., then $\langle$summarizer $m$:$n\rangle$, their $\langle$attribute 1$\rangle$ tends to be
    $\langle$summarizer $(m+1)$:1$\rangle$,..., and their $\langle$attribute $n\rangle$ tends
    to be $\langle$summarizer $(m+1)$:$n\rangle$ the next $\langle$time window$\rangle$.\\
\textbf{Univariate Example (If-Then Pattern)}: For \textbf{all of the} participants in this study, it is true that when their \textbf{calorie intake} follows the pattern of being \textbf{very high}, their \textbf{calorie intake} tends to be \textbf{high} the next \textbf{day}.

\textbf{Multivariate Example (If-Then Pattern)}: For \textbf{all of the} participants in this study, it is true that when their \textbf{calorie intake} follows the pattern of being \textbf{low}, their \textbf{calorie intake} tends to be \textbf{moderate} and their \textbf{carbohydrate intake} tends to be \textbf{moderate} the next \textbf{day}.
\end{description}
The population evaluation protoform for day if-then pattern protoform differs in the same way as the regular if-then pattern protoform.

\section{Summary Generation and Mining}
\label{sec:approach}
Having described the different protoforms and concrete examples of
summaries on the real user data, we now outline our summary generation
approach. 


\paragraph{Representing Time Series as Symbolic Sequences}
In order to find interesting discoveries from time-series data, such as frequent
patterns and anomalistic behavior, we first represent the raw
time-series data in symbolic form. To achieve this, we use the SAX
symbolic representation \cite{sax} for each time series, which also
makes it easier to represent the time series at different granularities.


\begin{figure}[!ht]
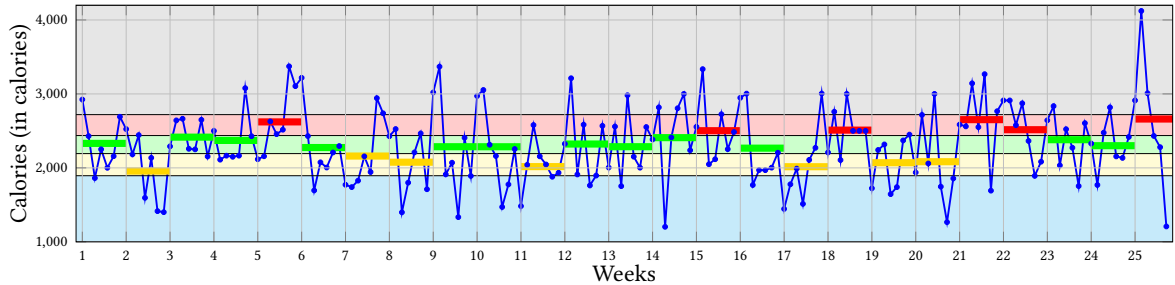
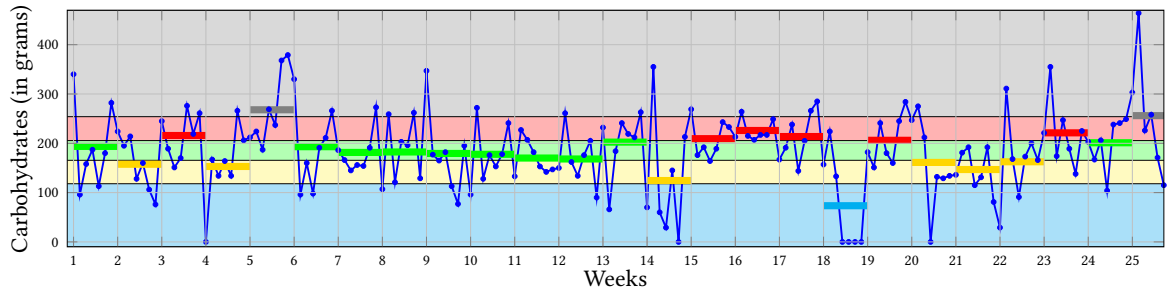

    \centering
    \pgfplotsset{every axis/.append style = {
            tick label style={font=\tiny}},
            x coord trafo/.code={\pgfmathparse{(#1-1)/7+1}\pgfmathresult},
        x coord inv trafo/.code={\pgfmathparse{#1}\pgfmathresult},
    }
    \centerline{
        \hspace{-0.25in}
        \subfloat[Calorie Intake Data: SAX week sequence `cbccdcbbccbcccdcbdbbddccd']{
            \label{fig:SAXcalorie}
            \scalebox{0.9}{
            \begin{tikzpicture}
            \begin{axis}[name=calorie_plot,%
                axis on top,
                width=7in, height=2in,
                grid=both,%
                xlabel=Weeks, xmin=0, xmax=175, ymin=1000,ymax=4200,%
                xtick distance=1, mark size = 1pt,%
                ylabel=Calories (in calories),%
                label shift = -5pt,
                ytick distance=1000, thick]
                \addplot+[color=blue, mark=*, mark options={fill=blue}] 
                    table[x=Day,y=Calories,col sep=comma]  {calorie_data.csv};
                \addplot[name path=A,black,no markers, line width=0pt]
                    coordinates {(0,0) (175,0)};
                \addplot[name path=B,black, no markers, line width=0pt]
                    coordinates {(0,1894) (175,1894)};
                \addplot[name path=C,black, no markers, line width=0pt]
                    coordinates {(0,2193) (175,2193)};
                \addplot[name path=D,black, no markers, line width=0pt]
                    coordinates {(0,2437) (175,2437)};
                \addplot[name path=E,black, no markers, line width=0pt]
                    coordinates {(0,2720) (175,2720)};
                \addplot[name path=F,black, no markers, line width=0pt]
                    coordinates {(0,4200) (175,4200)};
                \addplot[cyan!20] fill between[of=A and B];                
                \addplot[yellow!20] fill between[of=B and C];                
                \addplot[green!20] fill between[of=C and D];                
                \addplot[red!20] fill between[of=D and E];                
                \addplot[gray!20] fill between[of=E and F];                
\addplot[color=gold, no markers, line width=3pt] coordinates{
        \input{calorie_saxB.dat} 
};
\addplot[color=green, no markers, line width=3pt] coordinates{
        \input{calorie_saxC.dat} 
};
\addplot[color=red, no markers, line width=3pt] coordinates{
        \input{calorie_saxD.dat} 
};
            \end{axis}
            \end{tikzpicture}
        }}}
    \centerline{
        \hspace{-0.25in}
        \subfloat[Carbohydrate Intake Data: SAX week sequence `cbdbeccccccccbdddaccbbdce']{
            \label{fig:SAXcarb}
            \scalebox{0.9}{
            \begin{tikzpicture}
            \begin{axis}[name=carb_plot,%
                axis on top,
                label shift = -5pt,
                width=7in, height=2in,
                grid=major, mark size=1pt, %
                xlabel=Weeks, xmin=0, xmax=174, ymin=-10, ymax=470,%
                xtick distance=1, ylabel=Carbohydrates (in grams),%
                ytick distance=100, thick]
            \addplot+[color=blue, mark=*, mark options={fill=blue}] 
                table[x=Day,y=Carbohydrates,col sep=comma]  {carb_data.csv};
                \addplot[name path=A,black,no markers, line width=0pt]
                    coordinates {(0,-10) (175,-10)};
                \addplot[name path=B,black, no markers, line width=0pt]
                    coordinates {(0,118) (175,118)};
                \addplot[name path=C,black, no markers, line width=0pt]
                    coordinates {(0,165.5) (175,165.5)};
                \addplot[name path=D,black, no markers, line width=0pt]
                    coordinates {(0,205.5) (175,205.5)};
                \addplot[name path=E,black, no markers, line width=0pt]
                    coordinates {(0,254) (175,254)};
                \addplot[name path=F,black, no markers, line width=0pt]
                    coordinates {(0,470) (175,470)};
                \addplot[cyan!30] fill between[of=A and B];                
                \addplot[yellow!30] fill between[of=B and C];                
                \addplot[green!30] fill between[of=C and D];                
                \addplot[red!30] fill between[of=D and E];                
                \addplot[gray!30] fill between[of=E and F];                
\addplot[color=cyan, no markers, line width=3pt] coordinates{
(120, 73.4) (127, 73.4)
};
\addplot[color=gold, no markers, line width=3pt] coordinates{
        \input{carb_saxB.dat} 
};
\addplot[color=green, no markers, line width=3pt] coordinates{
        \input{carb_saxC.dat} 
};
\addplot[color=red, no markers, line width=3pt] coordinates{
        \input{carb_saxD.dat} 
};
\addplot[color=gray, no markers, line width=3pt] coordinates{
        \input{carb_saxE.dat} 
};
            \end{axis}
            \end{tikzpicture}
    }}
}
\vspace{-0.1in}
\caption{SAX Representation for the (a) Calorie and (b) Carbohydrate (b) Intake Data for a user from
MyFitnessPal dataset~\cite{mfp}. The different colored regions
correspond to the different letters -- a, b, c, d, and e -- that we map to
different summarizers -- very low (cyan), low (yellow),
moderate (green), high (red), and very high (gray), respectively. 
Each horizontal segment represents the SAX symbol for each
week, whereas individual data points can be seen to belong to the
different regions. The SAX representation (sequence over the letters
a-e) at the weekly granularity is shown for both calorie and
carbohydrate intake.}
\label{fig:sax}
\vspace{-0.1in}
\end{figure}

The symbols, in particular, are letters from some alphabet. Provided an
alphabet size $n$ and the time window size, SAX \textit{z}-normalizes
the raw data of each time series to a zero mean with a standard
deviation of 1. It, then, uses Piecewise Approximate Aggregation (PAA)
to reduce the dimensionality of each time series, depending on the time
window size. This reduction allows the ability to easily switch between
temporal granularities. After the data is projected onto its principle components
and normalized, SAX generates $n$ equiprobability bins based on the
standard Gaussian distribution with each segment represented by its
corresponding bin symbol.
Figure~\ref{fig:sax} shows the SAX representation of both calorie and
carbohydrate intake data for our example user. The SAX representation at the
daily level can be read by simply associating each colored region in the
figure with the corresponding letter from the alphabet comprising a, b,
c, d, and e (corresponding to `very low', `low', `moderate', `high',
`very high', respectively). The SAX representation at the week level
correspond to the similarly colored line segments. For example, as
observed in Figure~\ref{fig:SAXcarb}, the week level categorical
sequence for carbohydrates is: `cbdbeccccccccbdddaccbbdce'.




\subsection{Pattern Mining}
\label{sec:mining}
We employ two different types of pattern mining approaches, based on
clustering and frequent sequence mining.

\paragraph{Cluster-based patterns}
After partitioning the sub-time window SAX representation into time
window tuples (e.g., chunking a string of days into weeks), we combine
multiple time series into multivariate symbolic sequences. For example,
if one variable has the SAX sequence ``abacbbc'' and another the
sequence ``bccabcc,'' then the combined multivariate sequence is ``a-b,
b-c, a-c, c-a, b-b, b-c, c-c'', where the symbols in the corresponding
positions have been combined into an ``event.'' We, then, group these
combined sequences into clusters. For clustering, we use Squeezer
\cite{squeezer}, which is an online clustering algorithm for categorical
data that only needs a similarity threshold $s$ to find clusters. For
each tuple $t$, Squeezer assigns it to an existing cluster or creates a
new cluster based on the similarities between $t$ and the existing
clusters, using threshold $s$. A sampling-based approach is used to
determine $s$. We sample a fraction $f$ of the window tuples and
calculate the average similarity between each pair of tuples in the
sample, using
\begin{equation}
\label{eq:similarity}
sim(T_i; T_j) = \Big|\big\{A_k| T_i.A_k = T_j.A_k, 1 \leq k \leq n\big\}\Big|
\end{equation}
where $T_i$ and $T_j$ are tuples, $A_k$ is the SAX symbol at index $k$,
and $n$ is the time-window size. In other words, the similarity is based on the
number of matching symbols at corresponding positions. We repeat this
process $1/f$ times (e.g., if we sample $f=0.2$ or 20\% of the tuples,
we repeat the sampling $1/f=5$ times), and set $a$ as the mean of all
the average pair-wise similarities. Finally, we set $s = a + 1$, as
suggested in \cite{squeezer}.

Each cluster now contains non-consecutive time windows that have been
grouped together by similarity. From these clusters, we can use the
history of the attributes involved to ``predict'' what may happen in the
time window following the one we are interested in. Typically we 
choose the most recent time window in
an attempt to ``predict'' the future, although it may also be beneficial
to use another time window. If we were to use a time window other than
the most recent one, we can extract the expected result for the
following time window. In short, if we have a time window $TW_i$, we
should be able to use the result of similar time windows to see the
expected outcome of the following time window. 

For each cluster, we pair each tuple with the tuple that follows it,
e.g., pair each week in a cluster with the week following it. Next, we
replace the tuples, which are at the sub-time window level ($sTW$), with
the time-window level ($TW$) SAX symbols. These time-window level pairs
are used to generate cluster-based pattern summaries. In order to
describe a pattern, we map the letters (typically, in the last full
week) to their corresponding summarizers.

\paragraph{Frequent sequence mining for if-then patterns}
To generate if-then pattern summaries, we employ frequent sequence
mining over the symbolic SAX temporal data using SPADE \cite{spade}.
Frequent means that the pattern appears more than a user-specified
value called {\em minimum support}. The method outputs all of the frequent
sequence patterns found in the data.




For each frequent sequence, we map each of its prefixes to the following
suffixes. For instance, if ``abca'' is a frequent sequence, then we
consider the pairs: (`a', `bca'), (`ab', `ca'), and
(`abc', `a'). A similar approach is taken for multivariate data.
Next, we generate confidence values (or
conditional probability) of observing the suffix given the prefix, given
as
\begin{equation}\label{eq:conf}
P(\text{suffix} | \text{prefix}) = \frac{\text{count(prefix + suffix)}}{\text{count(prefix)}}
\end{equation}
where $count(seq)$ is the frequency of the sequence $seq$. We use a
minimum confidence threshold to retain only those frequent if-then
patterns of the form ``If \{prefix\}, then \{suffix\}'' with the highest
confidence values. Finally, these patterns are used to generate
summarizers to be presented to the user.

\subsection{Summary Generation}
\label{sec:generation}
In order to generate summaries, we fill in the blanks of protoforms
presented in Section \ref{sec:summaries} using summarizers from Table
\ref{tab:summarizers} and quantifiers from Table \ref{tab:assignments}.
The data we look at is also modified depending on the summary type. For
instance, for standard evaluation, the data is the past full week of
the data. 

As there are many possible combinations of summarizers and quantifiers
for each attribute, we choose a combination that is ``most
appropriate'' based on the \textit{average membership function} for a
summarizer $S$ and a quantifier $Q$. 
We denote $\mu_S$ as the membership function value for summarizer $S$ in
a time window $TW$. The membership value $\mu_S$ will either have the
value of 1 or 0, based on whether the value $v$ for
attribute $A$ of the time window follows the conclusion implied by the summarizer. For
example, when evaluating a goal for calorie intake where a user wishes
to eat at most 2,000 calories a day, the possible summarizers would be
``reached'' or ``did not reach,'' according to Table
\ref{tab:summarizers} (for the standard evaluation summary with a goal).
In this case, a value $v$ less than or equal to 2,000 would imply that
the user ``reached'' the goal, while a value $v$ greater than 2,000
would imply that the user ``did not reach'' the goal. 

From the $\mu_S$ for each time window, we calculate the aggregated
average 
\begin{equation}\label{eq:ratio}
r_S = \frac{1}{n}\sum_{i=1}^{n} \mu_S(y_i)
\end{equation}
where $y_i$ is a data point in the time-series data and $n$ is the size
or length of the time-series. This fraction $r_S$ indicates the percentage of the dataset
that agrees with the summarizer $S$.


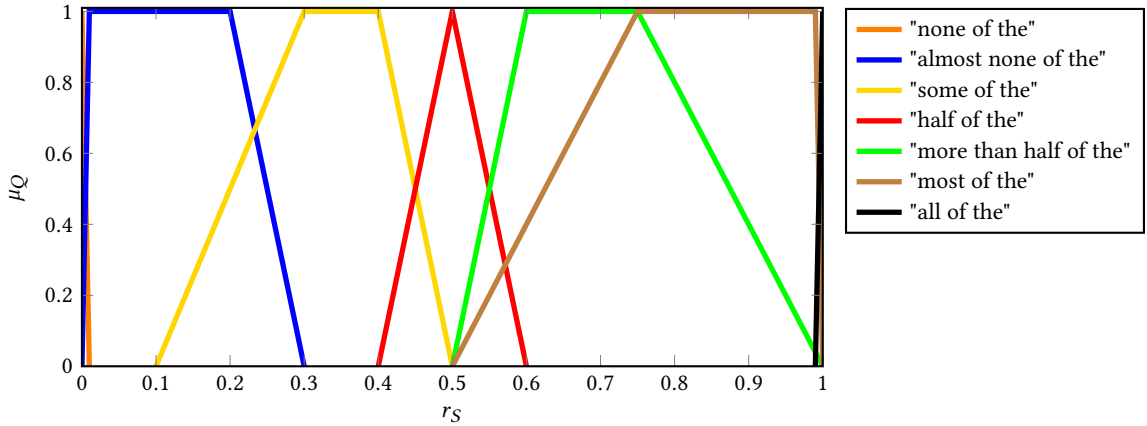
\begin{figure}[!h]
  \centering
  \small
  \hspace{-0.25in}
\begin{tikzpicture}
    \begin{axis}[%
        axis on top,
        legend style={font=\small},
        width=4.5in, height=2.5in,
        xlabel=$r_S$, ylabel=$\mu_Q$,
        xmin=0, xmax=1, ymin=0, ymax=1.01,%
        xtick distance=0.1, ytick distance=0.2,
        legend pos=outer north east,
        legend style={nodes={anchor=west}}, 
        every axis plot/.append style={ 
          line width=2pt,
          samples=2, 
        },
        thick]
        \addplot[color=orange, no markers] coordinates { (0,1) (0.01,0) };
        \addplot[color=blue, no markers] coordinates { 
                (0,0) (0.01,1) (0.2,1) (0.3, 0)
        };
        \addplot[color=gold, no markers] coordinates { 
                (0.1,0) (0.3,1) (0.4,1) (0.5, 0)
        };
        \addplot[color=red, no markers] coordinates { 
                (0.4,0) (0.5,1) (0.6, 0)
        };
        \addplot[color=green, no markers] coordinates { 
                (0.5,0) (0.6,1) (0.75, 1)(1,0)
        };
        \addplot[color=brown, no markers] coordinates { 
                (0.5,0) (0.75,1) (0.99,1) (1, 0)
        };
        \addplot[color=black, no markers] coordinates { 
                (0.99,0) (1,1)
        };
\legend{"none of the", "almost none of the", "some of the", 
    "half of the",
"more than half of the", "most of the", "all of the"};
    \end{axis}
\end{tikzpicture}
\vspace{-0.1in}
\caption{Quantifier Membership Functions}
  \label{fig:trapezoidal}
\vspace{-0.1in}
\end{figure}

Once we obtain the $r_S$ value for each summarizer $S$, we
use this value to determine the best quantifier for each summarizer.
We employ the use of
trapezoidal membership functions \cite{yagerapproach} $\mu_Q$ to
calculate how well $r_S$ fits each quantifier. As $\mu_S$ is the
membership function of a summarizer $S$, $\mu_Q$ is the membership
function of a quantifier $Q$.
For example, in Figure \ref{fig:trapezoidal} (in brown), for the quantifier
``most of the'', we have
\begin{equation}\label{eq:muQ}
\mu_Q(r_S) = 
    \begin{cases}
        4 r_S - 2 & 0.5 < r_S < 0.75 \\
        1 & 0.75 \leq r_S \leq 0.9 \\
        -10 r_S + 10 & 0.9 < r_S < 1 \\
        0 & \text{otherwise}
    \end{cases}
\end{equation}
We have defined membership functions for each possible
quantifier based on the approach in \cite{businessdata}. They were
designed to create trapezoidal functions that match with the values
we believe best fit each quantifier. 
The membership functions for the different quantifiers are
plotted in \cref{fig:trapezoidal}.

Once we have the best quantifier for each summarizer, we will have $k$
quantifier-summarizer candidate pairs. The pair that contains the
quantifier with the highest $\mu_Q$ will be chosen for the summary,
while the value of $\mu_Q$ eventually becomes the summary's {\em truth
value}. 
When the most appropriate summarizer and quantifier are found, they are
used within the protoform or template to generate the summary. 
As a result, we generate a list of candidate summaries, 
paired with a truth value using the $\mu_Q$ value of the quantifier
within the summary \cite{fuzzyquant}. Finally, we choose the summary
with the highest truth value $\mu_Q$, breaking ties by selecting the
summary with the quantifier that implies the largest amount (following
Yager's approach for text quality \cite{yagerapproach}). 

\subsubsection{Summary Metrics} \label{metrics}
To evaluate the summaries generated by our system, we use five
evaluation metrics, which help measure the Gricean maxims
\cite{grice}: the maxims of quality, quantity, relevance, and manner. These
maxims are well-known pragmatic
rules that are known to improve communication of information to humans
\cite{boran2016overview}, especially for natural language summaries. 
The evaluation metrics we use on our
summaries are the degrees of truth, imprecision, covering,
appropriateness, and coverage, along with the length quality
\cite{boran2016overview,businessdata,ifthen}.
For all of the metrics, they range from 0 to 1 in value, where 1 is the ideal value.

\paragraph{Degree of Truth $(T_1)$:}
First and foremost, we want the summaries that our framework generates to
convey the degree of truth. 
We use natural
language to summarize how often a finding may be true in the data. We
use  fuzzy quantifiers to describe the frequencies of
certain behaviors that best fit the percentage found in the data, 
although the
truthfulness of the overall summary may not be absolute. 
Zadeh's degree of truth \cite{fuzzyquant} determines which summaries are
actually true statements. We use this degree to measure to what extent our
summaries follow the maxim of quality, which states how true a summary
is and how much evidence supports it. 
As discussed above, Eq. \eqref{eq:ratio} calculates 
the percentage of the dataset that supports the
summarizer $S$.
Then, we calculate the $\mu_Q$ for the quantifier in
each quantifier-summarizer pair via Eq. \eqref{eq:muQ}, which in fact
represents a summary's truth value. 
For the remainder of this paper, we will refer to the degree of
truth as $T_1 = \mu_Q$.

\paragraph{Degree of Imprecision $(T_2)$:}
Also known as the degree of fuzziness, the degree of imprecision
measures how useful a summary is. It is highly possible that a summary
is generated that has a high degree of truth but is also a statement
that is not useful, such as ``All winter days are cold.'' 

Recall that $r_S$ indicates the fraction of the dataset that agrees with
the summarizer $S$.
In our summary generation approach, we keep track of percentages
$r_{S_j}$ of each possible summarizer $S_j$. In order to calculate the degree of imprecision,
we use the following equation:
\begin{equation}\label{eq:imprecision}
T_2 = 1 - \sqrt[m]{\prod_{j=1...m} r_{S_j}}
\end{equation}
where $m$ is the number of possible summarizers for the protoform type. 
Here, we compute the geometric mean of the percentages of agreement over
the possible summarizers $S_j$. For the case where a summary is obvious,
every summarizer $S_j$ would have a membership value $\mu_{S_j} > 0$ for
every (sub-)time window.
For example, if every day was snowy and cold, then it would
be unwise to output a summary describing this trend. When we subtract
the geometric mean from 1, the degree of imprecision represents the
extent to which the summary is useful. 

\paragraph{Degree of Covering $(T_3)$:}
The degree of covering measures how many points in a user's query
are covered by the summary. When we refer to the user's query, we are
referring to the subset $d$ of the dataset $D$ that is used to create
the summary. We wish to know how often the summary's conclusion (the
summarizer $S$) is true within the subset $d$. Within this domain, the
degree of covering can be expressed as the $r_S$ of the
summarizer $S$ of the summary restricted to $d$.
When we generate summaries, we specify time windows to look for
particular trends depending on the protoform type. In this way, the time
window specification is the ``query'' whereas the time window itself is
the subset $d$ of $D$. We already base the ratio $r_S$ off of how often
summarizer $S$ is true in $d$ so it in fact represents the degree of
covering:
\begin{equation}\label{eq:covering}
T_3 = r_S
\end{equation}
where $S$ is the best summarizer for the summary.
Although this degree is just the ratio of the subset $d$ that agrees
with the summary, it is useful to see the actual percentage of agreement
alongside the summary. In cases where the quantifier's definition is
more fuzzy (i.e., the range of the trapezoidal membership function is
especially large), it may be useful to know the exact percentage. For
example, if a summary uses the ``some of the'' quantifier, the trapezoidal
function corresponding to this quantifier ranges from an agreement of
10\% to 50\% of $d$. Even if the quantifier is not guaranteed to be
chosen unless the percentage is between 30\% and 40\% (see
Figure~\ref{fig:trapezoidal}), it is still
useful to have the ability to know what ``some of the'' actually means. 


\paragraph{Degree of Appropriateness $(T_4)$:}
The degree of appropriateness \cite{businessdata} also helps avoid
trivial multivariate summaries. The degree's value represents how interesting and
unexpected a finding in the summary may be. We use this degree to
measure to what extent our summaries follow the maxim of quantity, which
states how much information should be conveyed in a summary. 
When communicating with a human user, it is important that
our summaries avoid providing: 1) too much information to easily process
when reading the summary, or 2) too little information to fully
comprehend the findings implied and to act upon those findings.

To calculate the degree of appropriateness of a summary, the summary is
split into $K$ sub-summaries by attribute. For each sub-summary, the
percentage $r_k$ of the data where the membership value is $\mu_{S_k} > 0$ is calculated, with
$r_k$ given as:
\begin{equation}\label{eq:rk}
r_k = \frac{1}{n}\sum_{i=1}^{n} \mu_{S_k}(y_i)
\end{equation}
Afterwards, the product 
\begin{equation}\label{eq:rproduct}
r^* = \prod_{k=1}^{K} r_k
\end{equation}
of the percentages $r_k$ is calculated. Finally, the absolute difference
between $r^*$ and the summary's degree of covering $T_3$, given as 
\begin{equation}\label{eq:appropriateness}
T_4 = |r^* - T_3|
\end{equation}
yields the degree of appropriateness.
It should be noted that the degree of
appropriateness will be 0 for any univariate summary since this degree
requires relations between two or more variables.

This degree is mainly used to analyze relations in the data. For
example, if a user has a high calorie intake on
50\% of the days and a low carbohydrate intake on 50\% of the days, one
may expect that the user has a high calorie intake and a low
carbohydrate intake on 25\% of the days. This intuition corresponds to
the product of ratios $r^*$ above. If, however, the actual percentage of
days differs from 25\%, then we can say that the outcome is unexpected
and the difference represents the extent to which the outcome differs
from what was expected. 
In terms of the level of informativeness,
if the degree of appropriateness is 0, it is possible that the summary
states too much where the finding is too precise or too little where
the finding is too vague. 

\paragraph{Degree of Coverage $(T_5)$:}
We also want the summaries to be relevant to the user. It would
not be very useful to receive a summary that is not relevant to a
user's context or situation. The maxim of relevance states how relevant
the summary should be. It is calculated by using the degree of coverage
 (not to be confused with the degree of covering), which determines whether the
conclusion made by the summary is supported by enough data
\cite{ifthen}. If the summary is not supported by enough data, then it
may not be worth stating. 

We can use the ratio $r_S$ to find the
percentage of the data that agrees with the summary. 
The degree of coverage~\cite{ifthen} is given as:
\begin{equation}\label{eq:coverage}
T_5 = f(r_S) = \begin{cases}
        0 & r_S \le r_1 \\
        \dfrac{2(r_S - r_1)}{(r_2 - r_1)^2} & r_1 < r_S < \dfrac{r_1 + r_2}{2} \\
        1 - \dfrac{2(r_S - r_1)}{(r_2 - r_1)^2} & \dfrac{r_1 + r_2}{2} \le r_S < r_2 \\
        1 & r_S \ge r_2
    \end{cases}
\end{equation}
In the equation above, \citet{ifthen} use values of 0.02 and 0.15 for
$r_1$ and $r_2$, respectively. The definition of this function creates
a trapezoidal membership function (e.g., see   
Figure
\ref{fig:trapezoidal}). Where $r_S$ lies on this curve determines how
relevant the finding is. Intuitively, this is a fuzzy measure of
covering.

\paragraph{Length Quality $(T_6)$:}
Finally, we want for our summaries to be concise. The maxim of manner states how clear the summary should be.
The more words a user has to read,
the less invested they will be in what the summary means. In addition, the summary
may become more difficult to understand \cite{boran2016overview}. As the way we convey information is
extremely important within the personal health domain, we must evaluate the conciseness
as well as the comprehensiveness of our summaries.
We calculate the length measure \cite{businessdata} as follows:
\begin{equation}\label{eq:length}
T_6 = 2(0.5^{card S})
\end{equation}
In this case, $S$ is the set of summarizers included within the
    summary. This function generates an exponentially decreasing curve
    in the number of summarizers so that, the higher the number of conclusions within a summary, the lower the length
quality. 

\section{Experiments}
\label{sec:exp}
We ran experiments on multiple datasets to analyze the different types
of summaries we generate. In particular, we use real data from the {\em
MyFitnessPal food log dataset} \cite{mfp},
which consists of 587,187 days of food log data across 9,900 users over a
course of up to 180 days. Each entry logs a user's food items with
nutrient information, daily totals, and user's nutrient goals. We also
use user health data from {\em Insight4Wear}
\cite{rawassizadeh_scalable_2016},
which is a quantified-self/life-logging app, with about 11.5 million
records of information. It provides data gathered from mobile devices
that track step count, heart rate, and user activities for around
1,000 users.

For all of the example summaries reported earlier in the paper, we used a
default alphabet size of $n=5$, a time window of seven days, a minimum
support of 20\%, and a minimum confidence of 80\%, which 
comprise the default parameter values. We explore the
use of different sets of input parameters later in this section. Below,
we provide the results of our framework's summary generation on real user
data, and also show quantitative results in terms of evaluation metrics.
It is important to note that existing
systems for summary generation are either not publicly available, or
they do not handle time-series data; therefore, a direct comparison is
not feasible. Nevertheless, we qualitatively showcase how our framework
compares to other state-of-the-art works on temporal data from stock
market and weather domains. On the other hand, for reproducibility, our
implementation is open source, and can be downloaded from
\textbf{\url{https://github.com/harrij15/TemporalSummaries}}.


\subsection{Summary Generation}
We show summaries on calorie and carbohydrate intake from the
MyFitnessPal food log dataset and heart rate data
from Insight4Wear. All summaries are generated using the default input
parameters.

\subsubsection{Calorie and Carbohydrate Intake: MyFitnessPal Food Logs}
Based on the calorie and carbohydrate data shown in
Figures~\ref{fig:caloriedata} and \ref{fig:carbdata},
we display three lists of summaries generated for our user: for
calorie intake (univariate), for carbohydrate intake (univariate),
and for summaries handling both calorie and carbohydrate intake
(multivariate). For each list, there are also corresponding group-level
summaries evaluated on 389 users (15,915 summaries) from the food log
dataset. We selected users that have logged at least 175 days.  In
total, our system generates 113 summaries.

        
        
        
    

\paragraph{Calorie Intake: Univariate Summaries}
With the calorie intake data from Figure~\ref{fig:caloriedata}, we can use the summaries to draw a
picture of how the user usually handles their calories, what kinds of
conclusions we can draw from this data, and how our user compares to
the rest of the study population. Our system produces 19
individual-level summaries using 11 protoforms and 16 group-level
summaries using 5 protoforms. To avoid repetition, 11 representative
individual-level summaries are shown in Table \ref{tab:caloriesummaries}
and 9 group-level summaries are shown in Table
\ref{tab:groupcaloriesummaries}. 

\begin{table}[!ht]
\vspace{-0.1in}
\begin{center}
  \caption{Univariate Individual-Level Summaries for Calorie Intake Data}
  \label{tab:caloriesummaries}
  \scriptsize
  \begin{tabular}{p{1.5in}p{3in}p{0.1in}p{0.1in}p{0.1in}p{0.1in}p{0.1in}p{0.1in}}
    \toprule
    Protoform Type & Summary & $T_1$ & $T_2$ & $T_3$ & $T_4$ & $T_5$ & $T_6$\\\hline
    \midrule
    \texttt{Standard Evaluation (TW)} & In the past full week, your calorie intake has been moderate. & N/A & N/A & 1 & 0 & 1 & 1 \\\hline
    \texttt{Standard Evaluation (sTW)} & On some of the days in the past week, your calorie intake has been low. & 0.93 & 0.81 & 0.29 & 0 & 1 & 1\\\hline
    \texttt{Standard Evaluation + Goal} & On most of the days in the past week, you did not reach your goal to keep your calorie intake low. & 1 & 0.65 & 0.86 & 0 & 1 & 1\\\hline
    \texttt{Comparison} & Your calorie intake was about the same in week 24 as it was in week 12. & N/A & N/A & 1 & 0 & 1 & 1 \\\hline
    \texttt{Comparison + Goal} & You did about the same overall with keeping your calorie intake low in week 24 than you did in week 12. & N/A & N/A & 1 & 0 & 1 & 1\\\hline
    \texttt{Standard Trend} & Half of the time, your calorie intake increases from one day to the next. & 0.71 & 0.84 & 0.53 & 0 & 1 & 1\\\hline
    \texttt{Cluster-Based Pattern} & In week 24, your calorie intake was moderate, then very low, then high, then very high, then low, then moderate. During more than half of the weeks similar to week 24, your calorie intake dropped the next week. & 1 & 1 & 0.6 & 0 & 1 & 0.02\\\hline
    \texttt{Standard Pattern} & The last time you had a week similar to week 24, your calorie intake dropped the next week. & N/A & N/A & 1 & 0 & 1 & 1\\\hline
    \texttt{If-Then Pattern} & There is 100\% confidence that, when your calorie intake follows the pattern of being moderate, your calorie intake tends to be very low the next day. & 1 & 0.68 & 0.32 & N/A & 1 & 0.5\\\hline
    \texttt{Day If-Then Pattern} & There is 100\% confidence that, when your calorie intake follows the pattern of being very high on a Saturday, your calorie intake tends to be very high the next Sunday. & 0.7 & 0.76 & 0.24 & N/A & 1 & 0.5\\\hline
    \texttt{Day-Based Pattern} & Your calorie intake tends to be low on Tuesdays. & 0.9 & 0.81 & 0.28 & 0 & 1 & 1\\\hline
    \texttt{Goal Assistance} & In order to better follow the 2000-calorie diet, you should decrease your calorie intake. & N/A & N/A & N/A & N/A & N/A & 1\\
    \bottomrule
  \end{tabular}
\end{center}
\end{table}

\begin{table}[!ht]
  \caption{Univariate Group-Level Summaries for Calorie Intake Data}
  \label{tab:groupcaloriesummaries}
  \scriptsize
  \begin{tabular}{p{1.5in}p{3in}p{0.1in}p{0.1in}p{0.1in}p{0.1in}p{0.1in}p{0.1in}}
    \toprule
    Protoform Type & Summary & $T_1$ & $T_2$ & $T_3$ & $T_4$ & $T_5$ & $T_6$\\\hline
    \midrule
    \texttt{Standard Evaluation (TW)} & Some of the participants in this study had a moderate calorie intake in the past full week. & 1 & 0.84 & 0.37 & 0 & 1 & 1\\\hline
    \texttt{Standard Evaluation (sTW)} & Almost none of the participants in this study had a high calorie intake on more than half of the days in the past week. & 1 & 0.98 & 0.06 & 0 & 0.21 & 1\\\hline
    \texttt{Standard Evaluation + Goal} & Some of the participants in this study reached their goal to keep their calorie intake low on all of the days in the past week. & 0.84 & 0.87 & 0.42 & 0 & 1 & 1\\\hline
    \texttt{Comparison} & Some of the participants in this study had a higher calorie intake than they did in week 24. & 1 & 1 & 0.3 & 0 & 1 & 1\\\hline
    \texttt{Comparison + Goal} & Some of the participants in this study did not do as well with keeping their calorie intake low as they did in week 24. & 1 & 1 & 0.31 & 0 & 1 & 1\\\hline
    \texttt{Standard Trend} & More than half of the participants in this study increase their calorie intake from one day to the next half of the time. & 0.91 & 1 & 0.59 & 0 & 1 & 1\\\hline
    \texttt{Cluster-Based Pattern} & After looking at clusters containing weeks similar to this past one, it can be seen that some of the participants with these clusters may see a rise in their calorie intake next week. & 1 & 0.67 & 0.34 & 0 & 1 & 1\\\hline
    \texttt{Standard Pattern} & Based on the most recent weeks similar to this past one, it can be seen that half of the participants may see little to no change in their calorie intake next week. & 0.95 & 0.69 & 0.5 & 0 & 1 & 1\\\hline
    \texttt{If-Then Pattern} & For all of the participants in this study, it is true that when their calorie intake follows the pattern of being very high, their calorie intake tends to be high the next day. & 1 & 0 & 1 & 0 & 1 & 0.5\\\hline
    \texttt{Day If-Then Pattern} & For all of the participants in this study, it is true that when their calorie intake follows the pattern of being high on a Tuesday, their calorie intake tends to be moderate on a Wednesday.& 1 & 0 & 1 & 0 & 1 & 0.5\\\hline
    \texttt{Day-Based Pattern} & Some of the participants in this study tend to have a low calorie intake on Mondays. & 0.81 & 1 & 0.26 & 0 & 1 & 1\\\hline
    \texttt{Goal Assistance} & All of the participants in this study have been given advice to decrease their calorie intake. & 1 & 0 & 1 & 0 & 1 & 1\\
    \bottomrule
  \end{tabular}
\end{table}

From the individual-level summaries, it becomes apparent that the user
is struggling with their calorie intake. The standard evaluation (TW)
and goal evaluation summaries explain how our user has struggled in the
past week. We can gather from the comparison and goal comparison
summaries that the user is also performing worse than the week before, so
they are getting further from their health goals. From the standard
trend, the cluster-based pattern, and the standard pattern summaries,
our user can also see that they will most likely do even worse the next
week unless they make changes to their usual routine. In order to make
changes, the user can look at the standard evaluation (sTW), the (day)
if-then pattern, and the day-based pattern summaries to closely
look at their behavioral tendencies and see when they did things right.
Looking at the group-level summaries, the user seems to be performing as
well as some of the other users when comparing the summaries, although
performing fairly worse than the average user. 

When looking at the evaluation metrics in these tables, we can see that
some of the evaluation metrics do not apply to all of the
individual-level summary types (labeled as N/A). For the degrees of
truth $T_1$ and imprecision $T_3$, there are certain individual-level
summary types that do not use a ratio-based method on a subset $d$ of
the dataset $D$. 
The goal assistance summary depends only on the average
value of the last week's values in order to draw conclusions about how
well the user followed a certain diet in the past full week. In light of this,
only the length quality metric is applicable for these summaries.

\paragraph{Carbohydrate Intake: Univariate Summaries}
For carbohydrate intake data from Figure~\ref{fig:carbdata}, our system produces 19 individual-level
summaries using 9 protoforms and 14 group-level summaries using 6
protoforms. 
The conclusions we can draw from the carbohydrate intake are very
similar to what we drew from the calorie intake (using the same
protoforms). At the group level, the user can observe that most of the
other users struggled to reach their daily carbohydrate intake goals in
the past week. We omit the detailed results since they are
qualitatively similar to the calorie intake case. 

\begin{figure}[!ht]
        \centering
        \scalebox{0.9}{
            \begin{tikzpicture}

            \begin{axis}[width=6in,height=2in,axis y line*=left, xlabel=Days,ylabel=Calories (in calories)]
            \addplot[red] table[x=Day,y=Calories,col sep=comma,mark=none]  {calorie_data.csv};
            \end{axis}
            \begin{axis}[width=6in,height=2in,axis y line*=right, hide x axis, ylabel=Carbohydrates (in grams)]
            \addplot[blue] table[x=Day,y=Carbohydrates,col sep=comma,mark=none]  {carb_data.csv};
            \end{axis}
            \end{tikzpicture}
        }
\vspace{-0.1in}
        \caption{Calorie (red) and Carbohydrate (blue) Intake Data
        (superimposed for multivariate analysis).}
        \label{fig:calories}
\vspace{-0.1in}
\end{figure}
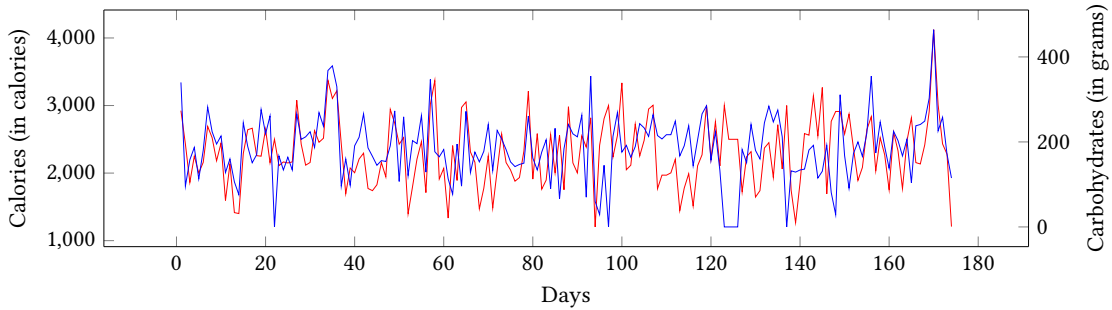

\paragraph{Calorie and Carbohydrate Intake: Multivariate Summaries}
What if there is a correlation between the calorie and carbohydrate
intake (see Figures~\ref{fig:caloriedata} and \ref{fig:carbdata}) for this particular user? We can find out by looking at the
multivariate summaries. 
Figure~\ref{fig:calories} shows both calories and carbohydrates superimposed by day for our
example user. Though it is easy to see that they are correlated, 
it is nevertheless hard to discern common trends and patterns directly
from the raw multivariate time-series data. In contrast, 
for the joint calorie and carbohydrate intake
data, our system produces 27 individual-level summaries using 11
different protoforms and 18 group-level summaries using 8 different
protoforms. Representative individual-level summaries (13 of them) are shown in
Table \ref{tab:caloriecarbsummaries} and group-level summaries (10
of them) are
shown in Table \ref{tab:groupcaloriecarbsummaries}. It appears that our
user is performing much better with carbohydrate intake when compared
to their performance with their calorie intake. The rest
of the users seem to perform well on Mondays.

\begin{table}[!ht]
  \scriptsize
\vspace{-0.1in}
  \caption{Multivariate Individual-Level Summaries for Calorie and Carbohydrate Intake}
  \label{tab:caloriecarbsummaries}
  \begin{tabular}{p{1.2in}p{3in}p{0.15in}p{0.15in}p{0.15in}p{0.15in}p{0.15in}p{0.15in}}
    \toprule
    Protoform Type & Summary & $T_1$ & $T_2$ & $T_3$ & $T_4$ & $T_5$ & $T_6$\\\hline
    \midrule
    \texttt{Standard Evaluation (TW)} & In the past full week, your calorie intake has been moderate and your carbohydrate intake has been moderate. & N/A & N/A & 1 & 0 & 1 & 0.5\\\hline
    \texttt{Standard Evaluation (sTW)} & On some of the days in the past week, your calorie intake has been low and your carbohydrate intake has been high. & 0.93 & 1 & 0.29 & 0.12 & 1 & 0.5\\\hline
    \texttt{Standard Evaluation (sTW) w/ qualifier} & On all of the days in the past week when your calorie intake was very low, your carbohydrate intake was moderate. & 1 & 1 & 1 & 0.96 & 0.99 & 0.5\\\hline
    \texttt{Standard Evaluation + Goal} & On some of the days in the past week, you did not reach your goal to keep your calorie intake low and you reached your goal to keep your carbohydrate intake low. & 0.71 & 1 & 0.43 & 0.06 & 1 & 0.5\\\hline
    \texttt{Comparison} & Your calorie intake was about the same and your carbohydrate intake was about the same in week 24 than they were in week 12. & N/A & N/A & 1 & 0 & 1 & 0.5\\\hline
    \texttt{Comparison + Goal} & You did about the same overall with keeping your calorie intake low and you did about the same overall with keeping your carbohydrate intake low in week 24 than you did in week 12. & N/A & N/A & 1 & 0 & 1 & 0.5\\\hline
    \texttt{Standard Trend} & Some of the time, your calorie intake increases and your carbohydrate intake increases from one day to the next. & 1 & 1 & 0.32 & 0.06 & 1 & 0.5\\\hline
    \texttt{Cluster-Based Pattern} & In week 24, your calorie intake was moderate, then very low, then high, then very high, then low, then moderate and your carbohydrate intake was moderate, then high, then very low, then high. During half of the weeks similar to week 24, your calorie intake dropped and your carbohydrate intake stayed the same the next week. & 1 & 1 & 0.5 & 0.25 & 1 & 0\\\hline
    \texttt{Standard Pattern} & The last time you had a week similar to week 24, your calorie intake dropped and your carbohydrate intake stayed the same the next week. & N/A & N/A & 1 & 0 & 1 & 0.5\\\hline
    \texttt{If-Then Pattern} & There is 100\% confidence that, when your calorie intake follows the pattern of being very high, your calorie intake tends to be very high and your carbohydrate intake tends to be very high the next day. & 0.7 & 0.76 & 0.24 & N/A & 1 & 0.25\\\hline
    \texttt{Day If-Then Pattern} & There is 100\% confidence that, when your calorie intake follows the pattern of being very high on a Saturday, your calorie intake tends to be very high the next Sunday and your carbohydrate intake tends to be very high the next Sunday. & 1 & 0.8 & 0.2 & N/A & 1 & 0.25\\\hline
    \texttt{Day-Based Pattern} & Your calorie intake tends to be very low and your carbohydrate intake tends to be very low on Mondays. & 1 & 1 & 0.04 & 0.02 & 0.05 & 0.5\\\hline
    \texttt{Goal Assistance} & In order to better follow the 2000-calorie diet, you should decrease your calorie intake and increase your carbohydrate intake.& N/A & N/A & N/A & N/A & N/A & 0.5\\
    \bottomrule
  \end{tabular}
\end{table}

\begin{table}[!ht]
  \vspace{-0.1in}
  \scriptsize
  \caption{Multivariate Group-Level Summaries for Calorie and Carbohydrate Intake}
  \label{tab:groupcaloriecarbsummaries}
  \begin{tabular}{p{1in}p{3.3in}p{0.15in}p{0.15in}p{0.15in}p{0.15in}p{0.15in}p{0.15in}}
    \toprule
    Protoform Type & Summary & $T_1$ & $T_2$ & $T_3$ & $T_4$ & $T_5$ & $T_6$\\\hline
    \midrule
    \texttt{Standard Evaluation (TW)} & Almost none of the participants in this study had a very high calorie intake and a high carbohydrate intake in the past full week. & 1 & 0.98 & 0.02 & 0 & 0 & 0.5\\\hline
    \texttt{Standard Evaluation (sTW)} & Almost none of the participants in this study had a very high calorie intake and a high carbohydrate intake on some of the days in the past week. & 1 & 1 & 0.03 & 0 & 0 & 0.5\\\hline
    \texttt{Standard Evaluation (sTW) w/ qualifier} & Some of the participants in this study had a very low carbohydrate intake, when they had a very low calorie intake on all of the days in the past week.  & 0.95 & 1 & 0.29 & 0 & 1 & 0.5\\\hline
    \texttt{Standard Evaluation + Goal} & Some of the participants in this study reached their goal to keep their calorie intake low and did not reach their goal to keep their carbohydrate intake low on all of the days in the past week. & 0.98 & 0.97 & 0.3 & 0 & 1 & 0.5\\\hline
    \texttt{Comparison} & Almost none of the participants in this study had a similar calorie intake and a higher carbohydrate intake in week 11 than they did in week 23. & 1 & 1 & 0.01 & 0 & 0 & 0.5\\\hline
    \texttt{Comparison + Goal} & Almost none of the participants in this study did about the same with keeping their calorie intake low and about the same with keeping their carbohydrate intake low in week 11 as they did in week 23. & 1 & 1 & 0.03 & 0 & 0 & 0.5\\\hline
    \texttt{Standard Trend} & Most of the participants in this study increase their calorie intake and increase their calorie intake from one day to the next some of the time. & 1 & 1 & 0.81 & 0 & 1 & 0.5\\\hline
    \texttt{Cluster-Based Pattern} & After looking at clusters containing weeks similar to this past one, it can be seen that almost none of the participants with these clusters may see a rise in their calorie intake and little to no change in their carbohydrate intake next week. & 1 & 0.91 & 0.16 & 0 & 1 & 0.5\\\hline
    \texttt{Standard Pattern} & Based on the most recent weeks similar to this past one, it can be seen that some of the participants may see little to no change in their calorie intake and little to no change in their carbohydrate intake next week. & 0.72 & 0.91 & 0.24 & 0 & 1 & 0.5\\\hline
    \texttt{If-Then Pattern} & For all of the participants in this study, it is true that when their calorie intake follows the pattern of being low, their calorie intake tends to be moderate and their carbohydrate intake tends to be moderate the next day. & 1 & 0 & 1 & 0 & 1 & 0.125\\\hline
    \texttt{Day If-Then Pattern} & For all of the participants in this study, it is true that when their calorie intake follows the pattern of being high on a Tuesday, their calorie intake tends to be moderate on a Wednesday and their carbohydrate intake tends to be moderate on a Wednesday. & 1 & 0 & 1 & 0 & 1 & 0.125\\\hline
    \texttt{General If-Then Pattern} & For most of the participants in this study, it is true that when they had a very low carbohydrate intake, they had a very low calorie intake. & 1 & 0.79 & 0.78 & 0 & 1 & 0.5\\\hline
    \texttt{Day-Based Pattern} & More than half of the participants in this study tend to have a very low calorie intake and a very low carbohydrate intake on Fridays. & 1 & 1 & 0.66 & 0 & 1 & 0.5\\\hline
    \texttt{Goal Assistance} & More than half of the participants in this study have been given advice to increase their calorie intake. & 1 & 0.93 & 0.7 & 0 & 1 & 1\\
    \bottomrule
  \end{tabular}
\end{table}

\begin{figure}[!ht]
    \centering        
    \begin{tikzpicture}
    \begin{axis}[xlabel=Days,ylabel=Heart Rate (in
        bpm),thick,width=5in,height=2in]
    \addplot table[x=Day,y=Heart Rate,col sep=comma,mark=none]  {hr_data.csv};
    \end{axis}
    \end{tikzpicture}
       
\vspace{-0.1in}
     \caption{Heart Rate Data Snippet (Insight4Wear)}
        \label{fig:heartrate}
\vspace{-0.1in}
\end{figure}
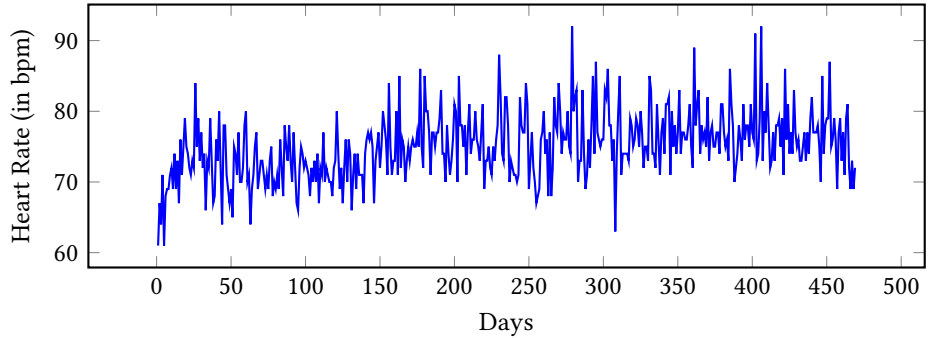

\subsubsection{Heart Rate: Insight4Wear}
Figure~\ref{fig:heartrate} shows a snippet of heart rate data for one
user that spans over 400 days. 
For creating the corresponding categorical sequence, 
SAX binning is not ideal because heart rate data has
very little temporal variation. Looking at the data, the data points are
strictly between 60 and 100 beats per minute (bpm). Due to the lack of
temporal variation, the SAX representations for each granularity will be
heavily affected. The letters chosen for each day or week will make data
points seem to differ greatly when, in reality, the differences are
minimal. For instance, the standard evaluation summaries at the daily
and weekly granularities both state the user's average daily heart rate
to be ``low,'' even though the heart rate is within a healthy range and
only differs slightly from other data points. As heart rate is more
about staying within a healthy range, it is better to create our own
discretization for this particular dataset.
Thus, given that the default equiprobable bins used in
SAX are
actually not ideal for heart rate data, we 
generate meaningful ranges for a heart rate; these are shown
and contrasted with the SAX symbols in Table \ref{tab:heartrate_map}. As
can be seen, a different set of summarizers is used as well.

\begin{table}[!ht]
    \small
\vspace{-0.1in}
  \caption{Mapping of Heart Rate Data Using SAX and Meaningful Ranges}
  \label{tab:heartrate_map}
\vspace{-0.1in}
  \begin{tabular}{cc||cc}
  \toprule
  \multicolumn{2}{c}{SAX} & \multicolumn{2}{c}{Meaningful Ranges}\\
    \toprule
    Symbol & Summarizer  &  Value Range & Summarizer\\
    \midrule
    a & very low & 0 - 50 & abnormally low\\
    b & low & 50 - 60 & low\\
    c  & moderate & 60 - 110 & within range\\
    d & high & 110 - 120 & high\\
    e & very high & 120 and up & abnormally high\\
    \bottomrule
  \end{tabular}
\end{table}

\begin{table}[!ht]
  \vspace{-0.1in}
\vspace{-0.1in}
  \caption{Summaries for Average Daily Heart Rate}
  \label{tab:heartrate2}
  \centering
  \scriptsize
  \begin{tabular}{p{1.5in}p{3in}p{0.1in}p{0.1in}p{0.1in}p{0.1in}p{0.1in}p{0.1in}}
    \toprule
    Protoform Type & Summary & $T_1$ & $T_2$ & $T_3$ & $T_4$ & $T_5$ & $T_6$\\\hline
    \midrule
    \texttt{Standard Evaluation (TW)} & In the past full week, your heart rate has been within range. & N/A & N/A & 1 & 0 & 1 & 1\\\hline
    \texttt{Standard Evaluation (sTW)} & On all of the days in the past week, your heart rate has been within range. & 1 & 1 & 1 & 0 & 1 & 1\\\hline
    \texttt{Standard Evaluation (sTW) + Goal} & On all of the days in the past week, you reached your goal to keep your heart rate within range. & 1 & 1 & 1 & 0 & 1 & 1\\\hline
    \texttt{Comparison} & Your heart rate was lower in week 67 than it was in week 33. & N/A & N/A & 1 & 0 & 1 & 1\\\hline
    \texttt{Comparison + Goal} & You did about the same overall with keeping your heart rate within range in week 67 than you did in week 33. & N/A & N/A & 1 & 0 & 1 & 1\\\hline
    \texttt{Standard Trend} & Half of the time, your heart rate increases from one day to the next. & 0.56 & 0.74 & 0.46 & 0 & 1 & 1\\\hline
    \texttt{Cluster-Based Pattern} & In week 67, your heart rate was within range. During more than half of the weeks similar to week 67, your heart rate dropped the next week. & 1 & 1 & 0.67 & 0 & 1 & 0.5\\\hline
    \texttt{Standard Pattern} & The last time you had a week similar to week 67, your heart rate dropped the next week. & N/A & N/A & 1 & 0 & 1 & 1\\\hline
    \texttt{If-Then Pattern} & There is 100\% confidence that, when your heart rate follows the pattern of being within range, your heart rate tends to be within range the next day. & 1 & 0 & 1 & N/A & 1 & 0.5\\\hline
    \texttt{Day If-Then Pattern} & There is 100\% confidence that, when your heart rate follows the pattern of being within range on a Saturday, your heart rate tends to be within range the next Sunday. & 1 & 0.33 & 0.67 & N/A & 1 & 0.5\\\hline
    \texttt{Day-Based Pattern} & Your heart rate tends to be within range on Wednesdays. & 1 & 1 & 1 & 0 & 1 & 1\\
    \bottomrule
  \end{tabular}
\end{table}

Our system produces 21 summaries using 9 different protoforms, with representative summaries shown in Table \ref{tab:heartrate2}.
Unlike the user for the calorie intake study, this user does not have
trouble satisfying the goal of keeping their heart rate within range.
For example, the if-then pattern summary suggests that, whenever the
user has six days of ``within range'' behavior, the heart rate on the
following day also remains within range. This study also showcases the
robustness of our framework, since we can generate meaningful summaries by
simply changing the symbolic mappings and adjusting the summarizers.

\subsection{Effect of Choosing Different Parameters}
In this section, we will explore different parameter values when
running our system on our user's calorie intake data.

\subsubsection{Time Window}
We tried different time windows to use in order to look for more
patterns. What if our user wished to look at months instead of weeks?
How about the entire time frame? For our earlier experiments, we used a
weekly time window with a daily sub-time window. We re-ran our
experiments for a monthly time window with a daily sub-time window, and
for no time window (where the entire time frame is evaluated).

When we switched to the monthly granularity, the system produced 17
individual-level summaries using 6 protoforms and 13 group-level
summaries using 5 protoforms. 
The change in time window affects every summary type except the standard
trend and the day-based pattern summaries since they do not depend on
the input time window. The output also does not contain
if-then pattern summaries. This is not very surprising since the calorie
intake data contains only six months of data (174 days), and thus there is
not enough data to extract meaningful or frequent monthly patterns.
The results are also very different with the group-level
summaries, although the same summary types are present since the set of
group-level types is derived from the set of individual-level summary
types. As for the summaries themselves, we see a difference in the
conclusion between the standard evaluation (sTW) summaries at the weekly
and monthly granularities; in particular, we observe that the days in the past week
are not very representative of the calorie intake for the entire month.

When we remove the time window, all summaries evaluate the entire time
frame. The system produces 10 summaries using 3 protoforms for both
individual-level and group-level summary output. The only summary types
that work without a time window are the standard evaluation (sTW), goal
evaluation, standard trend, and day-based pattern types. These summary
types can be used for the entire dataset where no time window needs to
be specified. Similar outcomes are observed for
group-level summaries.

\subsubsection{Alphabet Size}
The alphabet size determines the number of letters we use to discretize
the time-series data. The chosen default alphabet size is 5, which
allows our framework to use letters ``a'' through ``e'' in the alphabet.
What if our user wanted their summaries to be more/less precise? When we
change the alphabet size, we may be able to find different patterns as
the data points will be assigned different letters. Additionally, we
will have different sets of summarizers for the summaries we generate.
Table \ref{tab:alphasize} displays the number of individual- and
group-level summaries found using different
alphabet sizes --- 3, 5, and 7.

\begin{table}[!h]
\vspace{-0.1in}
  \caption{Number of Summaries Generated per Alphabet Size}
  \label{tab:alphasize}
\vspace{-0.1in}
  \small
  \centering
  \begin{tabular}{|c|c|c|}
    \toprule
    Alphabet Size & Individual-Level & Group-Level\\\hline
    3 & 22 & 16\\\hline
    5 & 19 & 16\\\hline
    7 & 17 & 16\\
    \bottomrule
  \end{tabular}
\end{table}

Overall, we can see a decrease in the number of individual-level
summaries whenever the alphabet size increases. This may reflect a
decrease in the number of if-then pattern summaries found as more
letters are used to create the symbolic representation.
To select an ideal
alphabet size for the summarization framework, we must consider both the
precision of the symbolic representation of the data as well as the
quality of the summaries we generate. 
We can follow the discussion outlined in \cite{sax}, who evaluate the
alphabet size using the tightness of lower bound metric.
When calculating this metric for each alphabet size
(from 3 to 10), they observed that the bounds are typically weakest when the
alphabet size is smaller. They also explain that it may be intuitive to
use larger alphabet sizes to better represent the data, although there
can be spatial concerns with a larger alphabet size. Therefore, the
recommended range (according to the authors) for a chosen alphabet size
is between 5 and 8 in order to balance the tightness of the lower bound
and the amount of space used. On the other hand, looking at the quality
of our summaries, we typically use a different summarizer for each
different letter we use. It is possible that the meaning of the
summarizers can become increasingly ambiguous as the number of
summarizers rises. Although this has not been fully tested, one can
imagine the difficulty of finding 10 distinct ways to describe a data
point within the range of ``high'' and ``low.'' Combined with the
conclusions from \citet{sax},  choosing
alphabet sizes of 5 or 7 may be the ideal choice for our scenario.

\begin{table}[!ht]
\vspace{-0.1in}
  \caption{Minimum Support and Confidence Thresholds}
  \label{tab:minsupconf}
\vspace{-0.1in}
  \small
  \centering
  \begin{tabular}{|c|c|c|}
    \toprule
    Minimum Support & Minimum Confidence & \# of If-Then Pattern Summaries\\\hline
    \midrule
    0 & 0 & 118\\\hline
    0 & 0.2 & 83\\\hline
    0 & 0.5 & 5\\\hline
    0 & 0.8 & 5\\\hline\hline
    0.2 & 0 & 15\\\hline
    0.2 & 0.2 & 15\\\hline
    0.2 & 0.5 & 3\\\hline
    0.2 & 0.8 & 3\\
    \bottomrule
  \end{tabular}
\end{table}

\subsubsection{Minimum Support and Confidence Thresholds}
These thresholds mainly control the
output of the if-then pattern summaries. 
The default thresholds are
20\% for minimum support and 80\% for minimum confidence. What if our
user wanted patterns that occur more or less frequently? We re-ran
our experiments for different minimum support and confidence thresholds
(e.g., 20\%, 50\%, or 80\%). We did not find any if-then patterns with a
minimum support of 50\% and 80\% and, thus, we show results for the lower
support threshold. Note that a minimum support of 0 means that we consider
all patterns that occur at least once in the data.
As we can see from the results in Table \ref{tab:minsupconf}, all 118 of
the frequent patterns found for our user's calorie intake data in
Figure~\ref{fig:caloriedata}
occur less than half of the time (since no sequences reached the 50\%
threshold). Only 15 sequences surpass the 20\% support threshold while
only 5 sequences surpass the 50\% confidence threshold. For the calorie
intake data in particular, the default thresholds filter out most of the
discovered if-then patterns to just three patterns.

These results beg a question: how many if-then patterns should we
    be generating? Ideally, we should show only some of the if-then
    pattern summaries to a user, since we do not want to overwhelm them.
This implies the need to be able to prioritize and select the if-then
patterns that are best to show to a user. 
At the same time, we believe all relevant patterns should be extracted
and used to create a health profile for the user, which can be useful
for other kinds of analysis.
For example, an if-then pattern that is
infrequent could turn into an anomaly later on that may prove useful to
mention to a user.


\begin{figure}[!htb]
    \centering
	\begin{tikzpicture}
    \begin{axis}[xlabel=Days,ylabel=Close Value,thick,legend pos=outer
        north east,width=5in,height=2in]
    \addplot table[x=Day,y=AAPL Close Value,col sep=comma,mark=none]  {stock_data.csv};
    \addlegendentry{Apple}
    \addplot table[x=Day,y=AET Close Value,col sep=comma,mark=none]  {stock_data.csv};
    \addlegendentry{Aetna}
    \end{axis}
    \end{tikzpicture}
\vspace{-0.1in}
    \caption{Close Value Data for Apple and Aetna}
    \label{fig:aplaet}
\vspace{-0.1in}
\end{figure}
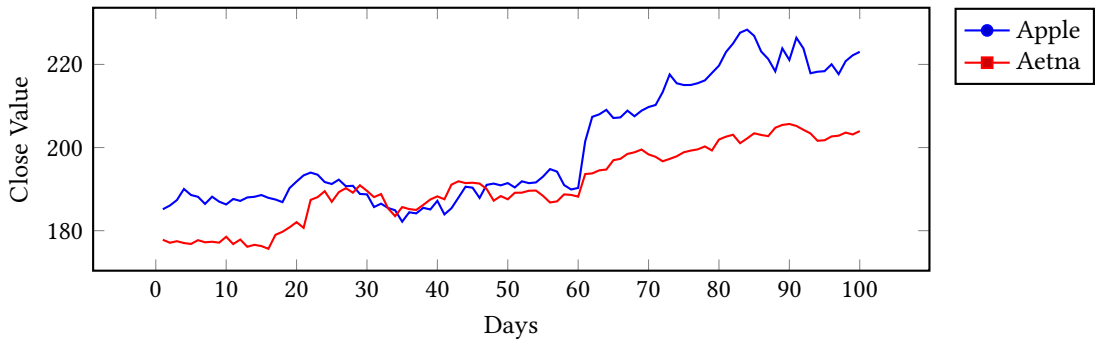

\subsection{Generalizability of our Framework: Application to Weather
and Stock Data}
Although there is a lack of open-source or publicly available 
automatic natural language summarization
systems in the personal health domain, we qualitatively compare our summary output
versus other systems. We decided to look at the stock market \cite{aoki2018} 
and weather \cite{wxsys} domains, which also demonstrates the
generalizability of our approach to time-series data from other domains
besides personal health.

\paragraph{Stock Market Data:} 
In the stock market domain, \citet{aoki2018} extended a neural encoder-decoder
model created by \citet{murakami2017} in order to generate comments
about the Nikkei stock market. They generate summaries
about the general trend of the stock market time-series ticker data,
such as ``\textit{Nikkei turns lower as yen's rise hits exporters}'' and
``\textit{Nikkei Stock Average opens at a high price after Dow Jones
Industrial Average closes at a high price}.'' Their main extension is
the ability to handle multivariate input. For our work, we can apply our
protoform-based approach to stock market data gathered using the REST
API from AlphaVantage \cite{alphavantage}. With this API, we 
retrieved a snippet of 100 days of Apple's and Aetna's stock market
data beginning from May 2018, as plotted in Figure \ref{fig:aplaet}. Our system is able to provide more
insights as shown in Table \ref{tab:stock}. It generated a total of 242
multivariate summaries, which were slightly modified (protoform-wise) to
match the stock market data. We find patterns that cannot be as easily
seen and our summaries say a lot more about the data. The
protoform-based approach also has better performance in terms of how
quickly the summaries are generated.

\begin{table}[!htbp]
\vspace{-0.1in}
  \caption{Apple and Aetna - Stock Market Summaries (AlphaVantage)}
  \label{tab:stock}
  \scriptsize
  \begin{tabular}{p{1.5in}p{3in}p{0.1in}p{0.1in}p{0.1in}p{0.1in}p{0.1in}p{0.1in}}
    \toprule
    Protoform Type & Summary & $T_1$ & $T_2$ & $T_3$ & $T_4$ & $T_5$ & $T_6$\\\hline
    \midrule
    \texttt{Standard Evaluation (TW)} & In the past full week, the AAPL close value has been very high and the AET close value has been very high. & N/A & N/A & 1 & 0 & 1 & 0.5\\\hline
    \texttt{Standard Evaluation (sTW)} & On all of the days in the past week, the AAPL close value has been very high and the AET close value has been very high. & 1 & 1 & 1 & 0 & 1 & 0.5\\\hline
    \texttt{Standard Evaluation (sTW) w/ qualifier} &  On all of the days in the past week when the AAPL close value was very high, the AET close value was very high. & 1 & 1 & 1 & 0 & 1 & 0.5\\\hline
    \texttt{Comparison} & The AAPL close value was higher and the AET close value was higher in week 14 than they were in week 7. & N/A & N/A & 1 & 0 & 1 & 0.5\\\hline
    \texttt{Standard Trend} & Some of the time, the AAPL close value increases and the AET close value increases from one day to the next. & 1 & 1 & 0.34 & 0.004 & 1 & 0.5\\\hline
    \texttt{Cluster-Based Pattern} & In week 14, your AAPL close value was very high and your AET close value was very high. During more than half of the weeks similar to week 14, the AAPL close value stayed the same and the AET close value stayed the same the next week. & 1 & 1 & 0.67 & 0 & 1 & 0.125 \\\hline
    \texttt{Standard Pattern} & The last time you had a week similar to week 14, your AAPL close value stayed the same and your AET close value stayed the same the next week. & N/A & N/A & 1 & 0 & 1 & 0.5 \\\hline
    \texttt{If-Then Pattern} & There is 100\% confidence that, when the AAPL close value follows the pattern of being high, the AET close value tends to be high, then high the next day. & 1 & 0.8 & 0.2 & N/A & 1 & 0.25\\\hline
    \texttt{Day If-Then Pattern} & There is 100\% confidence that, when the AET close value follows the pattern of being very low on a Thursday, the AAPL close value tends to be low the next Friday and the AET close value tends to be very low the next Friday. & 1 & 0.8 & 0.2 & N/A & 1 & 0.125\\\hline
    \texttt{General If-Then Pattern} & In general, if the AAPL close value is very low, then the AET close value is very low. & 0.55 & 1 & 0.45 & 0.41 & 0.7 & 0.5\\\hline
    \texttt{Day-Based Pattern} & the AAPL close value tends to be very low and the AET close value tends to be very low on Mondays. & 1 & 1 & 0.05 & 0.02 & 0.13 & 0.5\\
    \bottomrule
  \end{tabular}
\end{table}
\begin{figure}[!htbp]
    \centering
    \centerline{
        \subfloat[Average Temperature (Daily)]{
        \label{fig:avgtemp}
            \begin{tikzpicture}
            \begin{axis}[name=calorie_plot,xlabel=Days,ylabel= Temp (in
                $^\circ$F),thick,width=3in,height=1.75in]
            \addplot[red] table[x=Day,y=Average Temperature,col sep=comma,mark=none]  {weather_data.csv};
            \end{axis}
            \end{tikzpicture}
        }
        \subfloat[Average Wind Speed (Daily)]{
        \label{fig:avgwind}
            \begin{tikzpicture}
            \begin{axis}[name=calorie_plot,xlabel=Days,ylabel=Wind Speed
                (in MPH),thick,width=3in,height=1.75in]
            \addplot table[x=Day,y=Average Wind Speed,col sep=comma,mark=none]  {weather_data.csv};
            \end{axis}
            \end{tikzpicture}
        }
    }
    \centerline{
        \subfloat[Average Temperature (Hourly)]{
        \label{fig:avgtemphour}
            \begin{tikzpicture}
            \begin{axis}[name=calorie_plot,xlabel=Hours,ylabel=Temp (in
                $^\circ$F),thick,width=3in,height=1.75in]
            \addplot[red] table[x=Day,y=Average Temperature,col sep=comma,mark=none]  {weather_hourly_data.csv};
            \end{axis}
            \end{tikzpicture}
        }
        \subfloat[Average Wind Speed (Hourly)]{
        \label{fig:avgwindhour}
            \begin{tikzpicture}
            \begin{axis}[name=calorie_plot,xlabel=Hours,ylabel=Wind
                Speed (in MPH),thick,width=3in,height=1.75in]
            \addplot table[x=Day,y=Average Wind Speed,col sep=comma,mark=none]  {weather_hourly_data.csv};
            \end{axis}
            \end{tikzpicture}
    }}

\vspace{-0.1in}
    \caption{Average daily temperature and wind speed data for
        Huntsville, AL (for one year and for one day)}
        \label{fig:avgtempwind} 
\vspace{-0.1in}
\end{figure}
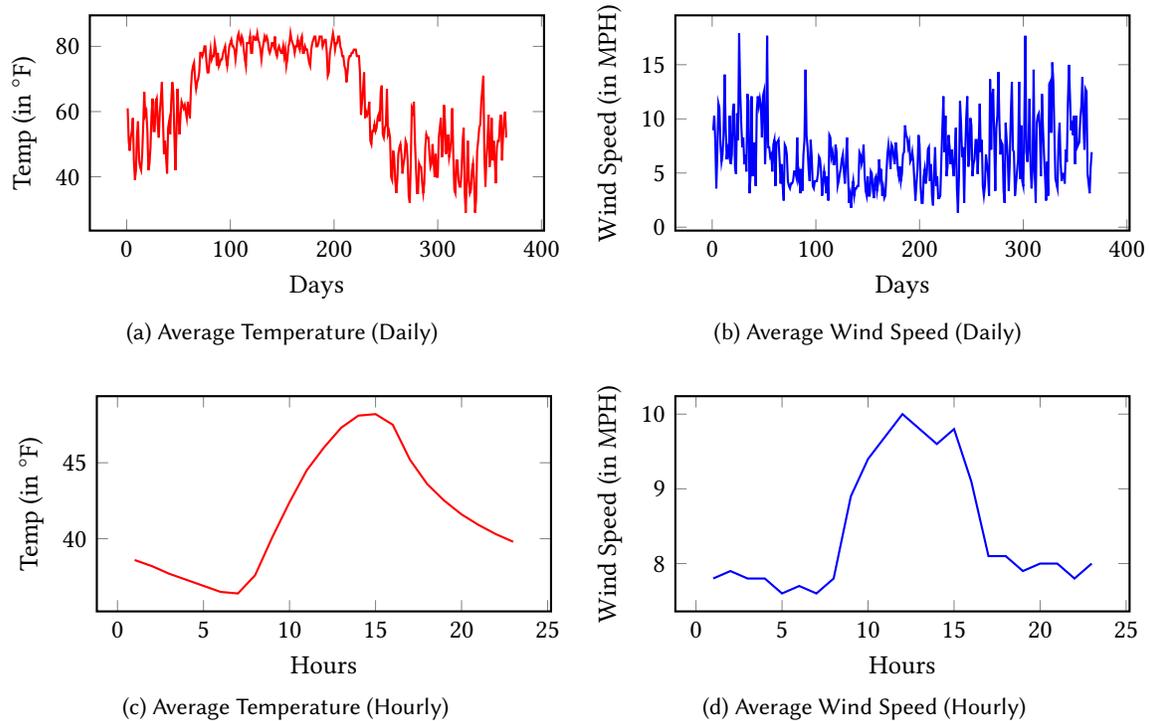

\paragraph{Weather Data:}
In the weather domain, the SUMTIME system \cite{sumtime} proposes a general time-series
summarization model. They focus their efforts on
the weather domain where they describe the forecast of the next
12-24 hours in natural language. An example summary is ``W 8-13
backing SW by mid afternoon and S 10-15 by midnight,'' which describes
wind direction and speed. Using our framework, we
generated
summaries describing the average temperature and the average wind speed
tracked by the weather station at the Huntsville International Airport
in Huntsville, Alabama. This data was provided by the National Centers For
Environmental Information (NCEI) \cite{ncei}. We used two datasets, one
containing a year of daily data between March 1, 2018 and March 1, 2019,
and the other containing a day of hourly data for January 1, 2010. We
display figures for temperature and wind speed daily data in Figures
\ref{fig:avgtemp} and \ref{fig:avgwind}.
Our framework generates 52 summaries at the weekly (TW) granularity, some of which can be found in Table \ref{tab:weather}. 
For the hourly data, we display average temperature and wind speed data
in Figures \ref{fig:avgtemphour} and \ref{fig:avgwindhour}. We generate
11 summaries in total with the multivariate summaries shown below. We
can see that some of the summary types do not show as we stick to mainly
sub-time window conclusions. Some of the protoforms were also modified
to account for the change in granularity.

\begin{table}[!hb]
\vspace{-0.1in}
  \caption{Huntsville, Alabama - Temperature and Wind Speed Summaries (NCEI)}
  \label{tab:weather}
\vspace{-0.1in}
  \footnotesize
  \begin{tabular}{p{1.5in}p{3in}p{0.1in}p{0.1in}p{0.1in}p{0.1in}p{0.1in}p{0.1in}}
    \toprule
    Protoform Type & Summary & $T_1$ & $T_2$ & $T_3$ & $T_4$ & $T_5$ & $T_6$\\\hline
    \midrule
    \texttt{Standard Evaluation (TW)} & In the past full week, the average temperature has been low and the average wind speed has been high. & N/A & N/A & 1 & 0 & 1 & 0.5\\\hline
    \texttt{Standard Evaluation (sTW)} & On some of the days in the past week, the average temperature has been low and the average wind speed has been very low. & 0.93 & 1 & 0.29 & 0.04 & 1 & 0.5\\\hline
    \texttt{Standard Evaluation (sTW) w/ qualifier} &  On all of the days in the past week when the average temperature was very low, the average wind speed was low. & 1 & 1 & 1 & 0.98 & 0.99 & 0.5\\\hline
    \texttt{Comparison} & The average temperature was lower and the average wind speed was higher in week 52 than they were in week 26. & N/A & N/A & 1 & 0 & 1 & 0.5\\\hline
    \texttt{Standard Trend} & Some of the time, the average temperature increases and the average wind speed increases from one day to the next. & 0.88 & 0.94 & 0.28 & 0.04 & 1 & 0.5\\\hline
    \texttt{Cluster-Based Pattern} & In week 52, your average temperature was low, then very low, then low and your average wind speed was very high, then moderate, then very high, then low, then very low. During all of the weeks similar to week 52, the average temperature stayed the same and the average wind speed stayed the same the next week. & 1 & 1 & 1 & 0 & 1 & 0.002\\\hline
    \texttt{Standard Pattern} & The last time there was a week similar to week 52, the average temperature stayed the same and the average wind speed stayed the same the next week. & N/A & N/A & 1 & 0 & 1 & 0.5\\\hline
    \texttt{If-Then Pattern} & There is 100\% confidence that, when the average temperature follows the pattern of being very low, the average temperature tends to be very low and the average wind speed tends to be very high the next day. & 0.92 & 0.79 & 0.21 & N/A & 1 & 0.25\\\hline
    \texttt{Day-Based Pattern} & The average temperature tends to be very low and the average wind speed tends to be very low on Wednesdays. & 1 & 1 & 0.06 & 0.17 & 0.17 & 0.5\\
    \bottomrule
  \end{tabular}
 \end{table}
 \begin{table}[!ht]
  \caption{Huntsville, Alabama - Hourly Temperature and Wind Speed Summaries (NCEI)}
  \label{tab:weatherhourly}
  \vspace{-0.1in}
  \footnotesize
  \begin{tabular}{p{1.5in}p{3in}p{0.1in}p{0.1in}p{0.1in}p{0.1in}p{0.1in}p{0.1in}}
    \toprule
    Protoform Type & Summary & $T_1$ & $T_2$ & $T_3$ & $T_4$ & $T_5$ & $T_6$\\\hline
    \midrule
    \texttt{Standard Evaluation (sTW)} & During almost none of the hours in the past day, the average temperature was very low and the average wind speed was very low. & 1 & 1 & 0.43 & 0.39 & 0.95 & 0.5\\\hline
    \texttt{Standard Evaluation (sTW) w/ qualifier} &  During most of the hours in the past day, when the average temperature was low, the average wind speed was low. & 1 & 1 & 0 & 0.09 & 0.95 & 0.5\\\hline
    \texttt{Comparison} & The average temperature was lower and the average wind speed was lower in hour 23 than they were in hour 11. & N/A & N/A & 1 & 0 & 1 & 0.5\\\hline
    \texttt{Standard Trend} & During some of the day, the average temperature decreases and the average wind speed decreases from one hour to the next. & 1 & 1 & 0.3 & 0.07 & 1 & 0.5\\\hline
    \bottomrule
  \end{tabular}
\end{table}

These two examples showcase the generalizability of our time-series
summary generation framework. They demonstrate that the framework is
generic and can be applied to different domains with slight changes to
the protoforms and the vocabulary (e.g., quantifiers, summarizers).

\subsection{Evaluation via User Study}

Having looked at various types of summaries produced by our system,
along with quantitative metrics, we wanted to evaluate the efficacy of
the summaries in terms of the understandability (readability and
comprehensiveness), the usefulness of our output summaries,
and how well they align with the source data or provenance.

To perform this evaluation, we designed a small user study comprised of 11
participants (8 male and 3 female) aged 21-25. As this user study is preliminary,
we only excluded potential recruits who were too familiar with our work. All participants 
had varying levels of fluency in English and knowledge pertaining to personal health. For our study, the
participants were asked 
to evaluate our summaries on three subjective metrics:
readability/comprehensiveness, usefulness, and data alignment. Each
individual completed a unique survey (i.e., each survey used a different
dataset) where they were first presented with a single time-series chart
of calorie intake data, similar to Figure \ref{fig:caloriedata}, and a
scenario. The scenario matches the running example within the paper
where a user is following a 2000-calorie diet. The horizontal colored
ranges were also described to the user. 
In fact, all of the figures presented above correspond to similar charts
and provenance shown to the users for various protoforms.
After reading the scenario, each
participant was asked to provide a textual description of the chart. We
requested this preliminary description to analyze what patterns
non-experts typically look for and how they describe those patterns.
After the description was provided, each participant was shown a number
of representative summaries generated for the calorie intake data, as
well as corresponding charts displaying the provenance of the discovered
patterns. Presented with one summary at a time, each participant was
asked to evaluate each summary over the three aforementioned metrics.
Each metric can be given a score from 1 (strongly disagree) to 5
(strongly agree) by the user based on their agreement with the following
statements:

\begin{description}
    \item (1) This summary is readable and comprehensible.
    
    \item (2) This summary is useful to me and my goals.
    
    \item (3) This summary aligns well with the data.
\end{description}

After evaluating the univariate summaries, the participants were
asked to provide another description of the data, to capture how the
freeform responses change after having seen our summaries.
Finally, another
variable was added in to display multivariate summaries and the process
was repeated. Each participant evaluated 14 summaries in total and was
exposed to most, if not all, of the summary types. Some participants did
not receive every summary type if the representative summary for the
summary type (chosen using a weighted sum of the objective evaluation
metrics $T_1$--$T_6$ from Section~\ref{metrics}) was determined to have been redundant or trivial. In this case, an
additional representative summary from another summary type served
as a replacement. 

The results for our user study are presented in Table
    \ref{tab:human_eval}. 
    We show the overall averages of
    the scores given to the subjective metrics (displayed at the
    bottom), as well as the average scores within the summary types.
Looking at the results, we can see that the participants had a fairly
strong agreement overall with the readability and the comprehensiveness
of our summaries at a score of 4.13 out of 5. As for usefulness and data
alignment, there is still agreement on average when it comes to
the usefulness of our summaries (3.48 out of 5) and how well they align
with the presented charts (3.58 out of 5). 
When looking at the averages within the summary types, it can be
seen that the goal evaluation summaries had the best scores in all three
categories. 
The standard trends summaries scored lowest on usefulness, although they have high scores
in the other categories. 
Over all of the three aspects, the cluster-based pattern
summaries received the lowest scores. We believe that the description of
the week before the actual summary may make it too complex of a read. It
is also possible that the pattern itself is not easily comprehensible.
The if-then pattern summaries also receive
lower scores in usefulness and data alignment, which may also stem from
pattern complexity. 
These results suggest that while our framework generates understandable
summaries that are useful, and there is great value in displaying the
provenance for illustrating the supporting data, there is still scope
for further improvement of the more complex summary types like the
cluster-based and if-then patterns, both in terms of the natural
language text and the accompanying provenance charts. This will provide
fruitful directions for our future work.


We also received a lot of helpful descriptions and feedback from the
participants that we will use to improve our system in the
future. From their descriptions, we are able to determine possible
future patterns to find in time-series data, such as variance,
consistency, and general drops/rises. It was interesting to find that
some of the participants' initial descriptions aligned with patterns our
summaries describe, such as the standard evaluation (w/ qualifier)
summaries.

\begin{table}[!h]
\vspace{-0.1in}
  \caption{Human Evaluation Scores}
  \label{tab:human_eval}
  \small
  \centering
  \begin{tabular}{|c|c|c|c|}
    \toprule
    \textbf{Summary Type} & \textbf{\scriptsize Readability/Comprehensiveness} & \textbf{Usefulness} & \textbf{Data Alignment}\\\hline
    Standard Evaluation (TW granularity) & 3.9 & 3.55 & 3.09\\\hline
    Standard Evaluation (sTW granularity) & 4.42 & 3.75 & 4\\\hline
    Standard Evaluation (sTW granularity w/ qualifier) & 4.29 & 3.65 & 4.53\\\hline
    Evaluation Comparison & 4 & 3.67 & 3.92\\\hline
    Goal Comparison & 4 & 3.73 & 4.09\\\hline
    Goal Evaluation & 4.55 & 4.45 & 4.18\\\hline
    Standard Trends & 4.5 & 2.5 & 4\\\hline
    Cluster-Based Pattern
	 & 2.73 & 2.91 & 2.55\\\hline
    Standard Pattern
	 & 4.5 & 3.38 & 3.38\\\hline
    If-Then Pattern
	 & 4.21 & 2.86 & 2.07\\\hline
    Day If-Then Pattern
	 & 3.82 & 3.18 & 3\\\hline
    General If-Then Pattern
	 & 4.33 & 3.4 & 3.8\\\hline
    Goal Assistance
	 & 4.36 & 3.36 & 4\\\hline
    Day-Based Pattern
	 & 4.5 & 4 & 3.33\\\hline\hline
	Overall & 4.13 & 3.48 & 3.58\\

    \bottomrule
  \end{tabular}
\end{table}

\section{Conclusion}
We presented a system to automatically generate summaries of a
user's personal health data. Unlike most previous approaches that either
focus on tabular, textual, or relatively simple trend summaries, we mine
interesting patterns from symbolic representations of numeric temporal
data, and propose a comprehensive set of useful summaries that cover a
wide range of scenarios. We showcase our work using real user data. Our
system is designed to extract comprehensible summaries to
better guide users towards their goals. To our knowledge, this is the
first comprehensive and systematic approach to generate natural language
summaries from time-series personal health data via protoforms. There is
\textit{no current system} that can automatically extract patterns and
clusters from time-series data and present them to the user in an
explainable manner in natural language. In fact, our approach is also \textit{generic} and
\textit{extensible} to other domains outside of the personal health domain.


It is important to note that our main contribution in this paper is the 
comprehensive framework for the generation of
useful and informative summaries of time-series data. 
We conducted a preliminary user study that confirm that our approach is
indeed effective and useful. However, it also elucidated aspects that
need improvement.
In the future, we aim to 
analyze how our
summaries ultimately impact the behavior of users via a larger user study.
This will allow us to focus on protoforms that are the most interesting and helpful to users,
along with the most comprehensible ways to put these findings into
words.

One limitation of our work is that our system is protoform- or
template-based.
In the future, we seek to automate the summarization process
where the use of protoforms is no longer needed, while retaining the system
 efficiency and summary readability. 
For example, we can extract temporal shapes and relationships 
to better summarize a time series by describing where interesting shapes
(e.g., certain spikes or drops) occur within and across time series.
These descriptions could be seen as creating a narrative about a time
series within a specific window when applied to the personal health
domain. We also aim to utilize deep learning to automatically
generate summaries based on the shapes found in the time series. 
\begin{acks}
This work is supported by IBM Research AI through the AI
Horizons Network, and was conducted under the auspices of the 
RPI-IBM Health Empowerment
through Analytics Learning and Semantics (HEALS) project.
\end{acks}


\bibliographystyle{ACM-Reference-Format}
\bibliography{acm19}


\begin{thebibliography}{52}


\ifx \showCODEN    \undefined \def \showCODEN     #1{\unskip}     \fi
\ifx \showDOI      \undefined \def \showDOI       #1{#1}\fi
\ifx \showISBNx    \undefined \def \showISBNx     #1{\unskip}     \fi
\ifx \showISBNxiii \undefined \def \showISBNxiii  #1{\unskip}     \fi
\ifx \showISSN     \undefined \def \showISSN      #1{\unskip}     \fi
\ifx \showLCCN     \undefined \def \showLCCN      #1{\unskip}     \fi
\ifx \shownote     \undefined \def \shownote      #1{#1}          \fi
\ifx \showarticletitle \undefined \def \showarticletitle #1{#1}   \fi
\ifx \showURL      \undefined \def \showURL       {\relax}        \fi
\providecommand\bibfield[2]{#2}
\providecommand\bibinfo[2]{#2}
\providecommand\natexlab[1]{#1}
\providecommand\showeprint[2][]{arXiv:#2}

\bibitem[\protect\citeauthoryear{Alvarez-Alvarez and Trivino}{Alvarez-Alvarez
  and Trivino}{2013}]%
        {gait}
\bibfield{author}{\bibinfo{person}{Alberto Alvarez-Alvarez} {and}
  \bibinfo{person}{Gracian Trivino}.} \bibinfo{year}{2013}\natexlab{}.
\newblock \showarticletitle{Linguistic description of the human gait quality}.
\newblock \bibinfo{journal}{\emph{Engineering Applications of Artificial
  Intelligence}} \bibinfo{volume}{26}, \bibinfo{number}{1}
  (\bibinfo{year}{2013}), \bibinfo{pages}{13 -- 23}.
\newblock
\showISSN{0952-1976}


\bibitem[\protect\citeauthoryear{Aoki, Miyazawa, Ishigaki, Goshima, Aoki,
  Kobayashi, Takamura, and Miyao}{Aoki et~al\mbox{.}}{2018}]%
        {aoki2018}
\bibfield{author}{\bibinfo{person}{Tatsuya Aoki}, \bibinfo{person}{Akira
  Miyazawa}, \bibinfo{person}{Tatsuya Ishigaki}, \bibinfo{person}{Keiichi
  Goshima}, \bibinfo{person}{Kasumi Aoki}, \bibinfo{person}{Ichiro Kobayashi},
  \bibinfo{person}{Hiroya Takamura}, {and} \bibinfo{person}{Yusuke Miyao}.}
  \bibinfo{year}{2018}\natexlab{}.
\newblock \showarticletitle{Generating Market Comments Referring to External
  Resources}. In \bibinfo{booktitle}{\emph{International Conference on Natural
  Language Generation}}.
\newblock


\bibitem[\protect\citeauthoryear{Association et~al\mbox{.}}{Association
  et~al\mbox{.}}{2019}]%
        {ada}
\bibfield{author}{\bibinfo{person}{American~Diabetes Association}
  {et~al\mbox{.}}} \bibinfo{year}{2019}\natexlab{}.
\newblock \showarticletitle{5. Lifestyle management: standards of medical care
  in diabetes—2019}.
\newblock \bibinfo{journal}{\emph{Diabetes Care}} \bibinfo{volume}{42},
  \bibinfo{number}{Supplement 1} (\bibinfo{year}{2019}),
  \bibinfo{pages}{S46--S60}.
\newblock


\bibitem[\protect\citeauthoryear{Baldwin, Martin, and Rossiter}{Baldwin
  et~al\mbox{.}}{1998}]%
        {baldwin}
\bibfield{author}{\bibinfo{person}{James Baldwin}, \bibinfo{person}{Trevor~P.
  Martin}, {and} \bibinfo{person}{Jonathan~M. Rossiter}.}
  \bibinfo{year}{1998}\natexlab{}.
\newblock \showarticletitle{Time series modelling and prediction using fuzzy
  trend information}.
\newblock \bibinfo{journal}{\emph{International Conference on Soft Computing
  and Information Intelligent Systems}} (\bibinfo{year}{1998}).
\newblock


\bibitem[\protect\citeauthoryear{Batyrshin and Sheremetov}{Batyrshin and
  Sheremetov}{2008}]%
        {BATYRSHIN2008}
\bibfield{author}{\bibinfo{person}{Ildar~Z. Batyrshin} {and}
  \bibinfo{person}{Leonid~B. Sheremetov}.} \bibinfo{year}{2008}\natexlab{}.
\newblock \showarticletitle{Perception-based approach to time series data
  mining}.
\newblock \bibinfo{journal}{\emph{Applied Soft Computing}} \bibinfo{volume}{8},
  \bibinfo{number}{3} (\bibinfo{year}{2008}), \bibinfo{pages}{1211 -- 1221}.
\newblock
\showISSN{1568-4946}
\newblock
\shownote{Forging the Frontiers -- Soft Computing.}


\bibitem[\protect\citeauthoryear{Boran, Akay, and Yager}{Boran
  et~al\mbox{.}}{2016}]%
        {boran2016overview}
\bibfield{author}{\bibinfo{person}{Fatih~Emre Boran}, \bibinfo{person}{Diyar
  Akay}, {and} \bibinfo{person}{Ronald~R Yager}.}
  \bibinfo{year}{2016}\natexlab{}.
\newblock \showarticletitle{An overview of methods for linguistic summarization
  with fuzzy sets}.
\newblock \bibinfo{journal}{\emph{Expert Systems with Applications}}
  \bibinfo{volume}{61} (\bibinfo{year}{2016}), \bibinfo{pages}{356--377}.
\newblock


\bibitem[\protect\citeauthoryear{Castillo-Ortega, Mar\'{i}n, S\'{a}nchez, and
  Tettamanzi}{Castillo-Ortega et~al\mbox{.}}{2011}]%
        {genetic}
\bibfield{author}{\bibinfo{person}{Rita Castillo-Ortega},
  \bibinfo{person}{Nicol\'{a}s Mar\'{i}n}, \bibinfo{person}{Daniel
  S\'{a}nchez}, {and} \bibinfo{person}{Andrea Tettamanzi}.}
  \bibinfo{year}{2011}\natexlab{}.
\newblock \showarticletitle{Linguistic Summarization of Time Series Data using
  Genetic Algorithms}. In \bibinfo{booktitle}{\emph{Conf.\ of the European
  Society for Fuzzy Logic and Technology}}.
\newblock


\bibitem[\protect\citeauthoryear{Cheung and Stephanopoulos}{Cheung and
  Stephanopoulos}{1990}]%
        {cheung1}
\bibfield{author}{\bibinfo{person}{Jarvis T.-Y. Cheung} {and}
  \bibinfo{person}{George Stephanopoulos}.} \bibinfo{year}{1990}\natexlab{}.
\newblock \showarticletitle{Representation of process trends -- {Part I}. {A}
  formal representation framework}.
\newblock \bibinfo{journal}{\emph{Computers \& Chemical Engineering}}
  \bibinfo{volume}{14}, \bibinfo{number}{4} (\bibinfo{year}{1990}),
  \bibinfo{pages}{495 -- 510}.
\newblock
\showISSN{0098-1354}


\bibitem[\protect\citeauthoryear{Choe, Lee, Lee, Pratt, and Kientz}{Choe
  et~al\mbox{.}}{2014}]%
        {choequantifiedself}
\bibfield{author}{\bibinfo{person}{Eun~Kyoung Choe}, \bibinfo{person}{Nicole~B.
  Lee}, \bibinfo{person}{Bongshin Lee}, \bibinfo{person}{Wanda Pratt}, {and}
  \bibinfo{person}{Julie~A. Kientz}.} \bibinfo{year}{2014}\natexlab{}.
\newblock \showarticletitle{Understanding quantified-selfers' practices in
  collecting and exploring personal data}. In \bibinfo{booktitle}{\emph{ACM
  Conference on Human Factors in Computing Systems}}.
\newblock


\bibitem[\protect\citeauthoryear{Codella, Partovian, Chang, and Chen}{Codella
  et~al\mbox{.}}{2018}]%
        {codella2018}
\bibfield{author}{\bibinfo{person}{James Codella}, \bibinfo{person}{Chohreh
  Partovian}, \bibinfo{person}{Hung-Yang Chang}, {and}
  \bibinfo{person}{Ching-Hua Chen}.} \bibinfo{year}{2018}\natexlab{}.
\newblock \showarticletitle{Data quality challenges for person-generated health
  and wellness data}.
\newblock \bibinfo{journal}{\emph{IBM Journal of Research and Development}}
  \bibinfo{volume}{62}, \bibinfo{number}{1} (\bibinfo{date}{Jan}
  \bibinfo{year}{2018}), \bibinfo{pages}{3:1--3:8}.
\newblock


\bibitem[\protect\citeauthoryear{Conde-Clemente, Alonso, Éldman O.~Nunes,
  Sanchez, and Trivino}{Conde-Clemente et~al\mbox{.}}{2017}]%
        {deforestation}
\bibfield{author}{\bibinfo{person}{Patricia Conde-Clemente},
  \bibinfo{person}{Jose~M. Alonso}, \bibinfo{person}{Éldman O.~Nunes},
  \bibinfo{person}{Angel Sanchez}, {and} \bibinfo{person}{Gracian Trivino}.}
  \bibinfo{year}{2017}\natexlab{}.
\newblock \showarticletitle{New types of computational perceptions: Linguistic
  descriptions in deforestation analysis}.
\newblock \bibinfo{journal}{\emph{Expert Systems with Applications}}
  \bibinfo{volume}{85} (\bibinfo{year}{2017}), \bibinfo{pages}{46 -- 60}.
\newblock
\showISSN{0957-4174}


\bibitem[\protect\citeauthoryear{Das, Lin, Mannila, Renganathan, and Smyth}{Das
  et~al\mbox{.}}{1998}]%
        {Das}
\bibfield{author}{\bibinfo{person}{Gautam Das}, \bibinfo{person}{King-Ip Lin},
  \bibinfo{person}{Heikki Mannila}, \bibinfo{person}{Gopal Renganathan}, {and}
  \bibinfo{person}{Padhraic Smyth}.} \bibinfo{year}{1998}\natexlab{}.
\newblock \showarticletitle{Rule Discovery from Time Series}. In
  \bibinfo{booktitle}{\emph{ACM SIGKDD Conference on Knowledge Discovery and
  Data Mining}}.
\newblock


\bibitem[\protect\citeauthoryear{Eciolaza, Pereira-Fariña, and
  Trivino}{Eciolaza et~al\mbox{.}}{2013}]%
        {driving}
\bibfield{author}{\bibinfo{person}{Luka Eciolaza}, \bibinfo{person}{Martín
  Pereira-Fariña}, {and} \bibinfo{person}{Gracian Trivino}.}
  \bibinfo{year}{2013}\natexlab{}.
\newblock \showarticletitle{Automatic linguistic reporting in driving
  simulation environments}.
\newblock \bibinfo{journal}{\emph{Applied Soft Computing}}
  \bibinfo{volume}{13}, \bibinfo{number}{9} (\bibinfo{year}{2013}),
  \bibinfo{pages}{3956 -- 3967}.
\newblock
\showISSN{1568-4946}


\bibitem[\protect\citeauthoryear{Elsworth and Güttel}{Elsworth and
  Güttel}{2020}]%
        {abba}
\bibfield{author}{\bibinfo{person}{Steven Elsworth} {and}
  \bibinfo{person}{Stefan Güttel}.} \bibinfo{year}{2020}\natexlab{}.
\newblock \showarticletitle{ABBA: Adaptive Brownian bridge-based symbolic
  aggregation of time series}.
\newblock \bibinfo{journal}{\emph{Data Mining and Knowledge Discovery}}
  \bibinfo{volume}{34} (\bibinfo{year}{2020}), \bibinfo{pages}{1175--1200}.
\newblock


\bibitem[\protect\citeauthoryear{Gatt, Portet, Reiter, Hunter, Mahamood,
  Moncur, and Sripada}{Gatt et~al\mbox{.}}{2009}]%
        {gatt}
\bibfield{author}{\bibinfo{person}{Albert Gatt}, \bibinfo{person}{Fran\c{c}ois
  Portet}, \bibinfo{person}{Ehud Reiter}, \bibinfo{person}{Jim Hunter},
  \bibinfo{person}{Saad Mahamood}, \bibinfo{person}{Wendy Moncur}, {and}
  \bibinfo{person}{Somayajulu Sripada}.} \bibinfo{year}{2009}\natexlab{}.
\newblock \showarticletitle{From Data to Text in the Neonatal Intensive Care
  Unit: Using NLG Technology for Decision Support and Information Management}.
\newblock \bibinfo{journal}{\emph{AI Commun.}} \bibinfo{volume}{22},
  \bibinfo{number}{3} (\bibinfo{date}{Aug.} \bibinfo{year}{2009}),
  \bibinfo{pages}{153--186}.
\newblock
\showISSN{0921-7126}


\bibitem[\protect\citeauthoryear{Grice}{Grice}{1967}]%
        {grice}
\bibfield{author}{\bibinfo{person}{Herbert~Paul Grice}.}
  \bibinfo{year}{1967}\natexlab{}.
\newblock \showarticletitle{Logic and Conversation}.
\newblock In \bibinfo{booktitle}{\emph{Studies in the Way of Words}},
  \bibfield{editor}{\bibinfo{person}{Paul Grice}} (Ed.).
  \bibinfo{publisher}{Harvard University Press}, \bibinfo{pages}{41--58}.
\newblock


\bibitem[\protect\citeauthoryear{Guimar{\~a}es and Ultsch}{Guimar{\~a}es and
  Ultsch}{1999}]%
        {temporalkc}
\bibfield{author}{\bibinfo{person}{Gabriela Guimar{\~a}es} {and}
  \bibinfo{person}{Alfred Ultsch}.} \bibinfo{year}{1999}\natexlab{}.
\newblock \showarticletitle{A Method for Temporal Knowledge Conversion}. In
  \bibinfo{booktitle}{\emph{Advances in Intelligent Data Analysis}},
  \bibfield{editor}{\bibinfo{person}{David~J. Hand}, \bibinfo{person}{Joost~N.
  Kok}, {and} \bibinfo{person}{Michael~R. Berthold}} (Eds.).
  \bibinfo{pages}{369--380}.
\newblock
\showISBNx{978-3-540-48412-7}


\bibitem[\protect\citeauthoryear{He, Xu, and Deng}{He et~al\mbox{.}}{2002}]%
        {squeezer}
\bibfield{author}{\bibinfo{person}{Zengyou He}, \bibinfo{person}{Xiaofei Xu},
  {and} \bibinfo{person}{Shengchun Deng}.} \bibinfo{year}{2002}\natexlab{}.
\newblock \showarticletitle{Squeezer: An efficient algorithm for clustering
  categorical data}.
\newblock \bibinfo{journal}{\emph{Journal of Computer Science and Technology}}
  \bibinfo{volume}{17} (\bibinfo{date}{09} \bibinfo{year}{2002}),
  \bibinfo{pages}{611--624}.
\newblock


\bibitem[\protect\citeauthoryear{H{\"o}ppner}{H{\"o}ppner}{2001}]%
        {Hoppner}
\bibfield{author}{\bibinfo{person}{Frank H{\"o}ppner}.}
  \bibinfo{year}{2001}\natexlab{}.
\newblock \showarticletitle{Learning Temporal Rules from State Sequences}. In
  \bibinfo{booktitle}{\emph{IJCAI Workshop on Learning from Temporal and
  Spatial Data}}.
\newblock


\bibitem[\protect\citeauthoryear{Kacprzyk and Wilbik}{Kacprzyk and
  Wilbik}{2008}]%
        {trends}
\bibfield{author}{\bibinfo{person}{Janusz Kacprzyk} {and} \bibinfo{person}{Anna
  Wilbik}.} \bibinfo{year}{2008}\natexlab{}.
\newblock \showarticletitle{Linguistic summarization of time series using fuzzy
  logic with linguistic quantifiers: a truth and specificity based approach}.
  In \bibinfo{booktitle}{\emph{International Conference on Artificial
  Intelligence and Soft Computing}}. \bibinfo{pages}{241--252}.
\newblock


\bibitem[\protect\citeauthoryear{Kacprzyk, Wilbik, and Zadrozny}{Kacprzyk
  et~al\mbox{.}}{2008}]%
        {kacprzyk2008linguistic}
\bibfield{author}{\bibinfo{person}{Janusz Kacprzyk}, \bibinfo{person}{Anna
  Wilbik}, {and} \bibinfo{person}{Slawomir Zadrozny}.}
  \bibinfo{year}{2008}\natexlab{}.
\newblock \showarticletitle{Linguistic summarization of time series using a
  fuzzy quantifier driven aggregation}.
\newblock \bibinfo{journal}{\emph{Fuzzy Sets and Systems}}
  \bibinfo{volume}{159}, \bibinfo{number}{12} (\bibinfo{year}{2008}),
  \bibinfo{pages}{1485--1499}.
\newblock


\bibitem[\protect\citeauthoryear{Kacprzyk, Wilbik, and Zadrozny}{Kacprzyk
  et~al\mbox{.}}{2010}]%
        {trendsapproach}
\bibfield{author}{\bibinfo{person}{Janusz Kacprzyk}, \bibinfo{person}{Anna
  Wilbik}, {and} \bibinfo{person}{Slawomir Zadrozny}.}
  \bibinfo{year}{2010}\natexlab{}.
\newblock \showarticletitle{An Approach to the Linguistic Summarization of Time
  Series Using a Fuzzy Quantifier Driven Aggregation}.
\newblock \bibinfo{journal}{\emph{Int. J. Intell. Syst.}} \bibinfo{volume}{25},
  \bibinfo{number}{5} (\bibinfo{date}{May} \bibinfo{year}{2010}),
  \bibinfo{pages}{411--439}.
\newblock
\showISSN{0884-8173}


\bibitem[\protect\citeauthoryear{Kacprzyk, Yager, and Zadrozny}{Kacprzyk
  et~al\mbox{.}}{2002}]%
        {businessdata}
\bibfield{author}{\bibinfo{person}{Janusz Kacprzyk}, \bibinfo{person}{Ronald~R.
  Yager}, {and} \bibinfo{person}{Slawomir Zadrozny}.}
  \bibinfo{year}{2002}\natexlab{}.
\newblock \bibinfo{booktitle}{\emph{Fuzzy Linguistic Summaries of Databases for
  an Efficient Business Data Analysis and Decision Support}}.
\newblock \bibinfo{publisher}{Springer US}, \bibinfo{address}{Boston, MA},
  \bibinfo{pages}{129--152}.
\newblock
\showISBNx{978-0-306-46991-6}
\urldef\tempurl%
\url{https://doi.org/10.1007/0-306-46991-X_6}
\showDOI{\tempurl}


\bibitem[\protect\citeauthoryear{Kaczmarek-Majer and
  Hryniewicz}{Kaczmarek-Majer and Hryniewicz}{2019}]%
        {forecast}
\bibfield{author}{\bibinfo{person}{Katarzyna Kaczmarek-Majer} {and}
  \bibinfo{person}{Olgierd Hryniewicz}.} \bibinfo{year}{2019}\natexlab{}.
\newblock \showarticletitle{Application of linguistic summarization methods in
  time series forecasting}.
\newblock \bibinfo{journal}{\emph{Information Sciences}}  \bibinfo{volume}{478}
  (\bibinfo{year}{2019}), \bibinfo{pages}{580 -- 594}.
\newblock
\showISSN{0020-0255}


\bibitem[\protect\citeauthoryear{Klein, Kim, Deng, Senellart, and Rush}{Klein
  et~al\mbox{.}}{2017}]%
        {klein}
\bibfield{author}{\bibinfo{person}{Guillaume Klein}, \bibinfo{person}{Yoon
  Kim}, \bibinfo{person}{Yuntian Deng}, \bibinfo{person}{Jean Senellart}, {and}
  \bibinfo{person}{Alexander~M. Rush}.} \bibinfo{year}{2017}\natexlab{}.
\newblock \showarticletitle{OpenNMT: Open-Source Toolkit for Neural Machine
  Translation}.
\newblock \bibinfo{journal}{\emph{CoRR}}  \bibinfo{volume}{abs/1701.02810}
  (\bibinfo{year}{2017}).
\newblock
\showeprint[arxiv]{1701.02810}


\bibitem[\protect\citeauthoryear{Koehn, Hoang, Birch, Callison-Burch, Federico,
  Bertoldi, Cowan, Shen, Moran, Zens, Dyer, Bojar, Constantin, and
  Herbst}{Koehn et~al\mbox{.}}{2007}]%
        {koehn}
\bibfield{author}{\bibinfo{person}{Philipp Koehn}, \bibinfo{person}{Hieu
  Hoang}, \bibinfo{person}{Alexandra Birch}, \bibinfo{person}{Chris
  Callison-Burch}, \bibinfo{person}{Marcello Federico}, \bibinfo{person}{Nicola
  Bertoldi}, \bibinfo{person}{Brooke Cowan}, \bibinfo{person}{Wade Shen},
  \bibinfo{person}{Christine Moran}, \bibinfo{person}{Richard Zens},
  \bibinfo{person}{Chris Dyer}, \bibinfo{person}{Ondrej Bojar},
  \bibinfo{person}{Alexandra Constantin}, {and} \bibinfo{person}{Evan Herbst}.}
  \bibinfo{year}{2007}\natexlab{}.
\newblock \showarticletitle{{M}oses: Open Source Toolkit for Statistical
  Machine Translation}. In \bibinfo{booktitle}{\emph{ACL Companion Volume: Demo
  and Poster Sessions}}.
\newblock


\bibitem[\protect\citeauthoryear{Le, Tran, and Nguyen}{Le
  et~al\mbox{.}}{2020}]%
        {csax}
\bibfield{author}{\bibinfo{person}{Xuan-May Le}, \bibinfo{person}{Tuan Tran},
  {and} \bibinfo{person}{Hien Nguyen}.} \bibinfo{year}{2020}\natexlab{}.
\newblock \showarticletitle{An improvement of SAX representation for time
  series by using complexity invariance}.
\newblock \bibinfo{journal}{\emph{Intelligent Data Analysis}}
  \bibinfo{volume}{24} (\bibinfo{date}{05} \bibinfo{year}{2020}),
  \bibinfo{pages}{625--641}.
\newblock


\bibitem[\protect\citeauthoryear{Lin, J.~Keogh, Wei, and Lonardi}{Lin
  et~al\mbox{.}}{2007}]%
        {sax}
\bibfield{author}{\bibinfo{person}{Jessica Lin}, \bibinfo{person}{Eamonn
  J.~Keogh}, \bibinfo{person}{Li Wei}, {and} \bibinfo{person}{Stefano
  Lonardi}.} \bibinfo{year}{2007}\natexlab{}.
\newblock \showarticletitle{Experiencing {SAX}: A Novel Symbolic Representation
  of Time Series}.
\newblock \bibinfo{journal}{\emph{Data Mining and Knowledge Discovery}}
  \bibinfo{volume}{15} (\bibinfo{date}{08} \bibinfo{year}{2007}),
  \bibinfo{pages}{107--144}.
\newblock


\bibitem[\protect\citeauthoryear{Maner and Joyce}{Maner and Joyce}{1997}]%
        {wxsys}
\bibfield{author}{\bibinfo{person}{Walter Maner} {and} \bibinfo{person}{Sean
  Joyce}.} \bibinfo{year}{1997}\natexlab{}.
\newblock \showarticletitle{WXSYS Weather Lore + Fuzzy Logic = Weather
  Forecasts}.
\newblock  (\bibinfo{date}{01} \bibinfo{year}{1997}).
\newblock
\urldef\tempurl%
\url{https://www.researchgate.net/publication/237546595_WXSYS_Weather_Lore_Fuzzy_Logic_Weather_Forecasts}
\showURL{%
\tempurl}


\bibitem[\protect\citeauthoryear{Menne, Durre, Korzeniewski, McNeal, Thomas,
  Yin, Anthony, Ray, Vose, Gleason, and Houston}{Menne et~al\mbox{.}}{2020}]%
        {ncei}
\bibfield{author}{\bibinfo{person}{Matthew~J. Menne}, \bibinfo{person}{Imke
  Durre}, \bibinfo{person}{Bryant Korzeniewski}, \bibinfo{person}{Shelley
  McNeal}, \bibinfo{person}{Kristy Thomas}, \bibinfo{person}{Xungang Yin},
  \bibinfo{person}{Steven Anthony}, \bibinfo{person}{Ron Ray},
  \bibinfo{person}{Russell~S. Vose}, \bibinfo{person}{Byron~E. Gleason}, {and}
  \bibinfo{person}{Tamara~G. Houston}.} \bibinfo{year}{2020}\natexlab{}.
\newblock \bibinfo{title}{Global Historical Climatology Network - Daily
  (GHCN-Daily), Version 3}.
\newblock
\newblock
\urldef\tempurl%
\url{https://www.ncei.noaa.gov/}
\showURL{%
\tempurl}


\bibitem[\protect\citeauthoryear{Moyse and Lesot}{Moyse and Lesot}{2016}]%
        {period}
\bibfield{author}{\bibinfo{person}{Gilles Moyse} {and}
  \bibinfo{person}{Marie-Jeanne Lesot}.} \bibinfo{year}{2016}\natexlab{}.
\newblock \showarticletitle{Linguistic summaries of locally periodic time
  series}.
\newblock \bibinfo{journal}{\emph{Fuzzy Sets and Systems}}
  \bibinfo{volume}{285} (\bibinfo{year}{2016}), \bibinfo{pages}{94 -- 117}.
\newblock
\showISSN{0165-0114}


\bibitem[\protect\citeauthoryear{Murakami, Watanabe, Miyazawa, Goshima, Yanase,
  Takamura, and Miyao}{Murakami et~al\mbox{.}}{2017}]%
        {murakami2017}
\bibfield{author}{\bibinfo{person}{Soichiro Murakami}, \bibinfo{person}{Akihiko
  Watanabe}, \bibinfo{person}{Akira Miyazawa}, \bibinfo{person}{Keiichi
  Goshima}, \bibinfo{person}{Toshihiko Yanase}, \bibinfo{person}{Hiroya
  Takamura}, {and} \bibinfo{person}{Yusuke Miyao}.}
  \bibinfo{year}{2017}\natexlab{}.
\newblock \showarticletitle{Learning to Generate Market Comments from Stock
  Prices}. In \bibinfo{booktitle}{\emph{Proceedings of the 55th Annual Meeting
  of the Association for Computational Linguistics}}.
\newblock


\bibitem[\protect\citeauthoryear{Papineni, Roukos, Ward, and Zhu}{Papineni
  et~al\mbox{.}}{2002}]%
        {bleu}
\bibfield{author}{\bibinfo{person}{Kishore Papineni}, \bibinfo{person}{Salim
  Roukos}, \bibinfo{person}{Todd Ward}, {and} \bibinfo{person}{Wei-Jing Zhu}.}
  \bibinfo{year}{2002}\natexlab{}.
\newblock \showarticletitle{Bleu: a Method for Automatic Evaluation of Machine
  Translation}. In \bibinfo{booktitle}{\emph{Annual Meeting of the Association
  for Computational Linguistics}}.
\newblock


\bibitem[\protect\citeauthoryear{Peel, Douglas, and Lawton}{Peel
  et~al\mbox{.}}{2007}]%
        {peeldiabetes}
\bibfield{author}{\bibinfo{person}{Elizabeth Peel}, \bibinfo{person}{Margaret
  Douglas}, {and} \bibinfo{person}{Julia Lawton}.}
  \bibinfo{year}{2007}\natexlab{}.
\newblock \showarticletitle{Self monitoring of blood glucose in type 2
  diabetes: longitudinal qualitative study of patients' perspectives}.
\newblock \bibinfo{journal}{\emph{BMJ}} \bibinfo{volume}{335},
  \bibinfo{number}{7618} (\bibinfo{date}{Sep} \bibinfo{year}{2007}),
  \bibinfo{pages}{493}.
\newblock


\bibitem[\protect\citeauthoryear{Rawassizadeh, Momeni, Dobbins, Gharibshah, and
  Pazzani}{Rawassizadeh et~al\mbox{.}}{2016}]%
        {rawassizadeh_scalable_2016}
\bibfield{author}{\bibinfo{person}{Reza Rawassizadeh}, \bibinfo{person}{Elaheh
  Momeni}, \bibinfo{person}{Chelsea Dobbins}, \bibinfo{person}{Joobin
  Gharibshah}, {and} \bibinfo{person}{Michael Pazzani}.}
  \bibinfo{year}{2016}\natexlab{}.
\newblock \showarticletitle{Scalable {Daily} {Human} {Behavioral} {Pattern}
  {Mining} from {Multivariate} {Temporal} {Data}}.
\newblock \bibinfo{journal}{\emph{IEEE Transactions on Knowledge and Data
  Engineering}} \bibinfo{volume}{28}, \bibinfo{number}{11}
  (\bibinfo{date}{Nov.} \bibinfo{year}{2016}), \bibinfo{pages}{3098--3112}.
\newblock


\bibitem[\protect\citeauthoryear{Reiter and Dale}{Reiter and Dale}{2000}]%
        {reiter}
\bibfield{author}{\bibinfo{person}{Ehud Reiter} {and} \bibinfo{person}{Robert
  Dale}.} \bibinfo{year}{2000}\natexlab{}.
\newblock \bibinfo{booktitle}{\emph{Building Natural Language Generation
  Systems}}.
\newblock \bibinfo{publisher}{Cambridge University Press}.
\newblock


\bibitem[\protect\citeauthoryear{Sanchez-Valdes, Alvarez-Alvarez, and
  Trivino}{Sanchez-Valdes et~al\mbox{.}}{2016}]%
        {selftrack}
\bibfield{author}{\bibinfo{person}{Daniel Sanchez-Valdes},
  \bibinfo{person}{Alberto Alvarez-Alvarez}, {and} \bibinfo{person}{Gracian
  Trivino}.} \bibinfo{year}{2016}\natexlab{}.
\newblock \showarticletitle{Dynamic linguistic descriptions of time series
  applied to self-track the physical activity}.
\newblock \bibinfo{journal}{\emph{Fuzzy Sets and Systems}}
  \bibinfo{volume}{285} (\bibinfo{year}{2016}), \bibinfo{pages}{162 -- 181}.
\newblock
\showISSN{0165-0114}


\bibitem[\protect\citeauthoryear{Sch{\"a}fer and H{\"o}gqvist}{Sch{\"a}fer and
  H{\"o}gqvist}{2012}]%
        {sfa}
\bibfield{author}{\bibinfo{person}{Patrick Sch{\"a}fer} {and}
  \bibinfo{person}{Mikael H{\"o}gqvist}.} \bibinfo{year}{2012}\natexlab{}.
\newblock \showarticletitle{SFA: A symbolic fourier approximation and index for
  similarity search in high dimensional datasets}. In
  \bibinfo{booktitle}{\emph{Proceedings of the 15th International Conference on
  Extending Database Technology}}. \bibinfo{pages}{516 -- 527}.
\newblock


\bibitem[\protect\citeauthoryear{Sripada, Reiter, Hunter, and Yu}{Sripada
  et~al\mbox{.}}{2003}]%
        {sumtime}
\bibfield{author}{\bibinfo{person}{Somayajulu~G. Sripada},
  \bibinfo{person}{Ehud Reiter}, \bibinfo{person}{Jim Hunter}, {and}
  \bibinfo{person}{Jin Yu}.} \bibinfo{year}{2003}\natexlab{}.
\newblock \showarticletitle{Generating English summaries of time series data
  using the Gricean maxims}. In \bibinfo{booktitle}{\emph{Proceedings of the
  ninth ACM SIGKDD international conference on Knowledge discovery and data
  mining}}. \bibinfo{pages}{187--196}.
\newblock


\bibitem[\protect\citeauthoryear{Sun and Costello}{Sun and Costello}{2018}]%
        {sundecision}
\bibfield{author}{\bibinfo{person}{Si Sun} {and} \bibinfo{person}{Kaitlin~L.
  Costello}.} \bibinfo{year}{2018}\natexlab{}.
\newblock \showarticletitle{Designing decision-support technologies for
  patient-generated data in type 1 diabetes}. In \bibinfo{booktitle}{\emph{AMIA
  Annual Proceedings}}. \bibinfo{pages}{1645–1654}.
\newblock


\bibitem[\protect\citeauthoryear{Torres}{Torres}{2019}]%
        {alphavantage}
\bibfield{author}{\bibinfo{person}{Romel Torres}.}
  \bibinfo{year}{2019}\natexlab{}.
\newblock \bibinfo{title}{Alpha Vantage}.
\newblock
\newblock
\urldef\tempurl%
\url{https://github.com/RomelTorres/alpha_vantage}
\showURL{%
\tempurl}


\bibitem[\protect\citeauthoryear{Ultsch}{Ultsch}{1993}]%
        {sig}
\bibfield{author}{\bibinfo{person}{A. Ultsch}.}
  \bibinfo{year}{1993}\natexlab{}.
\newblock \showarticletitle{Knowledge Extraction from Self-Organizing Neural
  Networks}. In \bibinfo{booktitle}{\emph{Information and Classification}},
  \bibfield{editor}{\bibinfo{person}{Otto Opitz}, \bibinfo{person}{Berthold
  Lausen}, {and} \bibinfo{person}{R{\"u}diger Klar}} (Eds.).
  \bibinfo{publisher}{Springer Berlin Heidelberg}, \bibinfo{address}{Berlin,
  Heidelberg}, \bibinfo{pages}{301--306}.
\newblock
\showISBNx{978-3-642-50974-2}


\bibitem[\protect\citeauthoryear{van~der Lee, Krahmer, and Wubben}{van~der Lee
  et~al\mbox{.}}{2018}]%
        {vanderlee}
\bibfield{author}{\bibinfo{person}{Chris van~der Lee}, \bibinfo{person}{Emiel
  Krahmer}, {and} \bibinfo{person}{Sander Wubben}.}
  \bibinfo{year}{2018}\natexlab{}.
\newblock \showarticletitle{Automated learning of templates for data-to-text
  generation: comparing rule-based, statistical and neural methods}. In
  \bibinfo{booktitle}{\emph{International Conference on Natural Language
  Generation}}.
\newblock


\bibitem[\protect\citeauthoryear{Weber and Achananuparp}{Weber and
  Achananuparp}{2016}]%
        {mfp}
\bibfield{author}{\bibinfo{person}{Ingmar Weber} {and}
  \bibinfo{person}{Palakorn Achananuparp}.} \bibinfo{year}{2016}\natexlab{}.
\newblock \showarticletitle{Insights from Machine-Learned Diet Success
  Prediction}. In \bibinfo{booktitle}{\emph{Pacific Symposium on
  Biocomputing}}.
\newblock


\bibitem[\protect\citeauthoryear{Wilbik and Kaymak}{Wilbik and Kaymak}{2015}]%
        {processes}
\bibfield{author}{\bibinfo{person}{Anna Wilbik} {and} \bibinfo{person}{Uzay
  Kaymak}.} \bibinfo{year}{2015}\natexlab{}.
\newblock \showarticletitle{Linguistic Summarization of Processes \textemdash a
  research agenda}. In \bibinfo{booktitle}{\emph{Conference of the
  International Fuzzy Systems Association and the European Society for Fuzzy
  Logic and Technology}}.
\newblock


\bibitem[\protect\citeauthoryear{Wilbik, Keller, and Alexander}{Wilbik
  et~al\mbox{.}}{2011}]%
        {eldercare}
\bibfield{author}{\bibinfo{person}{Anna Wilbik}, \bibinfo{person}{James~M.
  Keller}, {and} \bibinfo{person}{Gregory~L. Alexander}.}
  \bibinfo{year}{2011}\natexlab{}.
\newblock \showarticletitle{Linguistic summarization of sensor data for
  eldercare}.
\newblock \bibinfo{journal}{\emph{IEEE International Conference on Systems,
  Man, and Cybernetics}}.
\newblock


\bibitem[\protect\citeauthoryear{{Wu}, {Mendel}, and {Joo}}{{Wu}
  et~al\mbox{.}}{2010}]%
        {ifthen}
\bibfield{author}{\bibinfo{person}{Dongrui {Wu}}, \bibinfo{person}{Jerry~M.
  {Mendel}}, {and} \bibinfo{person}{Jhiin {Joo}}.}
  \bibinfo{year}{2010}\natexlab{}.
\newblock \showarticletitle{Linguistic summarization using IF-THEN rules}. In
  \bibinfo{booktitle}{\emph{International Conference on Fuzzy Systems}}.
  \bibinfo{pages}{1--8}.
\newblock


\bibitem[\protect\citeauthoryear{Yager}{Yager}{1982}]%
        {yagerapproach}
\bibfield{author}{\bibinfo{person}{Ronald~R. Yager}.}
  \bibinfo{year}{1982}\natexlab{}.
\newblock \showarticletitle{A new approach to the summarization of data}.
\newblock \bibinfo{journal}{\emph{Information Sciences}} \bibinfo{volume}{28},
  \bibinfo{number}{1} (\bibinfo{year}{1982}), \bibinfo{pages}{69 -- 86}.
\newblock


\bibitem[\protect\citeauthoryear{Zadeh}{Zadeh}{1975}]%
        {ZADEH1975}
\bibfield{author}{\bibinfo{person}{Lotfi~A. Zadeh}.}
  \bibinfo{year}{1975}\natexlab{}.
\newblock \showarticletitle{The concept of a linguistic variable and its
  application to approximate reasoning--I}.
\newblock \bibinfo{journal}{\emph{Information Sciences}} \bibinfo{volume}{8},
  \bibinfo{number}{3} (\bibinfo{year}{1975}), \bibinfo{pages}{199 -- 249}.
\newblock


\bibitem[\protect\citeauthoryear{Zadeh}{Zadeh}{1983}]%
        {fuzzyquant}
\bibfield{author}{\bibinfo{person}{Lotfi~A. Zadeh}.}
  \bibinfo{year}{1983}\natexlab{}.
\newblock \showarticletitle{A computational approach to fuzzy quantifiers in
  natural languages}.
\newblock \bibinfo{journal}{\emph{Computers \& Mathematics with Applications}}
  \bibinfo{volume}{9}, \bibinfo{number}{1} (\bibinfo{year}{1983}),
  \bibinfo{pages}{149 -- 184}.
\newblock


\bibitem[\protect\citeauthoryear{Zadeh}{Zadeh}{2002}]%
        {prototype}
\bibfield{author}{\bibinfo{person}{Lotfi~A. Zadeh}.}
  \bibinfo{year}{2002}\natexlab{}.
\newblock \showarticletitle{A prototype-centered approach to adding deduction
  capability to search engines-the concept of protoform}. In
  \bibinfo{booktitle}{\emph{IEEE Symposium on Intelligent Systems}}.
\newblock


\bibitem[\protect\citeauthoryear{Zaki}{Zaki}{2001}]%
        {spade}
\bibfield{author}{\bibinfo{person}{Mohammed~J. Zaki}.}
  \bibinfo{year}{2001}\natexlab{}.
\newblock \showarticletitle{SPADE: An Efficient Algorithm for Mining Frequent
  Sequences}.
\newblock \bibinfo{journal}{\emph{Machine Learning}} \bibinfo{volume}{42},
  \bibinfo{number}{1} (\bibinfo{date}{01 Jan} \bibinfo{year}{2001}),
  \bibinfo{pages}{31--60}.
\newblock


\end{thebibliography}


\end{document}


\maketitle

\appendix
\begin{table}[!h]
  \caption{Univariate Individual-Level Summaries for Carbohydrate Intake Data}
  \label{tab:carbsummaries}
  \small
  \begin{tabular}{p{1.5in}p{2.5in}p{0.15in}p{0.15in}p{0.15in}p{0.15in}p{0.15in}p{0.15in}}
    \toprule
    Protoform Type & Summary & $T_1$ & $T_2$ & $T_3$ & $T_4$ & $T_5$ & $T_6$\\\hline
    \midrule
    \texttt{Standard Evaluation (TW)} & In the past full week, your carbohydrate intake has been moderate. & N/A & N/A & 1 & 0 & 1 & 1 \\\hline
    \texttt{Standard Evaluation (sTW)} & On more than half of the days in the past week, your carbohydrate intake has been high. & 0.71 & 1 & 0.57 & 0 & 1 & 1\\\hline
    \texttt{Standard Evaluation + Goal} & On more than half of the days in the past week, you did not reach your goal to keep your carbohydrate intake low. & 0.71 & 0.51 & 0.57 & 0 & 1 & 1\\\hline
    \texttt{Comparison} & Your carbohydrate intake was about the same in week 24 as it was in week 12. & N/A & N/A & 1 & 0 & 1 & 1\\\hline
    \texttt{Comparison + Goal} & You did about the same overall with keeping your carbohydrate intake low in week 24 than you did in week 12. & N/A & N/A & 1 & 0 & 1 & 1\\\hline
    \texttt{Standard Trend} & Half of the time, your carbohydrate intake increases from one day to the next. & 1 & 0.82 & 0.5 & 0 & 1 & 1\\\hline
    \texttt{Cluster-Based Pattern} & This past week, your carbohydrate intake was moderate, then high, then very low, then high. During all of the weeks similar to this past one, your carbohydrate intake dropped the next week. & 1 & 1 & 0.6 & 0 & 1 & 0.03\\\hline
    \texttt{Standard Pattern} & The last time you had a week like this past one, your carbohydrate intake dropped the next week. & N/A & 1 & 1 & 0 & 1\\\hline
    \texttt{Day If-Then Pattern} & There is 100\% confidence that, when your carbohydrate intake follows the pattern of being low on a Wednesday, your carbohydrate intake tends to be low the next Thursday. & 1 & 0.8 & 0.2 & N/A & 1 & 0.5\\\hline
    \texttt{Day-Based Pattern} & Your carbohydrate intake tends to be very high on Saturdays. & 0.9 & 0.8 & 0.28 & 0 & 1 & 1 \\\hline
    \texttt{Goal Assistance} & In order to better to follow the 2000-calorie diet, you should increase your carbohydrate intake.& N/A & N/A & N/A & N/A & N/A & 1 \\
    \bottomrule
  \end{tabular}
\end{table}

\begin{table}[!h]
  \caption{Univariate Group-Level Summaries for Carbohydrate Intake Data}
  \label{tab:groupcarbsummaries}
  \small
  \begin{tabular}{p{1.5in}p{2.5in}p{0.15in}p{0.15in}p{0.15in}p{0.15in}p{0.15in}p{0.15in}}
    \toprule
    Protoform Type & Summary & $T_1$ & $T_2$ & $T_3$ & $T_4$ & $T_5$ & $T_6$\\\hline
    \midrule
    \texttt{Standard Evaluation (TW)} & Some of the participants in this study had a moderate carbohydrate intake in the past full week. & 1 & 0.85 & 0.35 & 0 & 1 & 1\\\hline
    \texttt{Standard Evaluation (sTW)} & Almost none of the participants in this study had a moderate carbohydrate intake on more than half of the days in the past week. & 1 & 1 & 0.04 & 0 & 0.05 & 1\\\hline
    \texttt{Standard Evaluation + Goal} & Some of the participants in this study did not reach their goal to keep their carbohydrate intake low on all of the days in the past week. & 0.76 & 0.9 & 0.42 & 0 & 1 & 1\\\hline
    \texttt{Comparison} & Some of the participants in this study had a higher carbohydrate intake in week 11 than they did in week 24. & 1 & 1 & 0.31 & 0 & 1 & 1\\\hline
    \texttt{Comparison + Goal} & Some of the participants in this study did not do as well with keeping their carbohydrate intake low in week 11 as they did in week 24. & 1 & 1 & 0.31 & 0 & 1 & 1 \\\hline
    \texttt{Standard Trend} & Half of the participants in this study increase their carbohydrate intake from one day to the next half of the time. & 0.52 & 1 & 0.55 & 0 & 1 & 1\\\hline
    \texttt{Cluster-Based Pattern} & After looking at clusters containing weeks similar to this past one, it can be seen that some of the participants with these clusters may see little to no change in their carbohydrate intake next week. & 1 & 0.67 & 0.32 & 0 & 1 & 1\\\hline
    \texttt{Standard Pattern} & Based on the most recent weeks similar to this past one, it can be seen that some of the participants may see a drop in their carbohydrate intake next week. & 1 & 0.67 & 0.33 & 0 & 1 & 1\\\hline
    \texttt{If-Then Pattern} & For all of the participants in this study, it is true that when their carbohydrate intake follows the pattern of being very low, their carbohydrate intake tends to be high the next day. & 1 & 0 & 1 & 0 & 1 & 0.5 \\\hline
    \texttt{Day If-Then Pattern} & For all of the participants in this study, it is true that when their carbohydrate intake follows the pattern of being high on a Tuesday, their carbohydrate intake tends to be moderate on a Wednesday. & 1 & 0 & 1 & 0 & 1 & 0.5 \\\hline
    \texttt{Day-Based Pattern} & Some of the participants in this study tend to have a low carbohydrate intake on Fridays. & 0.67 & 1 & 0.23 & 0 & 1 & 1\\
    \texttt{Goal Assistance} & Most of the participants in this study have been given advice to increase their carbohydrate intake. & 1 & 0.75 & 0.93 & 0 & 1 & 1\\\hline
    \bottomrule
  \end{tabular}
\end{table}

\begin{table}[!h]
  \caption{Univariate Individual-Level Summaries for Calorie Intake Data (TW = month)}
  \label{tab:caloriemonth}
  \small
  \begin{tabular}{p{1.5in}p{2.5in}p{0.15in}p{0.15in}p{0.15in}p{0.15in}p{0.15in}p{0.15in}}
    \toprule
    Protoform Type & Summary & $T_1$ & $T_2$ & $T_3$ & $T_4$ & $T_5$ & $T_6$\\\hline
    \midrule
    \texttt{Standard Evaluation (TW)} & In the past full month, your calorie intake has been moderate. & N/A & N/A & 1 & 0 & 1 & 1\\\hline
    \texttt{Standard Evaluation (sTW)} & On some of the days in the past month, your calorie intake has been high. & 1 & 0.82 & 0.3 & 0 & 1 & 1\\\hline
    \texttt{Standard Evaluation + Goal} & On more than half of the days in the past month, you did not reach your goal to keep your calorie intake low. & 1 & 0.56 & 0.73 & 0 & 1 & 1\\\hline
    \texttt{Comparison} & Your calorie intake was about the same in month 5 as it was in month 2. & N/A & N/A & 1 & 0 & 1 & 1\\\hline
    \texttt{Comparison + Goal} & You did about the same overall with keeping your calorie intake low in month 5 than you did in month 2. & N/A & N/A & 1 & 0 & 1 & 1\\\hline
    \texttt{Standard Trend} & Half of the time, your calorie intake increases from one day to the next. & 0.71 & 0.84 & 0.53 & 0 & 1 & 1\\\hline
    \texttt{Cluster-Based Pattern} & In month 4, your calorie intake was very high, then low, then very high, then high, then very low, then moderate, then very low, then moderate, then high, then low, then high, then low, then very high, then very low, then high, then very high, then high, then very high, then very low, then very high, then high. During all of the weeks similar to week 4, your calorie intake rose the next week. & 1 & 1 & 1 & 0 & 1 & 0\\\hline
    \texttt{Standard Pattern} & The last time you had a week similar to week 5, your calorie intake rose the next week. & N/A & N/A & 1 & 0 & 1 & 1\\\hline
    \texttt{Day If-Then Pattern} & There is 100\% confidence that, when your calorie intake follows the pattern of being very high on a Thursday, your calorie intake tends to be moderate the next Friday. & 1 & 0.83 & 0.17 & 1 & 0.5\\\hline
    \texttt{Day-Based Pattern} & Your calorie intake tends to be very high on Saturdays. & 1 & 0.82 & 0.32 & 0 & 1 & 1\\\hline
    \texttt{Goal Assistance} & In order to better to follow the 2000-calorie diet, you should decrease your calorie intake. & N/A & N/A & N/A & N/A & N/A & 1\\
    \bottomrule
  \end{tabular}
\end{table}

\begin{table}[!h]
  \caption{Univariate Group-Level Summaries for Calorie Intake Data (TW = month)}
  \label{tab:groupcaloriemonth}
  \small
  \begin{tabular}{p{1.5in}p{2.5in}p{0.15in}p{0.15in}p{0.15in}p{0.15in}p{0.15in}p{0.15in}}
    \toprule
    Protoform Type & Summary & $T_1$ & $T_2$ & $T_3$ & $T_4$ & $T_5$ & $T_6$\\\hline
    \midrule
    \texttt{Standard Evaluation (TW)} & Half of the participants in this study had a moderate calorie intake in the past full month. & 0.96 & 0.89 & 0.5 & 0 & 1 & 1\\\hline
    \texttt{Standard Evaluation (sTW)} & Almost none of the participants in this study had a high calorie intake on more than half of the days in the past month. & 1 & 1 & 0.01 & 0 & 0 & 1\\\hline
    \texttt{Standard Evaluation + Goal} & Some of the participants in this study reached their goal to keep their calorie intake low on most of the days in the past month. & 1 & 1 & 0.37 & 0 & 1 & 1\\\hline
    \texttt{Comparison} & Some of the participants in this study had a similar calorie intake in month 1 than they did in month 4. & 1 & 0.88 & 0.32 & 0 & 1 & 1\\\hline
    \texttt{Comparison + Goal} & Some of the participants in this study did about the same with keeping their calorie intake low in month 1 as they did in month 4. & 1 & 0.87 & 0.32 & 0 & 1 & 1\\\hline
    \texttt{Standard Trend} & More than half of the participants in this study increase their calorie intake from one day to the next half of the time. & 0.91 & 1 & 0.59 & 0 & 1 & 1\\\hline
    \texttt{Cluster-Based Pattern} & After looking at clusters containing months similar to this past one, it can be seen that half of the participants with these clusters may see little to no change in their calorie intake next month. & 0.64 & 0.68 & 0.46 & 0 & 1 & 1\\\hline
    \texttt{Standard Pattern} & Based on the most recent months similar to this past one, it can be seen that more than half of the participants may see little to no change in their calorie intake next month. & 0.94 & 0.68 & 0.49 & 0 & 1 & 1\\\hline
    \texttt{If-Then Pattern} & For all of the participants in this study, it is true that when their calorie intake follows the pattern of being very low, their calorie intake tends to be low the next day. & 1 & 0 & 1 & 0 & 1 & 0.5 \\\hline
    \texttt{Day If-Then Pattern} & For all of the participants in this study, it is true that when their calorie intake follows the pattern of being very low on a Monday, their calorie intake tends to be low on a Tuesday. & 1 & 0 & 1 & 0 & 1 & 0.5 \\\hline
    \texttt{Day-Based Pattern} & Some of the participants in this study tend to have a low calorie intake on Tuesdays. & 0.82 & 1 & 0.26 & 0 & 1 & 1\\\hline
    \texttt{Goal Assistance} & All of the participants in this study have been given advice to decrease their calorie intake. & 1 & 0 & 1 & 0 & 1 & 1\\
    \bottomrule
  \end{tabular}
\end{table}

\begin{table}[!h]
  \caption{Univariate Individual-Level Summaries for Calorie Intake Data (TW = None)}
  
  \label{tab:calorienotw}
  \small
  \begin{tabular}{p{1.5in}p{2.5in}p{0.15in}p{0.15in}p{0.15in}p{0.15in}p{0.15in}p{0.15in}}
    \toprule
    Protoform Type & Summary & $T_1$ & $T_2$ & $T_3$ & $T_4$ & $T_5$ & $T_6$\\\hline
    \midrule
    \texttt{Standard Evaluation (sTW)} & On some of the days, your calorie intake has been low. & 0.68 & 0.8 & 0.24 & 0 & 1 & 1 \\\hline
    \texttt{Standard Evaluation + Goal} & On more than half of the days, you did not reach your goal to keep your calorie intake low. & 1 & 0.57 & 0.75 & 0 & 1 & 1\\\hline

    \texttt{Standard Trend} & Half of the time, your calorie intake increases from one day to the next. & 0.71 & 0.84 & 0.53 & 0 & 1 & 1\\\hline
    \texttt{Day-Based Pattern} & Your calorie intake tends to be low on Thursdays. & 1 & 0.84 & 0.36 & 0 & 1 & 1\\
    \bottomrule
  \end{tabular}
\end{table}

\begin{table}[!h]
  \caption{Univariate Group-Level Summaries for Calorie Intake Data (TW = None)}
  \label{tab:groupcalorienotw}
  \small
  \begin{tabular}{p{1.5in}p{2.5in}p{0.15in}p{0.15in}p{0.15in}p{0.15in}p{0.15in}p{0.15in}}
    \toprule
    Protoform Type & Summary & $T_1$ & $T_2$ & $T_3$ & $T_4$ & $T_5$ & $T_6$\\\hline
    \midrule
    \texttt{Standard Evaluation (sTW)} & Some of the participants in this study had a low calorie intake on some of the days. & 1 & 1 & 0.32 & 0 & 1 & 1\\\hline
    \texttt{Standard Evaluation + Goal} & Half of the participants in this study reached their goal to keep their calorie intake low on most of the days. & 0.96 & 0.94 & 0.5 & 0 & 1 & 1\\\hline

    \texttt{Standard Trend} & More than half of the participants in this study increase their calorie intake from one day to the next half of the time. & 0.91 & 1 & 0.59 & 0 & 1 & 1\\\hline
    \texttt{Day-Based Pattern} & Some of the participants in this study tend to have a low calorie intake on Mondays. & 0.81 & 1 & 0.26 & 0 & 1 & 1\\
    \bottomrule
  \end{tabular}
\end{table}

\begin{table}[!h]
  \caption{Univariate Individual-Level Summaries for Calorie Intake Data (alphabet size = 3)}
  \label{tab:alpha3}
  \scriptsize
  \begin{tabular}{p{1.5in}p{2.5in}p{0.15in}p{0.15in}p{0.15in}p{0.15in}p{0.15in}p{0.15in}}
    \toprule
    Protoform Type & Summary & $T_1$ & $T_2$ & $T_3$ & $T_4$ & $T_5$ & $T_6$\\\hline
    \midrule
    \texttt{Standard Evaluation (TW)} & In the past full week, your calorie intake has been moderate. & N/A & N/A & 1 & 0 & 1 & 1\\\hline
    \texttt{Standard Evaluation (sTW)} & On more than half of the days in the past week, your calorie intake has been moderate. & 1 & 1 & 0.71 & 0 & 1 & 1\\\hline
    \texttt{Standard Evaluation + Goal} & On most of the days in the past week, you did not reach your goal to keep your calorie intake low. & 1 & 0.65 & 0.86 & 0 & 1 & 1\\\hline
    \texttt{Comparison} & Your calorie intake was about the same in week 24 as it was in week 12. & N/A & N/A & 1 & 0 & 1 & 1\\\hline
    \texttt{Comparison + Goal} & You did about the same overall with keeping your calorie intake low in week 24 than you did in week 12. & N/A & N/A & 1 & 0 & 1 & 1\\\hline
    \texttt{Standard Trend} & Half of the time, your calorie intake increases from one day to the next. & 0.71 & 0.84 & 0.53 & 0 & 1 & 1\\\hline
    \texttt{Cluster-Based Pattern} & In week 24, your calorie intake was moderate, then low, then moderate, then high, then moderate. During some of the weeks similar to week 24, your calorie intake rose the next week. & 1 & 0.68 & 0.4 & 0 & 1 & 0.03\\\hline
    \texttt{Standard Pattern} & The last time you had a week similar to week 24, your calorie intake dropped the next week. & N/A & N/A & 1 & 0 & 1 & 1\\\hline
    \texttt{Day If-Then Pattern} & There is 100\% confidence that, when your calorie intake follows the pattern of being high on a Saturday, your calorie intake tends to be high the next Sunday. & 1 & 0.64 & 0.36 & 0 & 1 & 0.5\\\hline
    \texttt{Day-Based Pattern} & Your calorie intake tends to be moderate on Mondays. & 1 & 1 & 0.32 & 0 & 1 & 1\\\hline
    \texttt{Goal Assistance} & In order to better to follow the 2000-calorie diet, you should decrease your calorie intake. & N/A & N/A & N/A & N/A & N/A & 1\\
    \bottomrule
  \end{tabular}
\end{table}

\begin{table}[!h]
  \caption{Univariate Group-Level Summaries for Calorie Intake Data (alphabet size = 3)}
  \label{tab:groupalpha3}
  \small
  \begin{tabular}{p{1.5in}p{2.5in}p{0.15in}p{0.15in}p{0.15in}p{0.15in}p{0.15in}p{0.15in}}
    \toprule
    Protoform Type & Summary & $T_1$ & $T_2$ & $T_3$ & $T_4$ & $T_5$ & $T_6$\\\hline
    \midrule
    \texttt{Standard Evaluation (TW)} & Half of the participants in this study had a moderate calorie intake in the past full week. & 0.7 & 0.69 & 0.53 & 0 & 1 & 1\\\hline
    \texttt{Standard Evaluation (sTW)} & Some of the participants in this study had a moderate calorie intake on more than half of the days in the past week. & 0.72 & 1 & 0.24 & 0 & 1 & 1\\\hline
    \texttt{Standard Evaluation + Goal} & Some of the participants in this study reached their goal to keep their calorie intake low on all of the days in the past week. & 0.84 & 0.87 & 0.42 & 0 & 1 & 1\\\hline
    \texttt{Comparison} & Some of the participants in this study had a similar calorie intake in week 11 than they did in week 24. & 1 & 1 & 0.3 & 0 & 1 & 1\\\hline
    \texttt{Comparison + Goal} & Some of the participants in this study did about the same with keeping their calorie intake low in week 11 as they did in week 24. & 1 & 1 & 0.3 & 0 & 1 & 1\\\hline
    \texttt{Standard Trend} & More than half of the participants in this study increase their calorie intake from one day to the next half of the time. & 0.91 & 1 & 0.59 & 0 & 1 & 1\\\hline
    \texttt{Cluster-Based Pattern} & After looking at clusters containing weeks similar to this past one, it can be seen that half of the participants with these clusters may see little to no change in their calorie intake next week. & 0.77 & 0.71 & 0.58 & 0 & 1 & 1\\\hline
    \texttt{Standard Pattern} & Based on the most recent weeks similar to this past one, it can be seen that more than half of the participants may see little to no change in their calorie intake next week. & 0.8 & 0.71 & 0.58 & 0 & 1 & 1\\\hline
    \texttt{If-Then Pattern} & For all of the participants in this study, it is true that when their calorie intake follows the pattern of being high, their calorie intake tends to be moderate the next day. & 1 & 0 & 1 & 0 & 1 & 0.5 \\\hline
    \texttt{Day If-Then Pattern} & For all of the participants in this study, it is true that when their calorie intake follows the pattern of being high on a Saturday, their calorie intake tends to be high on a Sunday. & 1 & 0 & 1 & 0 & 1 & 0.5 \\\hline
    \texttt{Day-Based Pattern} & Almost none of the participants in this study tend to have a low calorie intake on Mondays. & 1 & 1 & 0.02 & 0 & 0 & 1\\\hline
    \texttt{Goal Assistance} & All of the participants in this study have been given advice to decrease their calorie intake. & 1 & 0 & 1 & 0 & 1 & 1\\
    \bottomrule
  \end{tabular}
\end{table}

\begin{table}[!h]
  \caption{Univariate Individual-Level Summaries for Calorie Intake Data (alphabet size = 7)}
  \label{tab:alpha7}
  \small
  \begin{tabular}{p{1.5in}p{2.5in}p{0.15in}p{0.15in}p{0.15in}p{0.15in}p{0.15in}p{0.15in}}
    \toprule
    Protoform Type & Summary & $T_1$ & $T_2$ & $T_3$ & $T_4$ & $T_5$ & $T_6$\\\hline
    \midrule
    \texttt{Standard Evaluation (TW)} & In the past full week, your calorie intake has been moderate. & N/A & N/A & 1 & 0 & 1 & 1\\\hline
    \texttt{Standard Evaluation (sTW)} & On some of the days in the past week, your calorie intake has been low. & 0.93 & 1 & 0.29 & 0 & 1 & 1\\\hline
    \texttt{Standard Evaluation + Goal} & On most of the days in the past week, you did not reach your goal to keep your calorie intake low. & 1 & 0.65 & 0.86 & 0 & 1 & 1\\\hline
    \texttt{Comparison} & Your calorie intake was about the same in week 24 as it was in week 12. & N/A & N/A & 1 & 0 & 1 & 1\\\hline
    \texttt{Comparison + Goal} & You did about the same overall with keeping your calorie intake low in week 24 than you did in week 12. & N/A & N/A & 1 & 0 & 1 & 1\\\hline
    \texttt{Standard Trend} & Half of the time, your calorie intake increases from one day to the next. & 0.71 & 0.84 & 0.53 & 0 & 1 & 1\\\hline
    \texttt{If-Then Pattern} & There is 100\% confidence that, when your calorie intake follows the pattern of being very low, your calorie intake tends to be low the next day. & 0.9 & 0.72 & 0.28 & 0 & 1 & 0.5\\\hline
    \texttt{Day If-Then Pattern} & There is 100\% confidence that, when your calorie intake follows the pattern of being extremely high on a Saturday, your calorie intake tends to be extremely high the next Sunday. & 0.7 & 0.76 & 0.24 & 0 & 1 & 0.5\\\hline
    \texttt{Day-Based Pattern} & Your calorie intake tends to be low on Mondays. & 0.7 & 0.88 & 0.24 & 0 & 1 & 1\\\hline
    \texttt{Goal Assistance} & In order to better to follow the 2000-calorie diet, you should decrease your calorie intake. & N/A & N/A & N/A & N/A & N/A & 1\\
    \bottomrule
  \end{tabular}
\end{table}

\begin{table}[!h]
  \caption{Univariate Group-Level Summaries for Calorie Intake Data (alphabet size = 7)}
  \label{tab:groupalpha7}
  \small
  \begin{tabular}{p{1.5in}p{2.5in}p{0.15in}p{0.15in}p{0.15in}p{0.15in}p{0.15in}p{0.15in}}
    \toprule
    Protoform Type & Summary & $T_1$ & $T_2$ & $T_3$ & $T_4$ & $T_5$ & $T_6$\\\hline
    \midrule
    \texttt{Standard Evaluation (TW)} & Some of the participants in this study had a moderate calorie intake in the past full week. & 0.79 & 0.89 & 0.26 & 0 & 1 & 1\\\hline
    \texttt{Standard Evaluation (sTW)} & Almost none of the participants in this study had a low calorie intake on more than half of the days in the past week. & 1 & 1 & 0.03 & 0 & 0.02 & 1\\\hline
    \texttt{Standard Evaluation + Goal} & Some of the participants in this study reached their goal to keep their calorie intake low on all of the days in the past week. & 0.84 & 0.87 & 0.42 & 0 & 1 & 1\\\hline
    \texttt{Comparison} & Some of the participants in this study had a higher calorie intake in week 11 than they did in week 24. & 1 & 1 & 0.32 & 0 & 1 & 1\\\hline
    \texttt{Comparison + Goal} & Some of the participants in this study did not do as well with keeping their calorie intake low in week 11 as they did in week 24. & 1 & 1 & 0.32 & 0 & 1 & 1\\\hline
    \texttt{Standard Trend} & More than half of the participants in this study increase their calorie intake from one day to the next half of the time. & 0.91 & 1 & 0.59 & 0 & 1 & 1\\\hline
    \texttt{Cluster-Based Pattern} & After looking at clusters containing weeks similar to this past one, it can be seen that some of the participants with these clusters may see a drop in their calorie intake next week. & 1 & 0.68 & 0.3 & 0 & 1 & 1\\\hline
    \texttt{Standard Pattern} & Based on the most recent weeks similar to this past one, it can be seen that some of the participants may see a rise in their calorie intake next week. & 1 & 0.67 & 0.36 & 0 & 1 & 1\\\hline
    \texttt{If-Then Pattern} & For all of the participants in this study, it is true that when their calorie intake follows the pattern of being very low, their calorie intake tends to be high the next day. & 1 & 0 & 1 & 0 & 1 & 0.5 \\\hline
    \texttt{Day If-Then Pattern} & For all of the participants in this study, it is true that when their calorie intake follows the pattern of being moderate on a Monday, their calorie intake tends to be high on a Tuesday. & 1 & 0 & 1 & 0 & 1 & 0.5 \\\hline
    \texttt{Day-Based Pattern} & Almost none of the participants in this study tend to have a moderate calorie intake on Mondays. & 1 & 1 & 0.01 & 0 & 0 & 1\\\hline
    \texttt{Goal Assistance} & All of the participants in this study have been given advice to decrease their calorie intake. & 1 & 0 & 1 & 0 & 1 & 1\\
    \bottomrule
  \end{tabular}
\end{table}